%% file: main.tex
\documentclass[runningheads]{llncs}
\usepackage{etoolbox}
\usepackage{graphicx}

\usepackage[ruled,vlined]{algorithm2e}
\usepackage{placeins}
\usepackage{float}

\usepackage{xurl}

\usepackage{latexsym,verbatim}
\usepackage{xspace}
\usepackage{xcolor}

\usepackage{amsmath}
\usepackage{amssymb}

\usepackage{caption}

\usepackage{hyperref}

\usepackage{enumitem}
\setlist{topsep=2pt}

\AtBeginDocument{
	\setlength{\abovedisplayshortskip}{2pt plus 1pt minus 1pt}
	\setlength{\abovedisplayskip}{4pt plus 2pt minus 2pt}
	\setlength{\belowdisplayskip}{\abovedisplayskip}
	\setlength{\belowdisplayshortskip}{\abovedisplayshortskip}
}

\newcommand{\xseq}[2]{#1_1,\dots,#1_{#2}}

\newcommand{\FACTS}{\ensuremath{F}\xspace}
\newcommand{\PROP}{\ensuremath{\mathrm{PROP}}\xspace}
\newcommand{\PLit}{\ensuremath{\mathrm{PLit}}\xspace}
\newcommand{\LIT}{\ensuremath{\mathrm{Lit}}\xspace}
\newcommand{\ModLit}{\ensuremath{\mathrm{ModLit}}\xspace}

\newcommand{\Rex}{\ensuremath{\mathrm{REx}}\xspace}

\newcommand{\LABELS}{\ensuremath{\mathrm{Lab}}\xspace}

\newcommand{\Cost}{\ensuremath{\mathsf{C}}\xspace}
\newcommand{\Obl}{\ensuremath{\mathsf{O}}\xspace}
\newcommand{\Perm}{\ensuremath{\mathsf{P}}\xspace}

\newcommand{\TO}{\rightarrow}
\newcommand{\To}{\Rightarrow}
\newcommand{\defeater}{\leadsto}

\newcommand{\de}{\ensuremath{\mathrm{d}}\xspace}
\newcommand{\dft}{\ensuremath{\mathrm{dft}}\xspace}
\newcommand{\non}{\ensuremath{\mathcal{\sim }}}

\definecolor{ametista}{rgb}{0.6, 0.33, 0.73}
\definecolor{amethyst}{rgb}{0.6, 0.4, 0.8}

\newcommand{\viola}[1]{{\color{ametista} #1}}

\definecolor{green}{RGB}{48,184,136}
\definecolor{teal}{RGB}{0,115,119}
\definecolor{alma}{RGB}{171,27,31}
\definecolor{verdeVr}{RGB}{33,117,0}

\undef{\comment}
\usepackage{changes}

\definechangesauthor[color=ametista]{Ce}
\definechangesauthor[color=green]{Gui}
\definechangesauthor[color=alma]{Nino}
\definechangesauthor[color=verdeVr]{Matteo}

\newcommand{\set}[2][\relax]{\ensuremath{#1\{#2#1\}}}
\newcommand{\Set}[2][\relax]{%
  \ensuremath{
    \ifx#1\left
      #1\{#2\right\}
    \else\ifx#1\right
      \left\{#2#1\}
      \else #1\{#2#1\}\fi\fi}%
}

\def\dRule#1:#2=>#3#4{#1\colon #2\To_{#3}#4}
\newcommand{\drule}[4][\Cost]{#2\colon #3\To_{#1}#4}
\newcommand{\Drule}[4][\Cost]{#2\colon #3\defeater_{#1}#4}
\newcommand{\HB}{\mathit{HB}}
\newcommand{\MHB}{\mathit{MHB}}
\newcommand{\infd}{\mathit{infd}}

\begin{document}

\title{Deontic Meta-Rules}
\renewcommand{\orcidID}[1]{\relax}

\author{Francesco Olivieri\inst{1}\orcidID{0000-0003-0838-9850} \and
Guido Governatori\inst{2}\orcidID{0000-0002-9878-2762} \and
Matteo Cristani\inst{3}\orcidID{0000-0001-5680-0080} \and
Antonino Rotolo\inst{4}\orcidID{0000-0001-5265-0660} \and
Abdul Sattar\inst{1}\orcidID{0000-0002-2567-2052}
}
\authorrunning{Francesco Olivieri et al.}
%
\institute{School of Information and Communication Technology, IIIS, Griffith
University, Nathan, QLD 4111, Australia \ \email{\{f.oliveri;a.sattar\}@griffith.edu.au}
\and
Centre for Computational Law, Singapore Management University, Singapore
\email{guido@governatori.net} 
\and
University of Verona, Verona, 37136, Italy
\email{matteo.cristani@univr.it}
\and
Alma AI, University of Bologna, Bologna, 40121, Italy
\email{antonino.rotolo@unibo.it}
}

\maketitle              
\begin{abstract}

The use of meta-rules in logic, i.e., rules whose content includes other rules, has recently gained attention in the setting of non-monotonic reasoning:
a first logical formalisation and efficient algorithms to compute the (meta)-extensions of such theories were proposed in \cite{Olivieri202169}.
This work extends such a logical framework by considering the deontic aspect. The resulting logic will not just be able to model policies 
but also tackle well-known aspects that occur in numerous legal systems. The use of Defeasible Logic (DL) to model meta-rules in the application area we just alluded to has been investigated. Within this line of research, the study mentioned above was not focusing on the general computational properties of meta-rules. 

This study fills this gap with two major contributions. First, we introduce and formalise two variants of Defeasible Deontic Logic with Meta-Rules to represent (1) defeasible meta-theories with deontic modalities, and (2) two different types of conflicts among rules: Simple Conflict Defeasible Deontic Logic, and  Cautious Conflict Defeasible Deontic Logic. Second, we advance efficient algorithms to compute the extensions for both variants.

\end{abstract}

\section{Introduction and Background}\label{sec:Intro}

\input{Intro2}

\input{Methodology}

\section{Defeasible Deontic Logic with Meta-Rules}\label{sec:Logic}
\input{Logic}

\section{Algorithms}\label{sec:Algo}
\input{Algorithms}

\input{relatedworkmetalogic}

\section{Conclusions and Further Developments}\label{sec:Conc}
\input{Conclusions}


\bibliographystyle{splncs04}
\bibliography{bibexport}
		\end{document}

%% file: Intro2.tex
The AI community has devoted extensive research to the development of
rule-based systems in the normative domain. It is widely acknowledged that norms have, basically, a conditional structure like
\[
\textsc{If } a_1,\dots ,a_n \textsc{ Then } b
\]
where $a_1,\dots ,a_n$ are the applicability conditions of the norm, and $b$ denotes the legal effect which ought to follow when such applicability conditions hold. This, very general, view highlights a direct link between the concepts of norm and rule.

An important aspect of normative reasoning is the introduction of meta-norms (i.e., meta-rules): Norms conferring the power to produce other norms, justifying their choice or their enactment, as well as governing their application and dynamics \cite{hart:1994,Handbook:2021,GelGovRotSar:AIL:04:Normative}.

This paper develops, in the context of normative reasoning, a rule-based logical framework providing an efficient computation of meta-rules, i.e., rules having rules as their elements. We shall proceed with the simple idea that a rule is a (binary) relation between a set of (applicability) conditions and a conclusion. The meaning of such a relation is to determine under which conditions a conclusion can be derived. Meta-rules thus generalise such an idea by establishing that, in addition to standard conclusions and conditions, rules themselves can be the ``conclusion'' and the elements of the set of conditions. In other terms, new rules can be generated from other rules, along with conditions that are not rules per se.
 
An interesting discussion can be developed in the normative domain, as meta-rules (or rules with nested rules)  frequently occur not only in normative reasoning, but in other contexts as well including, for instance, policies for security systems. Often, when a set of rules represents a set of policies, we have to consider a policy that contains conditions (rules) about \emph{itself} (and also about other sets of rules/policies). Consider the example provided in \cite{ruleml05insu}, where a company has a security policy specifying that: (i) a piece of information is deemed confidential when its disclosure would harm the interests of the company, and (ii) confidential information must be protected (and hence cannot be disclosed). Such a policy can be naturally represented by the meta-rule
\[
 \big(\mathit{Disclose}(x) \rightarrow \mathit{HarmInterests}\big) \rightarrow \mathit{Confidential}(x).
\]
Now, in this policy, the condition about harming the interests should be represented by a hypothetical expression: an `$\textsc{If}\, (\dots)\, \textsc{Then}\, (\dots)$' rule would appear to be the most natural way to represent such a construct. Furthermore, the hypothesis is part of the conditions to define when a piece of information is confidential (in this case, it is the condition itself). Unfortunately, given its well-known paradoxes, we cannot use classical material implication ($\supset$). Consequently, if we model the policy as
\[
  (\mathit{Disclose}(x) \supset \mathit{HarmInterests}) \rightarrow \mathit{Confidential}(x),
\]
given the equivalence between `$\mathit{Disclose}(x) \supset \mathit{HarmInterests}$' and `$\neg\mathit{Disclose}(x) \vee \mathit{HarmInterests}$', we have the counter-intuitive scenarios where (1) if $x$ is not disclosed then $x$ is confidential (information that is not confidential, no matter if disclosed or not, it does not need to be protected), and (2) if, for any reason, company interests are harmed, then $x$ is confidential, for any piece of information. The policy can neither be defined as 
\[
  \big(\mathit{Disclose}(x) \wedge \mathit{HarmInterests}\big) \rightarrow \mathit{Confidential}(x),
\]
given that a disclosed piece of information (with the consequent harm of interests) can no longer be considered confidential. 

Another situation where meta-rules are useful is when the inclusion of a rule in a set of rules depends upon whether other rules are already in the system. For instance, we can have
\[
 r_1 \rightarrow r_2,
\]
indicating that the existence of rule $r_2$ in the system depends upon the existence in the system of rule $r_1$. For instance, this could be the case of an \emph{implementing decree} whose mere existence depends upon the existence of the law whose prescriptions are indeed implemented.
 
Typically, such dependencies among rules are stored externally. Still, suppose we model them directly into the system by using meta-rules. In that case, we can include (or remove) $r_2$ automatically, based on the other rules it depends upon (and thus automatising the system's maintenance functionalities). In addition, this feature is beneficial to system integration, as it supports context-dependent rules. 

The definition of context-dependent policies is valuable in many situations; for instance, the defence forces of a country can have different rules of engagement, depending on the environment in which they are situated. One might think that a simple (and somehow naive) way to achieve this would be to partition the rules into independent rule sets, one for each context, and then to use simpler rules (without nested rules). However, as discussed before, there could be dependencies among the rules, and the environments themselves could be defined in terms of rules. Accordingly, a clear partition of the simple rules might not be feasible. 

Another area of legal reasoning, where meta-rules proved to be essential in representing the legal processes at hand, is related to the field of norm change. As we argued (see also \cite{Governatori2009157}), many legislative instruments contain provisions (norms) about who has the power to create, modify, revoke, or abrogate other norms. If norms can be represented as rules \cite{Sartor05}, and there are norms `about other norms', then it would be natural to have rules whose content consists of other rules. 
Different variants of Defeasible Logic have been proposed \cite{Governatori2009157,Cristani201739} to incorporate meta-rules to describe the logical behaviour of norm changes. An important aspect of norm change is that some legal systems can specify that certain norms cannot exist (or cannot be in force) in a specific legal system. For instance, in the Italian Constitution, Article 27 prescribes that there cannot be norms in the Italian legal system prescribing Capital Punishment. This means that a meta-norm can speak about the positive existence of a rule and prevent a rule from being generated in the system itself after the enforcement of the meta-norm.  

Finally, meta-rules have been proposed to model private international law, which is the body of rules and principles governing the choice of law to be applied, when there are conflicts in the domestic law of different countries related to private legal facts and transactions \cite{stone}. Works such as \cite{DungS11,MalerbaRG16,Rotolo:2021} argued that we need a reasoning mechanism that allows us 
\begin{itemize}
    \item To conclude that if something holds in some legal system, then some norms hold in this or another legal system;
    
    \item To import, in a given system, any piece of information holding in another system.
\end{itemize}
Meta-rules are thus relevant to modelling such
an interaction.

In this paper, we extensively study a new computational framework (based on a variant of Defeasible Deontic Logic) that models rules and meta-rules for the normative domain. 
For meta-rules, we will allow only one level of nesting. 
Thus a rule can appear inside a meta-rule, but a meta-rule cannot appear inside other meta-rules. This restriction has two main motivations. First, as we discussed, meta-rule can represent norms or policies about other norms and policies; while it is possible to have multiple levels, they are relatively rare in real applications. Secondly, our work is partially motivated by computational aspects (e.g., maintaining a feasible computational complexity). It is unclear whether extending the language with nested meta-rule affords to keep the computational complexity under control.

We shall distinguish between the \emph{content} of a rule (which is a binary relation between a set of premises and a conclusion, both represented as propositions in an underlying, given language), and the \emph{name} (or  \emph{label}, the identifier) of the rule itself. In this setup, a rule can be understood as a function associating a label to the content of the rule. Similarly, a meta-rule is still a function that associates a label to the content, but now the elements of the binary relation corresponding to the content of the meta-rule can contain other rules.

\section{Synopsis and Structure of the paper}\label{sec:synopsis}
In devising a new computational logical system for meta-rules in deontic reasoning, we shall adopt the following road map and take the following steps:
\begin{itemize}
  \item We offer a conceptual analysis of meta-reasoning in the normative domain;
  \item We present the new logical framework;
  \item We investigate the computational properties of the logic.
\end{itemize}  

We thus commence with Section~\ref{sec:conceptual_framework} by setting up a conceptual framework where some basic philosophical problems are discussed; such a conceptual framework provides a discussion that frames meta-rules in legal reasoning and is hence the basis for our formal choices. For the sake of illustration, we discuss and consider (1) what are the complexities behind the definition of norm change mechanisms in domains such as the law and (2) the interplay of norm dynamics and deontic concepts such as permissions and permissive norms. 

Once this is done, we move on and provide the proper formalisation of the logical apparatus that models deontic and normative reasoning, and that which is enriched with meta-rules. Such a formalism must take as its starting point the above-mentioned conceptual analysis and be as rich as possible. 

To help the reader, the logic is presented progressively. As the technicalities of the proof-theoretic part can be harsh to grasp for a non-expert reader, a gentle introduction to the formal machinery is thus offered: in particular, we split the logic presentation and formalisation in two distinct sections.

Section \ref{sec:Method} presents the Defeasible Deontic Logic's framework of \cite{GovernatoriORS13}. Its main purpose is twofold: First, it introduces the reader to all the notions (some specific to Defeasible Logic) that will be needed later, such as: (i) explaining the meaning of a rule being applicable or discarded and their importance, (ii) representing the different types of rules -- strict, defeasible and defeaters -- (iii) modelling the distinction between constitutive and deontic rules, (iii) how such a framework is capable of solving conflicts, or again (iv) handling complex reasoning patterns related to contrary-to-duty and compensatory obligations (as well as various types of deontic statements). Second, all those notions are formalised: language, strict and defeasible rules vs constitutive, obligation and permission rules, and so on. For every definition and concept, we thoroughly explain their meaning and give examples.  

Following Section~\ref{sec:Logic} where we present Defeasible Deontic Logic with Meta-Rules: such a framework specialises the framework of \cite{GovernatoriORS13} by (i) representing the distinction between rules and meta-rules, (ii) introducing and formalising different \emph{variants} of \emph{conflict} among meta-rules, and naturally (iii) providing the proof theory extended with meta-rules and such conflicts. We end that section by proving the coherence and consistency of the logical apparatus.

Lastly, we prove that the logic proposed is correct and computationally efficient. Section~\ref{sec:Algo} thus advances algorithms that compute the extension of a given input logic and explain their behaviours through two examples. We end that section by studying their complexity and proving their correctness.

This work is concluded by Section~\ref{sec:RelatedWork}, that reviews current relevant literature, and by Section~\ref{sec:Conc} where we discuss how this research can be taken further.

\section{The Conceptual Framework}\label{sec:conceptual_framework}
It was argued in \cite{Governatori2009157} that meta-rules can describe the dynamics of any normative institution (such as a legal system) where norms are formalised and can be used to establish conditions for the creation and modification of other rules or norms. In turn, proper rules precisely correspond to norms in normative systems. In particular, it was pointed out that meta-rules can be represented in the language of Defeasible Logic as follows:\footnote{From now on, as formalised in the language in the two following sections, literals will be denoted by Latin letters, standard rules by Greek letters, and meta-rules by Greek letters preceded by lower-case `\emph{m}'.} 
\begin{gather*}
m\alpha\colon\xseq{a}{n}\To (\beta\colon\xseq{b}{m}\To c),
\end{gather*}
precisely to grasp norm change mechanisms in the law. For instance, if the rule `$\beta\colon\xseq{b}{m}\To c$' does not exist in the theory at hand, then the successful application of $m\alpha$ leads to deriving such a rule, which amounts to enacting $\beta$ as a new norm in the legal system. Similarly, if $\beta$ already exists but has the form $\beta\colon b_1\To c$', then the successful application of $m\alpha$ corresponds to modifying $\beta$ from `$\beta\colon b_1\To c$' into `$\beta\colon\xseq{b}{m}\To c$'.

In addition,  with meta-rules we can admit the \emph{negation of rules}. If we are able to conclude that a (positive) rule holds, then it means that we can insert the rule (the content of the rule, with a specific name) into the system, and we can use the resulting rule to derive new conclusions. For a negated rule, the meaning is that it is not possible to obtain a rule with that specific content (this can be either formally prescribed for the whole rule or in a way that results irrespective of the name). 

In this paper we go further. We discuss the conceptual and logical meanings of modalising rules via meta-rules, i.e., what we mean when we establish the obligatoriness of enacting a certain rule. Put it in this way, making a rule obligatory amounts to firing a meta-rules like the following:
\begin{gather*}
m\zeta\colon \xseq{a}{n}\To_{\Obl} (\gamma\colon\xseq{b}{m}\To_{\Obl} c).
\end{gather*}
Rule $\gamma$ is a standard deontic rule as described in previous works \cite{GovernatoriOSRC16,GovernatoriORS13,GovernatoriR08}: if $\xseq{a}{n}$ are the case, then rule $\gamma$ allows us to derive that $c$ is obligatory. If a similar intuition is extended to meta-rules, then $m\zeta$ asserts that enacting rule `$\gamma\colon\xseq{b}{m}\To_{\Obl} c$' is obligatory.

We should now ask: For whom `$\gamma\colon\xseq{b}{m}\To_{\Obl} c$' is obligatory? There is no unique answer, since the question has different meanings, depending on which normative domain is considered. Suppose we work, again, in the legal domain. In that case, we can imagine that a legislative authority imposes the obligation to enact a certain norm over another authority, lower in the legislative hierarchy but competent in enacting such a norm. Concrete examples occur in the law, when, for instance, European authorities impose on member states the implementation of European directives or when Constitutional courts require national parliaments to amend, or to integrate, the legislative corpus. 

In this investigation, we aim to unravel some important aspects of meta-norms, looking closely at the \textit{mechanism} that allows meta-rules to develop rules or prevent their existence. In the current investigation, we are not concerned with the complex aspects underlying the \textit{goals} (and potentially the \textit{intentions}) of these meta-rules. This is a different issue, which has been partly explored in the recent literature \cite{Cristani201739} but it is not the focus of this study.

We remark that structures like $m\zeta$ do not lead to a meaningless iteration of deontic modalities, an issue explored in early deontic literature such as in \cite{Barcan:1966,Goble:1966}. First of all, the obligatory enactment of a prescriptive norm is not equivalent to a simple iteration of obligations. In addition, while it looks meaningless to have an expression like $\Obl \Obl c$ standing for ``It is obligatory that it is obligatory that everyone keeps their promises'', an expression like ``Parking on highways ought to be forbidden'' makes sense \cite{Barcan:1966}. The latter example suggests that a norm forbidding parking on highways is obligatory, thus assuming a conceptual distinction between norms, on the one side, and obligations (and permissions) on the other side \cite{alchourron71normative,makinson99on}.  Our general approach, hence, clearly distinguishes norms from obligations:  obligations are the effects (i.e., the conclusions) of the application of prescriptive norms. Under this reading, the application of rule $\gamma$ leads to the obligation $\Obl c$, while the application of meta-rule $m\zeta$ leads to a state where norm $\gamma$ ought to be the case.

What about the permission? The elusive character of permission also affects the case of permissive meta-rules. Consider the meta-rule
\begin{gather*}
m\xi\colon \xseq{a}{n}\To_{\Perm} (\gamma\colon\xseq{b}{m}\To_{\Obl} c).
\end{gather*}
Does the well-known distinction between \emph{weak}
and \emph{strong} permission \cite{vonwright:1963} apply to this case as well? In the standard scenario, the former type of permission corresponds to saying that something is allowed by a code precisely when that code does not prohibit it. This idea is preserved when meta-rules are considered. Indeed, one may simply argue that, if the legal system \emph{does not} support the derivation of 
\begin{gather*}
\Obl\neg(\gamma\colon\xseq{b}{m}\To_{\Obl} c)\\
\Obl(\varphi\colon\xseq{b}{m}\To_{\Obl} \neg c),
\end{gather*}
then one may also conclude that the following holds:
\begin{gather*}
\Perm (\gamma\colon\xseq{b}{m}\To_{\Obl} c).
\end{gather*}
For similar reasons, since weak permission is the dual of obligation -- i.e., $\Obl A =_{\mathit{def}} \neg \Obl \neg A$ -- imposing consistency means $\Obl A \to \Perm A$,  and it hence looks reasonable that from `$\Obl(\gamma\colon\xseq{b}{m}\To_{\Obl} c)$' we can  obtain `$\Perm (\gamma\colon\xseq{b}{m}\To_{\Obl} c)$'.

As is well-known, the concept of strong permission is more complicated, as it amounts to saying that some $a$ is permitted by a code iff the code explicitly states that $a$ is permitted. Various sub-types of permissions can be thus identified, such as the following ones \cite{makinson-torre:2003,boella-torre-icail:2003,stolpe-jal:2010,GovernatoriORS13}:
\begin{description}
    \item[Static permission:] $X$ is a static permission, wrt a normative system, when $X$ is derived from a strong permission, i.e., from an explicit permissive norm.

    \item[Dynamic permission:] $X$ is a dynamic permission, wrt a normative system, when it guides the legislator by describing the limits on what may be prohibited typically without violating static permissions in the system.
    
    \item[Exemption:] $X$ is an exemption, wrt a normative system, when it is an exception of a prohibition contained in the system.
\end{description}
It seems reasonable that all these types may be extended to the case of meta-rules. The first case simply amounts to when a rule is explicitly permitted in the theory or is derived from a permissive meta-rule (such as $m\xi$). The third case is when, for instance, we have in the system two rules, such as 
\begin{gather*}
\Perm (\gamma:\xseq{b}{m}\To_{\Obl} c)\\
\Obl (\varphi:\xseq{b}{m}\To_{\Obl} \neg c),
\end{gather*}
and we know that $\gamma$ is stronger than $\varphi$, or, at a meta-rule level, we have
\begin{gather*}
m\xi\colon \xseq{a}{n}\To_{\Perm} (\gamma\colon\xseq{b}{m}\To_{\Obl} c) \\
m\chi\colon \xseq{a}{n}\To_{\Obl} (\varphi\colon\xseq{b}{m}\To_{\Obl} \neg c) 
\end{gather*}
and, similarly, $m\xi$ is stronger than $m\chi$. The idea of dynamic permission consists in preventing the theory from deriving any incompatible deontic rule, for example, by setting that the meta-rule $m\xi$ is stronger than any other meta-rule supporting any conflicting rule. 

The discussion above shows how crucial it is to establish some logical properties of permitted rules and determine when modalised rules conflict. Accordingly, desirable basic properties are the following:
\begin{gather}
\Perm (\xseq{a}{n}\To_{\Obl} b) \vdash \neg\Obl(\xseq{a}{n}\To_{\Obl} \neg b) \label{eq:reviewer1}\\
\Perm (\xseq{a}{n}\To_{\Box} b) \vdash \neg\Obl \neg (\xseq{a}{n}\To_{\Box} b), \label{eq:reviewer2}
\end{gather} 
\sloppy where (\ref{eq:reviewer2}) is trivially desirable because, for a rule $\phi$, $\Perm \phi$ should imply $\neg \Obl \neg \phi$. Instead, if the rule in the scope of the permission is a permissive norm (i.e., `$\xseq{a}{n}\To_{\Perm} b$'), then  principle (\ref{eq:reviewer1}) does not hold in general\footnote{Although we may have concrete examples where it seems a bit odd that the legislator explicitly issues \emph{both} `$\xseq{a}{n}\To_{\Perm} b$' and `$\xseq{a}{n}\To_{\Perm} \neg b$', this is deontically possible: see below. Many thanks to a reviewer for commenting on this point.}. Suppose that the normative system (for instance, based on constitutional values) permits the enactment of a norm that allows for the temporary limitation of liberties due to public health reasons. If so, it would not be contradictory that the normative system prescribes that the lack of limitation is also permitted. The following rules are in fact not necessarily incompatible:
\begin{gather*}
\mathit{public\_health}\To_{\Perm} \mathit{limit\_liberties}\\
\mathit{public\_health}\To_{\Perm} \neg \mathit{limit\_liberties}
\end{gather*} 
which, essentially, amount to saying that, under certain conditions, $\Perm \mathit{limit\_liberties}$ and $\Perm \neg \mathit{limit\_liberties}$ are deontically compatible.

Other properties depend upon to what extent we assume that the legislator was rational in a subtler way. For instance, the following one
\begin{gather}
\Perm (\xseq{a}{n}\To_{\Obl} b) \vdash \neg\Perm (\xseq{a}{n}\To_{\Obl} \neg b) \label{eq:DPermRuleInternal}
\end{gather}
can be rational, since it does not make much sense for a rational legislator to permit that two deontically incompatible rules are the case. 

Let us discuss further this case. 
If we accept it, one may similarly argue that we should also adopt the following (where $b$ is a literal):
\begin{gather}
\Perm  b \vdash \neg\Perm \neg b \label{eq:D_Perm}
\end{gather}
which, however, cannot be accepted. Why? Because the deontic facultativeness of a certain $b$ precisely amounts to stating $\Perm  b \wedge \Perm \neg b$, so  (\ref{eq:D_Perm}) would make facultativeness inconsistent.

What about facultative norms? We may have two cases like:
\begin{gather}
\Perm (\xseq{a}{n}\To_{\Obl} b) \qquad \Perm \neg (\xseq{a}{n}\To_{\Obl}  b) \label{eq:FacultativeRule1} \\
\Perm (\xseq{a}{n}\To_{\Obl} b) \qquad \Perm (\xseq{a}{n}\To_{\Obl}  \neg b). \label{eq:FacultativeRule2}
\end{gather}
Clearly, (\ref{eq:FacultativeRule1}) and (\ref{eq:FacultativeRule2}) are deontically different. While the former states that 
`$\xseq{a}{n}\To_{\Obl} b$' is facultative, the latter states that two deontically incompatible and different rules are permitted. 

What is facultative according to (\ref{eq:FacultativeRule2})? Certainly we cannot say that  $\Obl b$ is facultative, since we would need to permit two rules, one supporting $\Obl b$ -- which is the case: `$\xseq{a}{n}\To_{\Obl} b$' -- and one supporting $\neg \Obl b$ -- which we do not have because `$\xseq{a}{n}\To_{\Obl}  \neg b$' rather supports $\Obl \neg b$. Principle (\ref{eq:FacultativeRule2}) licenses that both $\Obl b$ and $\Obl \neg b$ are permitted given `$\xseq{a}{n}$'. Is it factually possible? Yes, it is. Is it deontically admissible? Perhaps, it is. Is it always deontically rational? We sometimes have arguments to answer: No. 

Consider the idea of static permission. Assume we would like to avoid having two deontically incompatible rules in the system. Then, it would not be reasonable to explicitly permit that both rules are the case. As in sceptical defeasible reasoning, if we have arguments for deriving $b$ and $\neg b$, we refrain from concluding anything on the assumption that $b$ and $\neg b$ are in contradiction; we could reason similarly when two conflicting deontic rules like `$\xseq{a}{n}\To_{\Obl} b$' and `$\xseq{a}{n}\To_{\Obl} \neg b$' are taken into account.\footnote{One may argue that, if the conflict between `$\xseq{a}{n}\To_{\Obl} b$' and `$\xseq{a}{n}\To_{\Obl} \neg b$' cannot be solved, this would imply (but would not be equivalent in Defeasible Logic to) $b$ being weakly permitted. Therefore, we would make a weird use of static positive permissions to state that $b$ is weakly permitted.}  

Consider instead the idea of dynamic permission: the aim here is limiting the legislator in dynamically adding prohibitions. If so, we would have that
\begin{gather*}
\Perm (\xseq{a}{n}\To_{\Obl} b) \text{ prevents the derivation of }\xseq{a}{n}\To_{\Obl} \neg b\\
\Perm (\xseq{a}{n}\To_{\Obl} \neg b) \text{ prevents the derivation of }\xseq{a}{n}\To_{\Obl} b.
\end{gather*}
If we have it (i.e., we prevent both derivations), it means that we want the system to be deontically indifferent with respect to $b$ whenever `$\xseq{a}{n}$' are the case. Suppose, however, that we also have other two norms like the following:
\begin{gather*}
\xseq{a}{n}\To_{\Obl} c \\
\Obl c \To_{\Obl} b.
\end{gather*}
Despite the fact that we prevent the derivation of both `$\xseq{a}{n}\To_{\Obl} \neg b$' and `$\xseq{a}{n}\To_{\Obl} b$', can we still say that the system is deontically indifferent with respect to $b$ given `$\xseq{a}{n}$'?

Accordingly, under the reading above, one may prudentially establish that the following meta-rules are somehow incompatible:
\begin{gather}
m\xi\colon \xseq{a}{n}\To_{\Perm} (\gamma\colon\xseq{b}{m}\To_{\Obl} c) \label{eq:1} \\
m\chi\colon \xseq{a}{n}\To_{\Perm} (\varphi\colon\xseq{b}{m}\To_{\Obl} \neg c) \label{eq:-1}
\end{gather}
In the above discussion, we assume that $\Perm$ is not the dual of $\Obl$, but rather as another $\Box$-operator. A logic for meta-rules assuming that \ref{eq:1} and \ref{eq:-1} \emph{are in conflict} is named \textbf{cautious}.

On the contrary, if we believe that the permission, as applied to rules, behaves exactly as when literals are modalised, then $m\alpha$ and $m\zeta$ are not in conflict, and this holds precisely because
\begin{gather}
\xi\colon \xseq{a}{n}\To_{\Perm} b\\
\psi\colon \xseq{a}{n}\To_{\Perm} \neg b 
\end{gather}
are likewise compatible in virtue of the intuition that $\Perm b$ and $\Perm \neg b$ are consistent. A logic for meta-rules assuming that \ref{eq:1} and \ref{eq:-1} \emph{are not in conflict} is named \textbf{simple}.

Finally, let us comment on applying the $\otimes$ operator \cite{GovernatoriORS13} to rules. In the standard reading of this operator, a rule like `$\xseq{a}{n}\To_{\Obl} b\otimes c$' means that if `$\xseq{a}{n}$' are the case, then $b$ is obligatory; on the contrary, if the obligation $b$ is not fulfilled, then the obligation $c$ is activated, and becomes in force until it is satisfied or violated. Since we argued that a legislator $L_1$ can impose on another legislator $L_2$ to enact a norm $\mu$, we can imagine a scenario where $L_2$ violates $\Obl\mu$. Still, we can also imagine that $L_1$ has considered a sanction as the result of such a violation, or rather that another normative solution is advanced by $L_2$. Accordingly, expressions such as
\begin{gather}
m\alpha\colon \xseq{a}{n}\To_{\Obl} (\gamma\colon\xseq{b}{m}\To_{\Obl} c)\otimes (\zeta\colon \xseq{b}{m} \To_{\Obl} d)
\end{gather}
are admissible. 
The $\otimes$ operator can also be seen as a preference operator, where the first element is the most preferred, and the last is the least of the acceptable options \cite{GovernatoriORS13,GovernatoriOSRC16}. According to this reading, meta-rule $m\alpha$ establishes that the underlying normative systems should introduce rule $\beta$ (if such a rule is not already in the system). Alternatively, a less preferable but still acceptable outcome is to impose $\zeta$. Generally, in real-life normative systems, the idea is to use this kind of structure to prescribe more and more stringent norms/policies.

%% file: Methodology.tex
\section{Defeasible Deontic Logic}\label{sec:Method}

Defeasible Logic (DL) \cite{nute,DBLP:journals/tocl/AntoniouBGM01} is a simple, flexible, and efficient rule-based non-monotonic formalism. Its strength lies in its constructive proof theory, allowing it to draw meaningful conclusions from (potentially) conflicting and incomplete knowledge base. In non-monotonic systems, more accurate conclusions can be obtained when more pieces of information become available.

Many variants of DL have been proposed for the logical modelling  of different application areas, specifically agents \cite{kravari2015,GovernatoriOSRC16,prima07:contextual}, legal reasoning \cite{Governatori2009157,Cristani201739}, and workflows from a business process compliance perspective \cite{DBLP:conf/ruleml/GovernatoriOSC11,Olivieri2015603}. 

In this research we focus on the Defeasible Deontic Logic (DDL) framework \cite{GovernatoriORS13} that allows us to determine what prescriptive behaviours are in force in a given situation. 

We start by defining the language of a defeasible deontic theory.

Let $\PROP$ be a set of propositional atoms, and $\LABELS$ be a set of arbitrary labels (the names of the rules). We use lower-case Roman letters to denote literals and lower-case Greek letters to denote rules.

Accordingly, $\PLit = \PROP \cup \set{\neg l\, |\, l \in \PROP}$ is the set of \emph{plain literals}. The set of \emph{deontic literals} $\ModLit = \set{\Box l, \neg \Box l\ |\ l \in \PLit \wedge \Box \in \set{\Obl, \Perm}}$. Finally, the set of \emph{literals} is $\LIT = \PLit \cup \ModLit$. The \emph{complement} of a literal $l$ is denoted by $\non l$: if $l$ is a positive literal $p$ then $\non l$ is $\neg p$, and if $l$ is a negative literal $\neg p$ then $\non l$ is $p$. Note that we will not have specific rules nor modality for prohibitions, as we will treat them according to the standard duality that something is forbidden iff the opposite is obligatory (i.e., $\Obl\neg p$).

\begin{definition}[Defeasible Deontic Theory]\label{def:DeonticTheory}
	A \emph{defeasible deontic theory} $D$ is a tuple $(\FACTS, R, >)$, where $\FACTS$ is the set of facts, $R$ is the set of rules, and $>$ is a binary relation over $R$ (called superiority relation).
\end{definition}	
Specifically, the set of facts $\FACTS \subseteq \PLit$ denotes simple pieces of information that are always considered to be true, like ``Sylvester is a cat'', formally $cat(Sylvester)$. 
In this paper, we subscribe to the distinction between the notions of obligations and permissions, and that of norms, where the norms in the system determine the obligations and permissions in force in a normative system. A Defeasible Deontic Theory is meant to represent a normative system, where the rules encode the norms of the systems, and the set of facts corresponds to a case. As we will see below, the rules are used to conclude the institutional facts, obligations and permissions that hold in a case. 
Accordingly,  we do not admit obligations and permissions as facts of the theory.

The set of rules $R$ contains three \emph{types} of rules: \emph{strict rules}, \emph{defeasible rules}, and \emph{defeaters}. Rules are also of two \emph{kinds}: 
\begin{itemize}
	\item \emph{Constitutive rules} (non-deontic rules) $R^\Cost$ model constitutive statements (count-as rules);
	\item \emph{Deontic rules} to model prescriptive behaviours, which are either \emph{obligation rules} $R^\Obl$ which determine when and which obligations are in force, or \emph{permission rules} which represent \emph{strong} (or \emph{explicit}) permissions $R^\Perm$.
\end{itemize}
Lastly, ${>} \subseteq R \times R$ is the \emph{superiority} (or \emph{preference}) relation, which is used to solve conflicts in case of potentially conflicting information.

A theory is \emph{finite} if the set of facts and rules are so.

A strict (constitutive) rule is a rule in the classical sense: whenever the premises are indisputable, so is the conclusion. The statement ``All computing scientists are humans'' is hence formulated through the strict rule\footnote{Here, we introduce informally the symbols to represent different types of rules, which are formally defined below in Definition~\ref{def:Rule}, where $\TO$ denotes a strict rule, $\To$ for a defeasible rule, and $\defeater$ for a defeater.}
\[
\mathit{CScientist}(X) \TO_\Cost \mathit{human}(X),
\]
as there is no exception to it%
\footnote{Like in \cite{DBLP:journals/tocl/AntoniouBGM01}, we consider only a propositional version of this logic, and we do not take into account function symbols. Every expression with variables represents the finite set of its variable-free instances.}.

On the other hand, defeasible rules are to conclude statements that can be defeated by contrary evidence. In contrast, defeaters are special rules whose only purpose is to prevent the derivation of the opposite conclusion. Accordingly, we can represent the statement ``Computing scientists travel to the city of the conference'' through a defeasible rule, whereas ``During pandemic travels might be prohibited'' through a defeater, like
\[
\mathit{CScientist}, \mathit{PaperAccepted} \To_\Cost \mathit{TravelConference}
\]
\[
\mathit{Pandemic} \defeater_\Cost \neg \mathit{TravelConference}.
\]
On the other hand, a prescriptive behaviour like ``At traffic lights it is forbidden to perform a U-turn unless there is a `U-turn Permitted' sign'' can be formalised via the general obligation rule
\[
\mathit{AtTrafficLight} \To_\Obl \neg \mathit{UTurn}
\]
and the exception through the permissive rule
\[
\mathit{UTurnSign} \To_\Perm \mathit{UTurn}.
\]
While \cite{GovernatoriORS13} discusses how to integrate strong and weak permission in 
defeasible deontic logic, in this paper, we restrict our attention to the notion of strong permission, namely, when permissions are explicitly stated using permissive rules, i.e., rules whose conclusion is to be asserted as a permission.  

Following the ideas of \cite{ajl:ctd}, obligation rules gain more expressiveness with the \emph{compensation operator} $\otimes$ for obligation rules, which is to model reparative chains of obligations. Intuitively, $a \otimes b$ means that $a$ is the primary obligation, but if for some reason we fail to obtain, to comply with, $a$ (by either not being able to prove $a$, or by proving $\non a$) then $b$ becomes the new obligation in force. This operator is used to build chains of preferences, called $\otimes$-expressions.

The formation rules for $\otimes$-expressions are: (i) every plain literal is an $\otimes$-expression, (ii) if $A$ is an $\otimes$-expression and $b$ is a plain literal then $A \otimes b$ is an $\otimes$-expression \cite{GovernatoriORS13}. 
 
 In general an $\otimes$-expression has the form `$c_1 \otimes c_2 \otimes \dots \otimes c_m$', and it appears as consequent of a rule `$A(\alpha) \hookrightarrow_\Obl C(\alpha)$' where $C(\alpha)=c_1 \otimes c_2 \otimes \dots \otimes c_m$; the meaning opf the $\otimes$-expression is: if the rule is allowed to draw its conclusion, then $c_1$ is the obligation in force, and only when $c_1$ is violated then $c_2$ becomes the new in force obligation, and so on for the rest of the elements in the chain. In this setting, $c_m$ stands for the last chance to comply with the prescriptive behaviour enforced by $\alpha$, and in case $c_m$ is violated as well, then we will result in a non-compliant situation.

For instance, the previous prohibition to perform a U-turn can foresee a compensatory fine, like
\[
\mathit{AtTrafficLight} \To_\Obl \neg \mathit{UTurn} \otimes \mathit{PayFine}.
\]
that has to be paid in case someone does perform an illegal U-turn.

It is worth noticing that we admit $\otimes$-expressions with only one element. The intuition, in this case, is that the obligatory condition does not admit compensatory measures or, in other words, that it is impossible to recover from its violation.

In this paper, we focus exclusively on the defeasible part of the logic ignoring the monotonic component given by the strict rules; consequently, we limit the language to the cases where the rules are either defeasible or defeaters. From a practical point of view, the restriction does not effectively limit the expressive power of the logic: a defeasible rule where there are no rules for the opposite conclusion, or where all rules for the opposite conclusion are weaker than the given defeasible rules, effectively behaves like a strict rule. Formally a rule is defined as below.

\begin{definition}[Rule]\label{def:Rule}
A \emph{rule} is an expression of the form $\alpha\colon A(\alpha) \hookrightarrow_\Box C(\alpha)$, where
	\begin{enumerate}
		\item $\alpha \in \LABELS$ is the unique name of the rule;
		\item $A(\alpha) \subseteq \LIT$ is the set of antecedents;
		\item An arrow ${\hookrightarrow} \in \set{\To, \defeater}$ denoting, respectively, defeasible rules, and defeaters;
		\item $\Box \in \{\Cost, \Obl, \Perm \}$;
		\item its consequent $C(\alpha)$, which is either
		\begin{enumerate}
			\item a single plain literal $l \in \PLit$, if either (i) ${\hookrightarrow} \equiv {\defeater}$ or (ii) $\Box \in \set{\Cost, \Perm}$, or
			\item an $\otimes$-expression, if $\Box \equiv \Obl$.
		\end{enumerate}	
	\end{enumerate}
\end{definition}
If $\Box = \Cost$ then the rule is used to derive non-deontic literals (constitutive statements), whilst if $\Box$ is $\Obl$ or $\Perm$ then the rule is used to derive deontic conclusions (prescriptive statements). The conclusion $C(\alpha)$ is, as before, a single literal in case $\Box = \{\Cost, \Perm\}$; in case $\Box = \Obl$, then the conclusion is an $\otimes$-expression. 
Note that $\otimes$-expressions can only occur in prescriptive rules though we do not admit them on defeaters (Condition 5.(a).i), see \cite{GovernatoriORS13} for a detailed explanation.

We use some abbreviations on sets of rules. The set of defeasible rules in $R$ is $R_\To$, the set of defeaters is $R_\dft$. $R^\Box[l]$ is the rule set appearing in $R$ with head $l$ and modality $\Box$, while $R^\Obl[l,i]$ denotes the set of obligation rules where $l$ is the $i$-th element in the $\otimes$-expression.  Given that the consequent of a rule is either a single literal or an $\otimes$-expression (that, due to the associative property, can be understood as a sequence of elements, and then as an ordered set), in what follows we are going to abuse the notation and use $l\in C(\alpha)$.  $R^\Box$ is the set of rules $\alpha\colon A(\alpha)\hookrightarrow_{\Box} C(\alpha)$ such that $\alpha$ \emph{appears in} $R$. For a theory as determined by Definitions~\ref{def:DeonticTheory} and \ref{def:Rule}, $\alpha$ appears in $R$ means that $\alpha\in R$; thus $R^\Perm$ is the set of permissive rules appearing in $R$. We use $R^\Diamond$ and $R^\Diamond[l]$ as shorthands for $R^\Obl\cup R^\Perm$ and $R^\Obl[l]\cup R^\Perm[l]$ respectively. The abbreviations can be combined. Finally, a literal $l$ appears in a theory $D$, if there is a rule $\alpha\in R$ such that $l\in A(\alpha)\cup C(\alpha)$.

\begin{definition}[Tagged modal formula]\label{def:TagFormula}
    A \emph{tagged modal formula} is an expression of the form
$\pm\partial_\Box l$, with the following meanings
	\begin{itemize}
		\item $+\partial_\Box l$: $l$ is \emph{defeasibly provable} (or simply provable) with mode $\Box$, 
		\item $-\partial_\Box l$:  $l$ is \emph{defeasibly refuted} (or simply refuted) with mode $\Box$;
	\end{itemize}
\end{definition}
Accordingly, the meaning of $+\partial_\Obl p$ is that
$p$ is provable as an obligation, and $-\partial_\Perm\neg p$ is that 
we have a refutation for the permisison of $\neg p$. Similarly, for the other combinations.

As we will shortly see (Definitions~\ref{def:StandardApplicability} and \ref{def:StandardDiscardability}), one of the key ideas of Defeasible Deontic Logic is that we use tagged modal formulas to determine what formulas are (defeasibly) provable or rejected given a theory and a set of facts (used as input for the theory). Therefore, when we have asserted the tagged modal formula $+\partial_\Obl l$ in a derivation (see Definition~\ref{def:Proof} below), we can conclude that the obligation of $l$ ($\Obl l$) follows from the rules and the facts and that we used a prescriptive rule to derive $l$; similarly for permission (using a permissive rule). However, the $\Cost$ modality is silent, meaning that we do not put the literal in the scope of the $\Cost$ modal operator, thus for $+\partial_\Cost l$, the derivation simply asserts that  $l$ holds (and not that $\Cost l$ holds, even if the two have the same meaning).  For the negative cases (i.e., $-\partial_\Box l$), the interpretation is that it is not possible to derive $l$ with a given mode. Accordingly, we read $-\partial_\Obl l$ as it is impossible to derive $l$ as an obligation. For $\Box\in\set{\Obl,\Perm}$ we are allowed to infer $\neg\Box l$, giving a constructive interpretation of the deontic modal operators. Notice that this is not the case for $\Cost$, where we cannot assert that $\non l$ holds (this would require $+\partial_\Cost\non l$); in the logic, failing to prove $l$ does not equate to proving $\neg l$.

We will use the term \emph{conclusions} and tagged modal formulas interchangeably. 
 
The definition of proof is also the standard in DDL. 

\begin{definition}[Proof]\label{def:Proof}
	Given a defeasible deontic theory $D$, a proof $P$ of length $m$ in $D$ is a finite sequence $P(1), P(2),\dots,P(m)$ of tagged modal formulas, where the proof conditions defined in the rest of this paper hold. 
\end{definition}
Hereafter, $P(1..n)$ denotes the first $n$ steps of  $P$, and we also use the notational convention $D\vdash \pm\partial_\Box l$, meaning that there is a proof $P$ for $\pm\partial_\Box l$ in $D$.	

Core notions in DL are that of \emph{applicability/discardability}. As knowledge in a defeasible theory is circumstantial, given a defeasible rule like `$\alpha\colon a, b \To_\Box c$', there are four possible scenarios: the theory defeasibly proves both $a$ and $b$, the theory proves neither, the theory proves one but not the other. Naturally, only in the first case, where both $a$ and $b$ are proved, we can use $\alpha$ to \emph{support/try to conclude} $\Box c$. Briefly, we say that a rule is \emph{applicable} when every antecedent's literal has been proved at a previous derivation step. Symmetrically, a rule is \emph{discarded} when one of such literals has been previously refuted. Formally:

\begin{definition}[Applicability]\label{def:StandardApplicability}
	Assume a deontic defeasible theory $D = (\FACTS, R, >)$.  We say that rule $\alpha \in R^\Cost \cup R^\Perm$ is \emph{applicable} at $P(n+1)$, iff for all $a \in A(\alpha)$
	\begin{enumerate}
		\item\label{item:l} if $a \in \PLit$, then $+\partial_\Cost a \in P(1..n)$,
		\item if $a = \Box q$, then $+\partial_\Box q \in P(1..n)$, with $\Box \in\set{\Obl, \Perm}$,
		\item\label{item:Diamondl} if $a = \neg \Box q$, then $-\partial_\Box q \in P(1..n)$, with $\Box \in \set{\Obl, \Perm}$.
	\end{enumerate}

	We say that rule $\alpha \in R^\Obl$ is \emph{applicable at index} $i$ \emph{and} $P(n+1)$ iff Conditions~\ref{item:l}--\ref{item:Diamondl} above hold and
	\begin{enumerate}[resume]
		\item\label{item:cj} $\forall c_j \in C(\alpha),\, j < i$, then $+\partial_\Obl c_j \in P(1..n)$ and $+\partial_\Cost \non c_j \in P(1..n)$\footnote{\label{foot:violation}As discussed above, we are allowed to move to the next element of an $\otimes$-expression when the current element is violated. To have a violation, we need (i) the obligation to be in force, and (ii) that its content does not hold. $+\partial_\Obl c_i$ indicates that the obligation is in force.  For the second part we have two options. The former, $+\partial_\Cost\non c_i$ means that we have ``evidence'' that the opposite of the content of the obligation holds.  The  latter  would be to have $-\partial_\Cost c_j \in P(1..n)$ corresponding to the intuition that we failed to provide evidence that the obligation has been satisfied.  It is worth noting that the former option implies the latter one. For a deeper discussion on the issue, see \cite{jurix2015burden}.}. 
	\end{enumerate}
	
\end{definition}

\begin{definition}[Discardability]\label{def:StandardDiscardability}
	Assume a deontic defeasible theory $D$, with $D = (\FACTS, R, >)$.  We say that rule $\alpha \in R^\Cost \cup R^\Perm$ is \emph{discarded} at $P(n+1)$, iff there exists $a \in A(\alpha)$ such that
	\begin{enumerate}
		\item\label{item:notl} if $a \in \PLit$, then $-\partial_\Cost l \in P(1..n)$, or
		\item if $a = \Box q$, then $-\partial_\Box q \in P(1..n)$, with $\Box \in \set{\Obl, \Perm}$, or
		\item\label{item:notDiamondl} if $a = \neg \Box q$, then $+\partial_\Box q \in P(1..n)$, with $\Box \in \set{\Obl, \Perm}$.
	\end{enumerate}

	We say that rule $\alpha \in R^\Obl$ is \emph{discarded at index } $i$ \emph{and} $P(n+1)$ iff either at least one of the Conditions~\ref{item:notl}--\ref{item:notDiamondl} above hold, or 
	\begin{enumerate}[resume]
		\item\label{item:notcj} $\exists c_j \in C(\alpha),\, j < i$ such that $-\partial_\Obl c_j \in P(1..n)$, or $-\partial_\Cost \non c_j \in P(1..n)$. 
	\end{enumerate}
	
\end{definition}
Note that discardability is obtained by applying the principle of \emph{strong negation} to the definition of applicability. The strong negation principle applies the function that simplifies a formula by moving all negations to an innermost position in the resulting formula, replacing the positive tags with the respective negative tags, and the other way around see \cite{DBLP:journals/igpl/GovernatoriPRS09}.
Positive proof tags ensure that there are effective decidable procedures to build proofs; the strong negation principle guarantees that the negative conditions provide a constructive and exhaustive method to verify that a derivation of the given conclusion is not possible. Accordingly, condition 3 of Definition~\ref{def:StandardApplicability} allows us to state that
$\neg\Box p$ holds when we have a (constructive) failure to prove $p$ with mode $\Box$ (for obligation or permission), thus it corresponds to a constructive version of negation as failure.  

We are finally ready to formalise the proof conditions, which are the standard conditions of DDL \cite{GovernatoriORS13}. We start with positive proof conditions for constitutive statements. In the following, we shall omit the explanations for negative proof conditions, when trivial, reminding the reader that they are obtained through the application of the strong negation principle to the positive counterparts.

\begin{definition}[Constitutive Proof Conditions]\label{def:StandardCostProof}
	
\begin{tabbing}
  $+\partial_\Cost l$: \=If $P(n+1)=+\partial_\Cost l$ then\+\\
  (1) \= $l \in \FACTS$, or\\
  (2) \> (1) \= $\non l \not\in \FACTS$, and\\
      \> (2) \= $\exists \beta \in R^\Cost_\To[l]$ s.t. $\beta$ is applicable, and\\
      \> (3) \=$\forall \gamma\in R^\Cost[\non l]$ either\\
        \>\>(1) \=$\gamma$ is discarded, or \\
        \>\>(2)  $\exists \zeta \in R^\Cost[l]$ s.t. \\
            \>\>\> (1) $\zeta$ is applicable and \\
            \>\>\> (2) $\zeta > \gamma$.
\end{tabbing}  

\begin{tabbing}
  $-\partial_\Cost l$: \=If $P(n+1)=-\partial_\Cost l$ then\+\\
  (1) \= $l \not\in \FACTS$ and either\\
  (2) \= (1) \= $\non l \in \FACTS$, or\\
  \> (2) $\forall \beta \in R^\Cost_\To[l]$, either $\beta$ is discarded, or\\
  \> (3) \=$\exists \gamma\in R^\Cost[\non l]$ such that\+\+\\
        \=(1) $\gamma$ is applicable, and\\
        \=(2) $\forall \zeta \in R^\Cost[l]$, either $\zeta$ is discarded, or $\zeta \not > \gamma$.
\end{tabbing}

\end{definition}
A literal is defeasibly proved if: it is a fact, or there exists an applicable, defeasible rule supporting it (such a rule cannot be a defeater), and all opposite rules are either discarded or defeated.  To prove a conclusion, not all the work has to be done by a stand-alone (applicable) rule (the rule witnessing condition (2.2): all the applicable rules for the same conclusion (may) contribute to defeating applicable rules for the opposite conclusion. Note that both $\gamma$ as well as $\zeta$ may be defeaters.

\begin{example}\label{ex:Standard}
	Let $D = (F = \set{a, b, c, d, e}, R, {>}  = \set{(\alpha, \varphi), (\beta, \psi)})$ be a theory such that 
\begin{align*}
    R = \{&\alpha\colon a \To_\Cost l &&\beta\colon b \To_\Cost l && \gamma\colon c \To_\Cost l\\ 
    &\varphi\colon d \To_\Cost \neg l && \psi\colon e \To_\Cost \neg l && \chi\colon g \To_\Cost \neg l \}.	 
\end{align*}
\end{example}
Here, $D \vdash +\partial_\Cost f_i$, for each $f_i \in \FACTS$ and, by Condition (1) of $+\partial$. Therefore, all rules but $\chi$ (which is discarded) are applicable: $\chi$ is indeed discarded since no rule has $g$ as consequent nor is a fact. The team defeat supporting $l$ is made by $\alpha$, $\beta$ and $\gamma$, whereas the team defeat supporting $\neg l$ is made by $\varphi$ and $\psi$. Given that $\alpha$ defeats $\varphi$ and $\beta$ defeats $\psi$, then we conclude that $D\vdash +\partial_\Cost l$. Note that, despise being applicable, $\gamma$ does not  effectively contribute in proving $+\partial_\Cost l$, i.e. $D$ without $\gamma$ would still prove $+\partial_\Cost l$.

Suppose to change $D$ such that both $\alpha$ and $\beta$ are defeaters. Even if $\gamma$ defeats neither $\varphi$ nor $\psi$, $\gamma$ is now needed to prove $+\partial l$ as Condition (2.2) requires that at least one applicable rule must be a defeasible rule.  
Below we present the proof conditions for obligations.

\begin{definition}[Obligation Proof Conditions]\label{def:StandardOblProof}

\begin{tabbing}
  $+\partial_\Obl l$: \=If $P(n+1)=+\partial_\Obl l$ then\+\\
  	$\exists \beta \in R^\Obl_\To[l,i]$ s.t.\\
  	(1) \=$\beta$ is applicable at index $i$ and\\
  	(2) $\forall \gamma \in R^\Obl[\non l, j] \cup R^\Perm[\non l]$ either \+\\
  	   (1) \= $\gamma$ is discarded (at index $j$), or\\
  	   (2) \= $\exists \zeta \in R^\Obl[l, k]$ s.t.\+\\ 
  	   	\= (1) $\zeta$ is applicable at index $k$ and\\ 
  	   	\= (2) $\zeta >  \gamma$.
\end{tabbing} 

\begin{tabbing}
  $-\partial_\Obl l$: \=If $P(n+1)=-\partial_\Obl l$ then\+\\
  	$\forall \beta \in R_\To^\Obl[l,i]$ either\\
  	(1) $\beta$ is discarded at index $i$, or\\
  	(2) \= $\exists \gamma \in R^\Obl[\non l, j] \cup R^\Perm[\non l]$ s.t. \+\\
  	   (1) \= $\gamma$ is applicable (at index $j$), and\\
  	   (2) \= $\forall \zeta \in R^\Obl[l, k]$ either\+\\ 
  	   	\= (1) $\zeta$ is discarded at index $k$, or\\
  	   	\= (2) $\zeta \not >  \gamma$.
\end{tabbing} 

\end{definition}
Note that: (i) in Condition (2) $\gamma$ can be a permission rule as explicit, opposite permissions represent exceptions to obligations, whereas $\zeta$ (Condition 2.2) must be an obligation rule as a permission rule cannot reinstate an obligation, and that (ii) $l$ may appear at different positions (indices $i, j,$ and $k$) within the three $\otimes$-chains.  The example below supports the intuition behind the restriction to obligation rules in Conditions (2.2).

\begin{example}
Suppose that the medical guidelines of a hospital forbid the use of opioids to sedate patients with an addiction history. However, at the same time, the guidelines mandate the same kind of drugs for terminal patients with cancer. However, physicians are permitted to refuse to treat patients with opioids based on moral ground objections. 
\begin{gather*}
 \alpha\colon \mathit{AddictionHistory} \To_\Obl \neg\mathit{Opioids}\\
 \beta\colon \mathit{TerminalCancer} \To_\Obl \mathit{Opioids}\\
 \gamma\colon \mathit{MoralGround} \To_\Perm \neg\mathit{Opioids}
\end{gather*}
where $\gamma >\beta$ and $\beta >\alpha$.  Is it forbidden to use opioids for a terminally ill patient with an addiction history where moral ground objections apply? Here, rule $\gamma$ establishes an exemption from the obligation to prescribe opioids; however, the opposite course of action is asmissible, and the use of opioids for terminally ill cancer patients appears to be not forbidden (with or without moral ground objections). 

\end{example}
 
\noindent Below, we introduce the proof conditions for permissions.

\begin{definition}[Permission Proof Conditions]\label{def:StandardPermProof}

\begin{tabbing}
  $+\partial_\Perm l$: \=If $P(n+1)=+\partial_\Perm l$ then\+\\
  	(1) \= $+\partial_\Obl l \in P(1..n)$, or\\
  	(2) \= $\exists \beta \in R_\To^\Perm[l]$ s.t.\+\\
  	    (1) $\beta$ is applicable and\\
  	    (2) \= $\forall \gamma \in R^\Obl[\non l, j]$ either \+\\
  		    (1) \=$\gamma$ is discarded at index $j$, or\\
            (2) \=$\exists \zeta \in R^\Perm[l] \cup R^\Obl[l,k]$ s.t.\+\\
                (1) \= $\zeta$ is applicable (at index $k$) and\\
                (2) $\zeta >  \beta$.
\end{tabbing}

\begin{tabbing}
  $-\partial_\Perm l$: \=If $P(n+1)=-\partial_\Perm l$ then\+\\
  	(1) \= $-\partial_\Obl l \in P(1..n)$, and\\
  	(2) \= $\forall \beta \in R_\To^\Perm[l]$ either\+\\ 
  	    (1) $\beta$ is discarded or\\
  	    (2) \= $\exists \gamma \in R^\Obl[\non l, j]$ s.t. \+\\
  		    (1) \=$\gamma$ is applicable at index $j$ and\\
            (2) \=$\forall \zeta \in R^\Perm[l]  \cup R^\Obl[l,k]$ either\+\\
                (1) \= $\zeta$ is discarded (at index $k$), or\\
                (2) $\zeta \not >  \beta$.
\end{tabbing} 

\end{definition}
Condition (1) allows us to derive a permission from the corresponding obligation. Thus it corresponds to the $\Obl a\to\Perm a$ axiom of Deontic Logic. Condition (2.2) considers as possible counter-arguments \emph{only} obligation rules in situations where both $\Perm l$ and $\Perm \neg l$ hold are allowed.
We refer the readers interested in a deeper discussion on how to model permissions and obligations in DDL to \cite{GovernatoriORS13}.

Hereafter, whenever the applicability conditions of a given rule are not relevant for the example, we will set the corresponding set of antecedents to empty as such rules are vacuously applicable.

\begin{example}\label{ex:StandardDeontic}
	Assume the theory of Example~\ref{ex:Standard}, where we extend the rule set and the superiority relation as follows
\begin{align*}
    R &\cup \set{\zeta\colon \emptyset \To_\Obl \neg l \otimes p\quad \eta\colon \emptyset \To_\Perm l \quad \nu\colon \neg\Obl l \To_\Cost q }\\
    >  &\cup \set{(\zeta, \eta)}.
\end{align*}

\end{example}
Since $\zeta > \eta$, we conclude $D\vdash +\partial_\Obl \neg l$, $D\vdash -\partial_\Perm l$ due to Condition (3) of $-\partial_\Perm$, and $D\vdash -\partial_\Obl l$ due Condition (1) of $-\partial_\Obl$ as there are no obligation rules supporting $l$. Condition (3) of Definition~\ref{def:StandardApplicability} is satisfied, which makes $\nu$ applicable and hence $D\vdash +\partial_\Cost q$.

Given that, from Example~\ref{ex:Standard}, $D\vdash +\partial_\Cost l$, Condition (4) of Definition~\ref{def:StandardApplicability} is true which makes $\zeta$ applicable at index 2 for $p$: since the are no deontic rules supporting either $\Obl\neg p$ or $\Perm\neg p$, we conclude that $D\vdash +\partial_\Obl p$.

The set of positive and negative conclusions of a theory is called \emph{extension}. The extension of a theory is computed based on the literals that appear in it; more precisely, the literals in the Herbrand Base of the theory $\mathit{HB}(D)=\set{l,\non l \in\PLit|\,l \text{ appears in }D}$.

\begin{definition}[Extension]\label{def:StandardExtension}
	Given a defeasible deontic theory $D$, we define the \emph{extension} of $D$ as 
	\[E(D) = (+\partial_\Cost, -\partial_\Cost,+\partial_\Obl, -\partial_\Obl,+\partial_\Perm, -\partial_\Perm),\] 
	where $\pm\partial_\Box = \set{l \in \mathit{HB}(D) |\, D\vdash \pm\partial_\Box l}$, with $\Box \in \set{\Cost, \Obl, \Perm}$.
\end{definition}
In Example~\ref{ex:StandardDeontic}, $E(D)$ consists of the following sets:
\begin{align*}
+\partial_\Cost&=\set{l, q}\cup \FACTS,   
&-\partial_\Cost &= \set{\neg l, \neg q}\cup\set{\neg f_i|f_i\in\FACTS},\\
+\partial_\Obl &= \set{\neg l, p}, 
&-\partial_\Obl &= \set{l, q, \neg q}
	\cup \FACTS\cup\set{\neg f_i|f_i\in\FACTS}
\\ 
+\partial_\Perm&= \set{\neg l, p}, 
&-\partial_\Perm &= \set{l, \neg p, q, \neg q}\cup \FACTS\cup\set{\neg f_i|f_i\in\FACTS}.
\end{align*}

%% file: Logic.tex
Some of the DL application fields we alluded to in the previous section require modelling contexts and using rules in the scope of other rules. Accordingly, extensions of the logic have been developed to capture such features by adopting meta-rules.  However, the work on meta-rules in Defeasible Logic focused on defining the extensions of the logic, specifically the proof-theoretic features, neglecting to investigate the computational aspects. 

A recent work \cite{Olivieri202169} provides a first (to the best of our knowledge) complete logical framework that encodes meta-rules within a non-monotonic logic.
This paper generalises the model by:
\begin{itemize}
    \item Adding deontic operators for obligation ($\Obl$) and permission ($\Perm$);
    \item Adding the compensation operator $\otimes$;
    \item Applying deontic and compensation operators both to literals and to rules;
    \item Deeply investing situations where rules would provide conflicting information wrt other rules.
\end{itemize}
To this end, we first discuss the types of meta-rules conflicts we are going to model in Section~\ref{sub:conflicts}. Then, in Section~\ref{sec:commonV1V2} we enrich the language  with meta-rules extending the definitions provided in Section~\ref{sec:Method}. In addition, we provide the formal definitions for rules and meta-rules conflicts. Specifically, we give two definitions of conflicting rules; each version gives rise to a variant of Defeasible Deontic Logic with meta-rules. 
Sections~\ref{sec:Variante1} and \ref{sec:Variante2} are dedicated to proof conditions for the two variants. Finally, Section~\ref{subsec:Properties} provides some theoretical results for the two variants.   

\subsection{Conflicting Rules}
\label{sub:conflicts}
The richness of this extension makes the logic very expressive and leads to examining a potentially large variety of conflicts between meta-rules. Consider the following meta-rules:
\begin{gather*}
   \zeta\colon A(\zeta) \To_\Box (\alpha)\\
\nu\colon A(\nu) \To_\Box   (\beta).
\end{gather*}
On the one hand, we can simply extend to rules the intuition covering literals: as literal $s$ conflicts with another literal $l$ whenever $s$ is the complement of $l$ (i.e., $s=\non l$), rule $\nu$ conflicts with rule $\zeta$ whenever $\beta=\non \alpha$ (thus, when $\alpha =\neg \beta$ or $\beta= \neg \alpha$). 

What does it mean to apply $\neg$ to a rule? In our context -- normative reasoning -- if a rule $\alpha$ corresponds to a norm, then $\neg \alpha$ means that such a norm does not exist in the normative system or is removed.

\begin{example}\label{example:standard}
Consider this case from Italian Law. The Parliament issues the following Legislative Act (n. 124, 23 July 2008):  
\begin{quote}[{\bf Target of the modification}]
Except for the cases mentioned under Articles 90 and 96 of the Constitution, criminal proceedings against the President of the Republic, the President of the Senate, the President of the House of Representatives, and the Prime Minister, are suspended for the entire duration of tenure. [\dots]
\end{quote}
Suppose the Constitutional Court declares that the legislative act above is illegitimate and states its annulment. This means that such a norm no longer exists in the system. Hence, if issuing the norm corresponds to
\begin{gather*}
\zeta\colon \mathit{Parliament}, \mathit{Promulgation} \To_C (\mathit{L.\; 124}\colon\mathit{Crime}, \mathit{Tenure} \To_{\Obl} \mathit{Suspended})
\end{gather*}
then the annulment of $\mathit{L.\; 124}$ can be reconstructed by having, in the theory, a meta-rule such as
\begin{gather*}
\nu\colon \mathit{Constitutional\_Court} \To_C \neg (\mathit{L.\; 124}\colon\mathit{Crime}, \mathit{Tenure} \To_{\Obl} \mathit{Suspended}).
\end{gather*}
\end{example}
Clearly, $\nu$ conflicts with $\zeta$.

If conflicts between meta-rules occur only in cases like this, our variant of the logic extends the standard intuitions of rule conflict in DDL to meta-rules. On the basis of the discussion in Section \ref{sec:conceptual_framework}, we call this variant \textit{simple}. 

However, we can inspect the deontic meaning of rules occurring in the head of meta-rules, an inspection that allows us to introduce a second variant of the logic with meta-rules. Again based on the discussion in Section \ref{sec:conceptual_framework}, this variant, which we call \textit{cautious}, considers the following scenarios of when meta-rules conflict.

\paragraph{Case 1.}
Consider the following meta-rules:
\begin{gather*}
\zeta\colon A(\zeta) \To_\Box (\alpha\colon a\To_{\Box} b)\\
\nu\colon A(\nu) \To_\Box   (\beta\colon a \To_{\Box} \neg b)
\end{gather*}
The two meta-rules can be seen as incompatible (especially when $\To_{\Box}$ is $\To_{\Obl}$), since rules $\alpha$ and $\beta$ have exactly the same antecedent, they are labelled by the same $\Box$-modality but support complementary conclusions. A variant is
\begin{gather*}
\zeta\colon A(\zeta) \To_{\Box} (\alpha\colon a\To_{\Obl} b)\\
\nu\colon A(\nu) \To_{\Box}   (\beta\colon a \To_{\Perm} \neg b)
\end{gather*}
where the modalities of $\alpha$ and $\beta$ are different (specifically, one is obligation and the other permission), but we have a conflict as well.

\paragraph{Case 2.}
Consider the following meta-rules:
\begin{gather*}
\zeta\colon A(\zeta) \To_\Box (\alpha\colon a\To_{\Obl} b\otimes c)\\
\nu\colon A(\nu) \To_\Box   (\beta\colon a \To_{\Obl} \neg b\otimes d).
\end{gather*}
This is somehow similar to the previous case, but (i) it focuses on the obligation rules, and (ii) the head of rules are $\otimes$-expressions. Again, $\alpha$ and $\beta$ have exactly the same antecedent, but, although the heads state consistent compensations, their primary obligations are incompatible. Therefore, $\zeta$ and $\nu$ are in conflict because they support the conclusions $\Obl b$ and $\Obl \neg b$.  

\paragraph{Case 3.}
Consider the following meta-rules:
\begin{gather*}
\zeta\colon A(\zeta) \To_\Box (\alpha\colon a\To_{\Obl} b\otimes d)\\
\nu\colon A(\nu) \To_\Box   (\beta\colon a \To_{\Obl} c\otimes \non d).
\end{gather*}
Following the analysis for Case 2 above but for opposite reasons: the two meta-rules are not in conflict. Rules $\alpha$ and $\beta$ have exactly the same antecedent, but, although the heads state inconsistent compensations, their primary obligations are compatible. Ergo, $\zeta$ and $\nu$ are \emph{not} in conflict because support the conclusions $\Obl b$ and $\Obl c$.  

\paragraph{Case 4.}
Consider the following meta-rules:
\begin{gather*}
\zeta\colon A(\zeta) \To_\Box (\alpha\colon a\To_{\Obl} b)\\
\nu\colon A(\nu) \To_\Box   (\beta\colon a \To_{\Obl} \neg b\otimes d).
\end{gather*}
This is similar to Case 2; hence $\zeta$ and $\nu$ are in conflict.   

\paragraph{Case 5.}
Consider the following meta-rules:
\begin{gather*}
\zeta\colon A(\zeta) \To_\Box (\alpha\colon a\To_{\Obl} b)\\
\nu\colon A(\nu) \To_\Box   (\beta\colon a \To_{\Obl} b\otimes d).
\end{gather*}
Strictly speaking, these meta-rules are not in conflict as they support the derivation of rules both supporting $\Obl b$. On a different level -- the compliance angle -- we can argue that a theory containing both meta-rules is somehow odd since $\beta$ specifies that the violation of $b$ is compensated by $d$; hence, a situation $s$ satisfying $a$, $\neg b$ and $c$ complies with $\beta$. On the contrary, $\alpha$ does not admit a compensation for the violation of $b$, and the situation $s$ contravenes $\alpha$. Therefore, a system where both $\alpha$ and $\beta$ are in force admits a situation that is, at the same time, compliant and non-compliant.

\paragraph{Case 6.}
Consider the following meta-rules:
\begin{gather*}
\zeta\colon A(\zeta) \To_\Box (\alpha\colon a\To_{\Obl} b\otimes d)\\
\nu\colon A(\nu) \To_\Box   (\beta\colon a \To_{\Obl} b\otimes \neg d)
\end{gather*}
Following the analysis of other cases, the two meta-rules conflict. Rules $\alpha$ and $\beta$ have exactly the same antecedent and prove the same primary obligation, but they also state inconsistent compensations. In fact, if $\Obl b$ is violated, $\alpha$ and $\beta$ support opposite conclusions: $\Obl d$ and $\Obl \neg d$.  

\paragraph{Case 7.}
Consider the following meta-rules:
\begin{gather*}
\zeta\colon A(\zeta) \To_\Box (\alpha\colon a\To_{\Obl} b\otimes c)\\
\nu\colon A(\nu) \To_\Box   (\beta\colon a \To_{\Obl} b\otimes \neg d).
\end{gather*}
The two meta-rules are not in conflict, even though they can be seen as deontically odd: the `apparent oddity' lies in the fact that the same violation of $\Obl b$ leads to different compensations.

Consider this second legal example to further illustrate the intuitions behind the cautious logic's perspective.

\begin{example}\label{ex:derogation1}
Consider Art. 3 of the Italian Constitution:
\begin{quote}
Article 3\\
All citizens have equal social status and are equal before the
law, without regard to their sex, race, language, religion, political
opinions, and personal or social conditions.
\end{quote}
Suppose the Constitution is amended by stating the following:
\begin{quote}
In derogation to the provisions set out in Article 3, paragraph 1, of
the Constitution, EU citizens may have different social statuses when Italy is no longer a member state of the EU.
\end{quote}

If Art.~3 can be represented as follows
\[
\mathit{Art.\; 3}\colon \mathit{Citizen} \To_{\Obl} \mathit{Equal\_status}
\]
its amendment corresponds to applying the following meta-rule:
\[
\begin{array}{l}
derog_{Art.\;3}\colon \neg\mathit{EU}  \To_C 
(\mu\colon\mathit{EU\_Citizen} \To_{\Perm} \non \mathit{Equal\_status}).
\end{array}
\]
The example corresponds to the second variant of Case 1 above. Similar examples can be elaborated to illustrate the other scenarios previously discussed.
\end{example}

\subsection{Basics}
\label{sec:commonV1V2}

The new language accommodates both rules and meta-rules, where a meta-rule is a rule whose elements are other rules. We are now providing the formal machinery to capture such notions properly. The main idea is to re-use as much as possible the definitions and constructions given in Section~\ref{sec:Method} and to revise/extend them to include the new notions. In this way, the variants of the Defeasible Deontic Logic with meta-rules we are going to present in this Section are conservative extensions of the logic of Section~\ref{sec:Method}. Accordingly, for instance, there is no need to give a new definition of what a derivation is or to adjust the conditions when literals and deontic literals are provable. 

For the terminology used, we reserve the term \emph{standard rules} for expressions satisfying Definition~\ref{def:Rule} (rules containing only literals), and we call \emph{meta-rules} expressions containing other (standard) rules. Notice that we do not allow for the nesting of meta-rules.

In most cases, we do not need to distinguish between standard and meta-rules: Definition~\ref{def:RuleFinal} captures both into a single definition of what a rule is (when we speak of a generic rule, it can be either a standard rule or a meta-rule). 

\begin{definition}[Rule Expressions]\label{def:RuleExpression}
	Given a standard rule $\viola\alpha$, $\alpha$ and $\neg \alpha$ are \emph{rule expressions}; we use \Rex to denote the set of rule expressions. 
	If $\beta$ is a rule expression, then, for $\Box\in\set{\Obl,\Perm}$ $\Box \beta$ and $\neg \Box \beta$ are \emph{deontic rule expressions}.
\end{definition}
We also extend the definition of $\otimes$-expression allowing every element to be either a plain literal or a rule expression. 
Accordingly, the following is an $\otimes$-expression:
\[
a \otimes (\beta\colon b \To_\Obl c) \otimes (\gamma\colon d,\Obl a\To_\Obl e\otimes \neg f).
\]
Note that we can mix rules and literals in reparation chains. This allows us to represent situations where, for instance, an entity needs to enforce a particular policy and is subject to a sanction if it does not.

Consequently, we redefine the notion of rule as follows.

\begin{definition}[Rule]\label{def:RuleFinal}
	A \emph{rule} is an expression $\alpha: A(\alpha) \hookrightarrow_\Box C(\alpha)$, where
	\begin{enumerate}
		\item $\alpha \in \LABELS$ is the unique name of the rule.
		\item $A(\alpha)$ is a (possibly empty) set of elements $a_1, \dots, a_n$, where each $a_i$ is either a literal, a rule expression, or a deontic rule expression;
		\item $\hookrightarrow \in \set{\To, \defeater}$ with the same meaning as before;
		\item $\Box = \{\Cost, \Obl, \Perm \}$;
		\item its consequent $C(\alpha)$, which is either
		\begin{enumerate}
			\item a single plain literal $l \in \PLit$, or a rule expression $\beta$, if either (i) $\hookrightarrow \equiv \defeater$ or (ii) $\Box \in \set{\Cost, \Perm}$, or
			\item an $\otimes$-expression, if $\Box \equiv \Obl$.
		\end{enumerate}	
	\end{enumerate}
\end{definition}
We extend the convention used to represent the complement of a literal to the cases of (deontic) rule expressions: (i) if $\beta=\alpha$ then $\non \beta= \neg\alpha$, and (ii) if $\beta=\neg\alpha$ then $\non\beta=\alpha$.

Definition~\ref{def:RuleFinal} allows us to reuse the notation specified to identify particular sets of rules given in Section~\ref{sec:Method}.  Moreover, we can continue to use the notion of defeasible deontic theory just as defined in Definition \ref{def:DeonticTheory} but using the reviseddefinition of rule we just gavce. We extend the terminology and we say that a rule $\alpha$ \emph{appears in a rule} $\beta$, if $\alpha\in A(\beta)\cup C(\beta)$, and that $\alpha$ \emph{appears in a set of rules} $S$, if either $\alpha\in S$, or $\exists\beta\in S$ such that $\alpha$ appears in $\beta$. 

As standard rules may now be conclusions of meta-rules, we have to update the definition of tagged modal formula given in Definition~\ref{def:TagFormula} as follows.

\begin{definition}[Tagged modal formula]\label{def:MetaConclusion}
	An expression is a \emph{tagged modal formula} if satisfies Definition~\ref{def:TagFormula}, or has one of the following forms:
	\begin{itemize}
		\item $+\partial^m_\Box \alpha$: meaning that rule $\alpha$ is defeasibly provable with mode $\Box$, for $\Box\in\set{\Cost,\Obl,\Perm}$;
		\item $-\partial^m_\Box \alpha$:  meaning that rule $\alpha$ is defeasibly refuted with mode $\Box$, for $\Box\in\set{\Cost,\Obl,\Perm}$.
	\end{itemize}
\end{definition}
The intuition here is the same as for literals. In a defeasible theory with meta-rules, the conclusion of a derivation can be either a (deontic) literal or a (deontic) rule. Accordingly, when a theory proves $+\partial^m_{\Obl}\alpha$, it is obligatory to have the rule $\alpha$; when we prove a rule with mode $\Cost$, the meaning is that the rule is in the system and can produce its conclusion; $-\partial^m_{\Cost}\alpha$, on the other hand, indicates that rule $\alpha$ is not present in the underlying normative system. 

What about the case when we have the negation of a rule, $\neg\alpha$? For $+\partial^m_\Cost\neg\alpha$, two interpretations are possible: (i) an affirmative statement that $\alpha$ is not in the system, and (ii) that $\alpha$ has been removed from the rules of the system. The second reading was used in \cite{Governatori2009157} to model abrogation and annulment. For \Obl, a prohibition is defined as $\Obl\neg$. Accordingly, $+\partial^m_{\Obl}\neg\alpha$ specifies that, in the normative system, it is forbidden to have rule $\alpha$. For instance, Article 27 of the Italian Constitution makes capital punishment not admissible in the Italian Legal System.

The definition of Proof does not change wrt Definition~\ref{def:Proof}. However, we need to reformulate definitions of a rule being applicable/discarded to accommodate the case when a standard rule (or, more properly, a rule expression) is in the antecedent of a meta-rule.  A meta-rule is applicable when (i) the rule itself is provable and (ii) the rule expressions appearing in it are provable (or discarded) with the appropriate mode.

\begin{definition}[Applicability]\label{def:MetaApplicability}
	Assume a deontic defeasible theory $D$ with $D = (\FACTS, R, >)$.  We say that rule $\alpha \in R^\Cost \cup R^\Perm$ is \emph{applicable} at $P(n+1)$, iff  $+\partial^{m}_{\Cost}\alpha\in P(1..n)$ and for all $a \in A(\alpha)$ Conditions~\ref{item:l}--\ref{item:Diamondl} of Definition~\ref{def:StandardApplicability} hold and
	\begin{enumerate}
		\item\label{item:Beta} if $a = \beta \in \Rex$, then $+\partial_\Cost \beta \in P(1..n)$,

		\item if $a = \Box \beta$, $\beta \in \Rex$, then $+\partial_\Box \beta \in P(1..n)$, $\Box\in\set{\Obl,\Perm}$,

		\item\label{item:nonDiamondnonBeta} if $a = \neg\Box \beta$, $\beta \in \Rex$, then $-\partial_\Box \beta \in P(1..n)$, $\Box\in\set{\Obl,\Perm}$.
	\end{enumerate}
	We say that rule $\alpha \in R^\Obl$ is \emph{applicable at index} $i$ \emph{and} $P(n+1)$ iff  $+\partial^{m}_{\Cost}\alpha\in P(1..n)$ and Conditions~\ref{item:l}--3 of Definition~\ref{def:StandardApplicability} and Conditions \ref{item:Beta}-\ref{item:nonDiamondnonBeta} above hold, and for all $c_j \in C(\alpha),\, j < i$
	\begin{enumerate}[resume]
	    \item
	    \begin{enumerate}[label={\alph*.}]
	        \item if $c_j = l$, then  $+\partial_\Obl l \in P(1..n)$ and $+\partial_\Cost \non l \in P(1..n)$, $l \in \PLit$;  
	        \item if $c_j = \beta$, then  $+\partial_\Obl \beta \in P(1..n)$ and $-\partial_\Cost \beta \in P(1..n)$. 
	    \end{enumerate}
		
	\end{enumerate}
	
\end{definition}
Note that, whilst 4.a is the same of Condition 4 of Definition 4, when $c_j$ is a rule then we require in Condition 4.b only $-\partial_\Cost \beta$ and not $+\partial_\Cost\non\beta$. Both conditions indicate that rule $\beta$ is not a rule effective. As discussed in Footnote~\ref{foot:violation}, the former means that there is no evidence that the rule is in the system, while the latter guarantees that the rule is removed from or prevented from being in the system. In a positivist view, a normative system explicitly gives the norms that hold in it. Consequently, the set of effective norms (the norms that can produce an effect) is constrained by the norms that appear in the normative system. We have argued that $-\partial^m_\Cost\alpha$ states that rule/norm $\alpha$ is not in the system (or that it is not effective). Therefore, this is enough to contravene the obligation of the norm to be effective in the normative system.

\begin{definition}[Discardability]\label{def:MetaDiscardability}
	Assume a deontic defeasible theory $D$ with $D = (\FACTS, R, >)$.  We say that rule $\alpha \in R^\Cost \cup R^\Perm$ is \emph{discarded} at $P(n+1)$, iff $-\partial^{m}_{\Cost}\alpha\in P(1..n)$ or at at least one of the  Conditions~\ref{item:notl}--\ref{item:notDiamondl} of Definition~\ref{def:StandardDiscardability} holds, or there exists $a \in A(\alpha)$ such that

	\begin{enumerate}
		\item\label{item:notBeta} if $a = \beta \in \Rex$, then $-\partial_\Cost \beta \in P(1..n)$,
        \item if $a = \Box \beta$, $\beta \in \Rex$, then $-\partial_\Box \beta \in P(1..n)$, $\Box\in\set{\Obl,\Perm}$
		\item\label{item:notDiamondBeta} if $a = \neg\Box \beta$, $\beta \in \Rex$, then $+\partial_\Box \beta \in P(1..n)$, $\Box\in\set{\Obl,\Perm}$.
	\end{enumerate}
	We say that rule $\alpha \in R^\Obl$ is \emph{discarded at index} $i$ \emph{and} $P(n+1)$ iff $-\partial^{m}_{\Cost}\alpha\in P(1..n)$ or at least one of the Conditions~\ref{item:notl}--3 of Definition~\ref{def:StandardDiscardability} or of the Conditions~\ref{item:notBeta}--\ref{item:notDiamondBeta} above holds or there exists $c_j \in C(\alpha),\, j < i$ such that
	\begin{enumerate}[resume]
		\item 
		\begin{enumerate}[label={\alph*.}]
	        \item if $c_j = l$, then $-\partial_\Obl l \in P(1..n)$ or	$-\partial_\Cost \non l \in P(1..n)$, $l \in \PLit$;  
	        \item if $c_j = \beta$, then  $-\partial_\Obl \beta \in P(1..n)$ or $+\partial_\Cost \beta \in P(1..n)$. 
	    \end{enumerate}
	\end{enumerate}

\end{definition}
Before moving to the presentation of the two variants of the logic, we introduce a key feature of our logical apparatus: the notion of \emph{conflict of rules}. The definition below formalises the intuitions discussed at the beginning of this section. 

\begin{definition}[Simple Conflict]\label{def:SimpleConflict}
    Given two rules or rule expressions $\alpha$ and $\beta$, we say that $\alpha$ and $\beta$ \emph{simply conflict} iff
    \begin{enumerate}
        \item
        \begin{description}
            \item $\alpha\colon A(\alpha) \hookrightarrow_{\Box} C(\alpha)$ 
            \item $\beta$ is $\non(\gamma\colon A(\alpha) \hookrightarrow_{\Box} C(\alpha))$
        \end{description}
        \item  If $\nu\in C(\alpha)$ at index $i$, $\zeta\in C(\beta)$ at index $j$, and $\nu$ and $\zeta$ simply conflict.
    \end{enumerate}
\end{definition}

\begin{definition}[Cautious Conflict]\label{def:RationalConflict}
    Given two rules or rule expressions $\alpha$ and $\beta$, we say that $\alpha$ and $\beta$ \emph{cautiously conflict} iff
    \begin{enumerate}
        \item 
        \begin{description}
            \item $\alpha\colon A(\alpha) \hookrightarrow_{\Box} C(\alpha)$ 
            \item $\beta$ is $\non(\gamma\colon A(\alpha) \hookrightarrow_{\Box} C(\alpha))$

        \end{description}
        \item 
        \begin{description}
            \item $\alpha\colon A(\alpha) \hookrightarrow_{\Box} C(\alpha)$, 
            \item $\beta\colon A(\alpha) \hookrightarrow_\Box  \non C(\alpha)$;
        \end{description}
            \item 
        \begin{description}
            \item $\alpha\colon A(\alpha) \hookrightarrow_{\Obl} C(\alpha)$, 
            \item $\beta\colon A(\alpha) \hookrightarrow_{\Perm} \non C(\alpha)$;
        \end{description}
        \item
            \begin{description}
            \item $\alpha\colon A(\alpha)  \To_{\Obl} c_1\otimes\dots\otimes c_m$,
            \item $\beta\colon A(\alpha)  \To_{\Obl} d_1\otimes\dots\otimes d_n$,  and
                \begin{enumerate}
                    \item $\exists i \leq m, n,\ \forall k < i.\, c_k = d_k \wedge c_i = \non d_i$; or
                    \item $m < n \wedge \forall i \leq m$ $c_i=d_i$.
                \end{enumerate}
            \end{description}
        \item If $\nu\in C(\alpha)$ at index $i$, $\zeta\in C(\beta)$ at index $j$, and $\nu$ and $\zeta$ cautiously conflict.
    \end{enumerate}
\end{definition}
Given two rules or rule expressions $\alpha$ and $\beta$ that (simply/cautiously) conflict, we will say that $\alpha$ (simply/cautiously) conflicts with $\beta$ (and the other way around).
Notice that the definitions above apply to both standard rules and meta-rules; only the recursive conditions (condition 2 
in Defintion~\ref{def:SimpleConflict} and condition 5 of Definition~\ref{def:RationalConflict}) are limited to meta-rules.  
It is easy to verify that every simple conflict is also a cautious conflict. 
\begin{example}
The most immediate example of a simple conflict is given by two rule expressions like
\[
	\dRule\alpha:a=>\Cost b \qquad
	\neg(\dRule\alpha:a=>\Cost b).
\]
Here the two expressions share the same rule label. However, we have a simple conflict even when the rule label is different. Thus, 
\[
	\dRule\beta:a,\Obl b=>\Perm c\qquad
	\neg(\dRule\gamma:a,\Obl b=>\Perm c)
\]
simply conflict with each other. 

Two meta-rules are in a simple conflict relation when their conclusions are in the same relation. Therefore, 
\[	\dRule\epsilon:A(\epsilon)=>\Obl{a\otimes(\dRule\zeta:c=>\Cost d)}\qquad		\dRule\eta:A(\eta)=>\Obl{a\otimes\neg(\dRule\theta:c=>\Cost d)}
\]
simply conflict with each other since the initial part of the $\otimes$-chain in their conclusion is the same, but the next elements are conflicting rule expressions. 

For the notion of cautious conflict, the rules
\[
	\dRule\alpha:a=>\Cost b \qquad
	\dRule\kappa:a=>\Cost \neg b
\]
and the meta-rules
\[
	\dRule\lambda:{(\dRule\alpha:a=>\Cost b),e}=>\Obl f 
	\qquad
	\dRule\mu:{(\dRule\alpha:a=>\Cost b),e}=>\Perm\neg f 
\]
conflict since, for each pair, they have the same antecedent and opposite conclusions (the meta-rules also show an $\Obl$/$\Perm$ conflict).  The next pair of meta-rules
\begin{gather*}
\dRule\nu:\Obl({\dRule\alpha:a=>\Cost b)}=>\Obl{(\dRule\beta:a,\Obl b=>\Perm c)
 			\otimes
 			(\dRule\xi:f=>\Obl g\otimes h)}
\\
\dRule o:{\neg(\dRule\kappa:a=>\Cost \neg b)}=>\Obl{c\otimes d
 			\otimes
 			(\dRule\pi:f=>\Obl g\otimes \neg h)}
\end{gather*}
cautiously conflict because the rules $\xi$ and $\pi$ do: they have the same antecednet, the same element at index 1, but complementary literals at index 2. Notice, that $\nu$ and $o$ can have different antecedents, and the conflcliting rules in their conclusions can appear at different indices.  
\end{example}
In the rest of the section, we shall introduce two variants of the logic, where in addition to proving literals we prove rules. The two variants share the proof conditions for literals given in Section~\ref{sec:Method} (Definitions \ref{def:StandardCostProof}--\ref{def:StandardPermProof}), whilst they differ from each other in the proof conditions to derive (standard) rules.

\subsection{Simple Conflict Defeasible Deontic Logic}\label{sec:Variante1}

We are now ready to present the proof conditions for the Simple Conflict variant of Defeasible Deontic Logic. In this variant, we focus on the case of applicable meta-rules for a standard rule and its negation. In other words, there are two meta-rules: one to introduce a rule in the system and one to remove it (or to prevent it from being inserted). To handle this case, we restrict the logic to the notion of simple conflicts. 

\begin{definition}[Constitutive Meta-Proof Conditions -- Simple Variant]\label{def:ConstMetaPT-V1}

\begin{tabbing}
$+\partial_\Cost^m \alpha$: \=If $P(n+1)=+\partial_\Cost^m \alpha$ then\+\\
  (1) $\alpha\in R$ or\\
  (2) \=(1) $\forall \omega\in R$, $\omega$ does not simply conflict with $\alpha$, and
  \+\\
        (2) \=$\exists \beta \in R_\To^\Cost[(\alpha: A(\alpha) \hookrightarrow_\Box c)]$ s.t.\+\\
            (1) $\beta$ is applicable, and\\
            (2) \=$\forall \gamma\in R^\Cost[\non(\varphi: A(\alpha) \hookrightarrow_\Box c)]$  either\+\\
                (1) $\gamma$ is discarded, or \\
                (2) \= $\exists \zeta \in R^\Cost[(\chi : A(\alpha) \hookrightarrow_\Box c)]$ s.t.\+\\
	                (1) $\chi \in \set{\alpha, \varphi}$,\\
		            (2) $\zeta$ is applicable, and\\
		            (3) $\zeta > \gamma$.
\end{tabbing}

\begin{tabbing}
  $-\partial_\Cost^m \alpha$: \=If $P(n+1)=-\partial_\Cost^m \alpha$ then\+\\
    (1) $\alpha\notin R$ and\\
    (2) \=(1) $\exists\rho\in R$, $\rho$ simply conflicts with $\alpha$, or\+\\
        (2) \= $\forall \beta \in R_\To^\Cost[(\alpha: A(\alpha) \hookrightarrow_\Box c)]$ either \+\\ 
            (1) $\beta$ is discarded, or\\
            (2) \=$\exists \gamma\in R^\Cost[\non(\varphi: A(\alpha) \hookrightarrow_\Box c)]$, s.t. \+\\
                (1) $\gamma$ is applicable \\
                (2) \= $\forall \zeta \in R^\Cost[(\chi : A(\alpha) \hookrightarrow_\Box c)]$ \+\\
	                (1) $\chi \notin \set{\alpha, \varphi}$, or\\
		            (2) $\zeta$ is discarded, or\\
		            (3) $\zeta \not> \gamma$.
\end{tabbing} 

\end{definition}
For the explanation, we focus on the positive case. We first remark that the general structure of the proof conditions for rules is the same as that for literals. Either the conclusion is given in the theory, or there is an applicable rule for it and all possible attacks are either discarded or defeated. Ergo, the first option to prove a rule is to see if it is one of the rules given in $R$. Notice that this step applies to both standard rules and meta-rules (Condition 1). If rule $\alpha$ is not in $R$, Condition (2) checks that rules simply conflicting with $\alpha$ are not in the set of the initial, given rules $R$; if this step succeeds, we then look for an applicable meta-rule whose conclusion is $\alpha$. It is worth noticing that, with the extended definition of applicability, $\beta$ has to be provable, and this condition applies to the proof conditions for literals. Next, we have to examine all meta-rules for the ``opposite'' of $\alpha$: here, we consider all rule expressions starting with a negation and having the same antecedent as well as the same consequent of $\alpha$; nevertheless, they can have a different label (we do not impose that label $\varphi$ must be $\alpha$). The reason is that it would be irrational to introduce a rule (with a given name) and, at the same time to remove it (potentially with a different name). For all attacking meta-rules, we have to rebut them. As usual, there are two ways: either the attacking rule is discarded or is defeated. However, contrary to the attacking phase, the defeating meta-rules must be for either $\alpha$ or $\varphi$; in other terms, we cannot reinstate a specific rule using other generic rules. 

\begin{example}\label{ex:MetaConstV1-1}
Consider the theory $D$ where $a\in F$, the following rules are in $R$ \footnote{From now on, we use the convention to use $\dots$ to indicate the antecedent of rules that are applicable and where the content of the antecedent is not relevant to the discussion.} 
\begin{align*}
&\drule{\beta}{\dots}{(\drule{\alpha}{a}{b})} &
&\drule{\eta}{\dots}{c}&
&\Drule{\lambda}{\dots}{(\drule{\alpha}{a}{b})}\\
&\drule{\gamma}{c}{\neg(\drule{\epsilon}{a}{b})} &
&\drule{\theta}{\dots}{\neg(\drule{\eta}{\dots}{c})} &
&\drule{\mu}{d}{\neg(\drule{\alpha}{a}{b})}
\end{align*}
and $\lambda>\gamma$.

It is immediate to see that the meta-rules $\beta,\gamma,\theta,\lambda,\mu$ are provable being in $R$. Also we have $+\partial^m_\Cost\eta$, while there is an applicable meta-rule for $\neg\eta$ ($\theta$), $\eta\in R$ and, by Condition (1), this takes precedence, and by the conjunction of Conditions (1) and (2.1)  for $-\partial^m_\Cost$, we conclude $-\partial^m_\Cost\neg\eta$. At this stage, $\eta$ is applicable and, since there are no rules for $\neg c$, we derive $+\partial_\Cost c$; this makes $\gamma$ applicable. Now, we have an applicable meta-rule, $\beta$, for $\alpha$. We have to look for meta-rules for standard rules that simply conflict with $\alpha$; we have two candidates: $\gamma$ and $\mu$. There is no rule for $d$; hence $-\partial_\Cost d$ holds, and  $\mu$ is discarded. On the other hand, as we just discussed, $\gamma$ is applicable. Thus we have to determine if it is defeated or not. Again, two options: $\beta$ itself, but $\beta$ is not stronger than $\gamma$; and $\lambda$ which, even if it is a defeater, succeeds in defeating $\gamma$, we hence derive $+\partial^m_\Cost\alpha$.  Finally, since $\alpha$ is applicable and there are no rules for $\neg b$, then $+\partial_\Cost b$ holds in $D$. 

Suppose now that we extend the theory by inserting the rule $\rho$
\[
    \drule[\Obl]{\rho}{(\drule{\alpha}{a}{b})}{e}
\]
in $R$. Since $\rho$ is in $R$, then it is derivable. Given that we proved $+\partial^m_\Cost\alpha$, $\rho$ is applicable, we can now prove $+\partial_\Obl e$. This shows that it is possible to have a meta-rule whose conclusion is not a rule but a literal. Suppose now that, in addition to update $R$ with $\rho$, we also replace $\lambda$ with 
\[
\drule{\nu}{\dots}{\neg(\drule{\sigma}{a}{b})}
\]
and include $\nu>\gamma$.
Given that $\sigma$ is neither $\alpha$ nor $\epsilon$, we cannot use $\nu$ to reinstate $\alpha$ (even if $\nu$ is stronger than $\gamma$). This means, that now $D\vdash-\partial_\Cost^m\alpha$, and then $D\vdash-\partial_\Obl e$. However, we can use $\nu$ in both Conditions (2.2) and (2.2.2.2), to prove $+\partial^m_\Cost\sigma$; again we have an applicable undefeated rule for $b$, and $+\partial_\Cost b$ continues to hold.
\end{example}

\begin{definition}[Obligation Meta-Proof Conditions -- Simple Variant]\label{def:OblMetaPT-V1}

\begin{tabbing}
  $+\partial^m_\Obl \alpha$: \=If $P(n+1)=+\partial^m_\Obl \alpha$ then\+\\
    $\exists \beta \in R_\To^\Obl[(\alpha: A(\alpha) \hookrightarrow_\Box c), i]$ s.t.\\ 
    (1) $\beta$ is applicable at index $i$, and\\
    (2) \=$\forall \gamma\in R^\Obl[\non(\psi: A(\alpha) \hookrightarrow_\Box c), j] \cup R^\Perm[\non(\psi: A(\alpha) \hookrightarrow_\Box c)]$ either\\
        \>(1) $\gamma$ is discarded (at index $j$), or \\
        \>(2) \= $\exists \zeta \in R^\Obl[(\chi : A(\alpha) \hookrightarrow_\Box c), k]$ s.t.\\
		    \>\>(1) $\chi \in \set{\alpha, \psi}$,\\
		    \>\>(2) $\zeta$ is applicable at index $k$, and\\
		    \>\>(3) $\zeta > \gamma$.
\end{tabbing} 

\begin{tabbing}
  $-\partial^m_\Obl \alpha$: \=If $P(n+1)=-\partial^m_\Obl \alpha$ then\+\\
    $\forall \beta \in R_\To^\Obl[(\alpha: A(\alpha) \hookrightarrow_\Box c), i]$ either\\
    (1) $\beta$ is discarded at index $i$, or\\
    (2) \=$\exists \gamma\in R^\Obl[\non(\psi: A(\alpha) \hookrightarrow_\Box c), j] \cup R^\Perm[\non(\psi: A(\alpha) \hookrightarrow_\Box c)]$, s.t.\\
        \>(1) $\gamma$ is applicable (at index $j$), and \\
        \>(2) \= $\forall \zeta \in R^\Obl[(\chi : A(\alpha) \hookrightarrow_\Box c), k]$ either\\
		    \>\>(1) $\chi \not\in \set{\alpha, \psi}$, or\\
		    \>\>(2) $\zeta$ is discarded at index $k$, or\\
		    \>\>(3) $\zeta \not> \gamma$.
\end{tabbing} 
\end{definition}
The conditions to derive when a rule is proved or refuted as an obligation combine: (i) the issues to derive a literal as an obligation, and (ii) the above elements for proving a rule. Note that a deontic rule expression can appear in the body of a rule but cannot stand on its own. Thus $F$ and $R$ cannot contain expressions like $\Obl\alpha$.

\begin{example}\label{ex:MetaObligation}
Let $D$ be a theory where $a,d\in\FACTS$, $R$ contains the following rules
\begin{align*}
 &\dRule\alpha:\dots=>\Obl {
    (\dRule\gamma:a=>\Cost c) \otimes
    c \otimes
    (\dRule\epsilon: d=>\Obl {e\otimes f})
} &
&\dRule\beta:\dots=>\Cost {(\dRule\eta: a=>\Cost {\neg c})}\\
&\dRule\theta:\Perm({\dRule\epsilon: d=>\Obl {e\otimes f}})=>\Obl {\neg(\dRule\kappa:\Obl e,\neg e=>\Obl h)} &
& \dRule\lambda:g=>\Perm {(\dRule\kappa:\Obl e,\neg e=>\Obl h)}
\end{align*}
and $\lambda>\theta$.

Let us consider the following derivation (where, for the sake of clarity, we removed trivial and irrelevant steps):

\begin{tabbing}xxx\=xxxxxxxxxxxxxxxxxxxxx\=\kill
 1. \>$+\partial^m_\Obl (\dRule\gamma:a=>\Cost c)$  \> $\alpha$ applicable for $\gamma$ at index $1$, $R^\Diamond[\neg\alpha]=\emptyset$ \\
 2. \>$-\partial^m_\Cost (\dRule\gamma:a=>\Cost c)$ \> $\gamma\notin R$ and $R^\Cost[\gamma]=\emptyset$ \\
 3. \>$+\partial_\Obl c$ \>  1. and 2. $\alpha$ applicable for $c$ at index $2$, $R^\Diamond[\neg c]=\emptyset$\\
 4. \>$+\partial^m_\Cost (\dRule\eta:a=>\Cost \neg c)$ \> $\beta$ applicable and $R^\Cost[\neg\eta]=\emptyset$\\
 5. \>$+\partial_\Cost a$ \> $a\in\FACTS$\\
 6. \>$+\partial_\Cost \neg c$ \> 4. and 5. $\eta$ applicable and $R^\Cost[\neg c]=\emptyset$\\
 7. \>$+\partial^m_\Obl(\dRule\epsilon: d=>\Obl {e\otimes f})$ \> 3. and 6. $\alpha$ applicable for $\epsilon$ at index 3, $R^\Diamond[\neg\epsilon]=\emptyset$\\
 8. \>$+\partial^m_\Perm(\dRule\epsilon: d=>\Obl {e\otimes f})$ \> $+\partial^m_\Obl(\epsilon)\in P(1..7)$, Condition (1) of $+\partial^m_\Perm$ of Def. \ref{def:PermMetaPT-V1}\\
 9. \>$-\partial_\Cost g$ \> $g\notin\FACTS$ and $R^\Cost[g]=\emptyset$\\
 10.\>$+\partial^m_\Obl \neg(\dRule\kappa:\Obl e,\neg e=>\Obl h)$ \> $\theta$ applicable (7.), $\lambda$ discarded (9.)\\
 11. \>$+\partial_\Cost d$ \> $d\in\FACTS$\\
 12. \>$+\partial_\Obl e$ \> 7. and 11. $\epsilon$ applicable for $e$ at index 1, $R^\Diamond[\neg e]=\emptyset$\\
 13. \>$-\partial_\Cost \neg e$ \> $\neg e\notin\FACTS$ and $R^\Cost[\neg e]=\emptyset$\\
 14. \>$-\partial_\Obl f$ \> 13. $\epsilon$ discarded for $f$ at index 2
\end{tabbing}
The derivation above mostly illustrates the proof conditions for $+\partial^m_\Obl$ with repeated use of Condition 5 of Definition~\ref{def:MetaApplicability}, and particularly the distinction between rules and literals appearing in $\otimes$-expressions: compare the combinations Steps 1. and 2. to obtain Step 3., and Steps 3. and 6. to conclude Step 7.  Note that most cases of discarded occur when there are no rules for the opposite. 

Let us focus on the rules in the theory. First of all, rule $\theta$ shows that it is possible to have rule expressions in the antecedent of meta-rules. Second, let us analyse the meaning of rules $\theta$ and $\lambda$: $\lambda$ allows $\kappa$ to be an admissible norm in the normative system. In case $\epsilon$ is present and effective (produces the effects of its conclusion, $\Obl e$), $\kappa$ corresponds to a contrary-to-duty norm whose normative effect is in force in case of a violation. Typically, this kind of expression represents an accessory penalty (see \cite{icail2015:thou} for the distinction between \emph{compensatory obligations} and \emph{contrary-to-duty obligations}). On the other hand, $\epsilon$ establishes a compensatory condition for violating the obligation of $e$. Accordingly, $\theta$ prescribes that it is forbidden to have an accessory penalty if a compensatory condition is admissible.   
\end{example}

\begin{definition}[Permission Meta-Proof Conditions -- Simple Variant]\label{def:PermMetaPT-V1}
\begin{tabbing}
  $+\partial^m_\Perm \alpha$: \=If $P(n+1)=+\partial^m_\Perm \alpha$ then\+\\
    (1) \= $+\partial^m_\Obl \alpha \in P(1..n)$, or\\
    (2) \= $\exists \beta \in R_\To^\Perm[(\alpha: A(\alpha) \hookrightarrow_\Box c)]$ s.t.\+\\ 
        (1) $\beta$ is applicable, and\\
        (2) \=$\forall \gamma\in R^\Obl[\non(\varphi: A(\alpha) \hookrightarrow_\Box c), j]$  either\\
        \>(1) $\gamma$ is discarded at index $j$, or \\
        \>(2) \= $\exists \zeta \in R^\Diamond[(\chi : A(\alpha) \hookrightarrow_\Box c), k]$ s.t.\\
	        \>\>(1) $\chi \in \set{\alpha, \varphi}$,\\
		    \>\>(2) $\varepsilon$ is applicable (at index $k$), and\\
		    \>\>(3) $\varepsilon > \gamma$.
\end{tabbing}

\begin{tabbing}
  $-\partial^m_\Perm \alpha$: \=If $P(n+1)=-\partial^m_\Perm \alpha$ then\+\\
    (1) \= $-\partial^m_\Obl \alpha \in P(1..n)$, and\\
    (2) \= $\forall \beta \in R_\To^\Perm[(\alpha: A(\alpha) \hookrightarrow_\Box c)]$ either \+\\ 
        (1) $\beta$ is discarded, or\\
        (2) \=$\exists \gamma\in R^\Obl[\non(\varphi: A(\alpha) \hookrightarrow_\Box c), j]$, s.t.\\
        \>(1) $\gamma$ is applicable at index $j$, and \\
        \>(2) \= $\forall \zeta \in R^\Diamond[(\chi : A(\alpha) \hookrightarrow_\Box c), k]$ either\\
	        \>\>(1) $\chi \not\in \set{\alpha, \varphi}$, or\\
		    \>\>(2) $\varepsilon$ is discarded (at index $k$), or\\
		    \>\>(3) $\varepsilon \not > \gamma$.
\end{tabbing}
\end{definition}
Similarly to $\partial_\Obl^{m}$, the conditions to determine the provability of rules as permission combine the elements of $\partial_\Perm$ and the new features included in $\partial^m_\Cost$. As illustrated by Example~\ref{ex:MetaObligation} above, Condition (1) corresponds to the `obligation implies permission principle' (axiom D), characteristic of Deontic Logic.

\begin{example}\label{ex:MetaPerm-V1}
Consider theory $D$ with the following four applicable rules
\begin{align*}
    &\dRule\beta:\dots=>\Perm {(\dRule\alpha:a=>\Perm b)} &
    &\dRule\gamma:\dots=>\Cost {(\dRule\alpha:a=>\Perm b)}\\
    &\dRule\eta:\dots=>\Perm {\neg(\dRule\alpha:a=>\Perm b)} &
    &\dRule\theta:\dots=>\Cost {\neg(\dRule\alpha:a=>\Perm b)}.
\end{align*}
It is immediate to see that $\beta$ and $\eta$ conflict with each other, and so are $\gamma$ and $\theta$ (in both cases, the conclusion of a rule is the complement of the other). However, in Condition (2.2) of $+\partial^m_\Perm$, only obligation rules attack a permissive rule, and no obligation rules conflict with $\beta$ (and symmetrically $\eta$); therefore, Condition (2.2) is vacuously satisfied: we have $+\partial^m_\Perm\alpha$ and $+\partial^m_\Perm\neg\alpha$. Conversely, we have to evaluate $\theta$ according to Condition (2.2.2) of $+\partial^m_\Cost$ to determine whether $D\vdash+\partial^m_\Cost\alpha$. Since $\theta$ is neither discarded nor defeated, $\alpha$ is not defeasibly provable, and we can repeat the argument for $\neg\alpha$. Consequently, $D\vdash-\partial^m_\Cost\alpha$ and $D\vdash-\partial^m_\Cost\neg\alpha$. The example illustrates the case where it is possible for a normative system to make admissible a norm and, at the same time, it is admissible not to have that norm. However, it is not possible to have a norm and not to have the same norm (or alternatively, to include a norm and remove it at the same time).  
\end{example}
We will use $D\vdash_S\pm\# c$ to indicate that there is a proof of $\pm\# c$ from $D$ using the proof conditions for Simple Conflict Defeasible Deontic Logic.

\subsection{Cautious Conflict Defeasible Deontic Logic}\label{sec:Variante2}

We now introduce a second variant of the logic with meta-rules. This variant considers a different idea of when meta-rules conflict. In the previous variant, the idea was that a conflict occurs when two statements fire concurrently, one affirming that a rule holds (is, or is inserted, in the system), and the second denying the rule (i.e., asserting that the rule is not in the system, has been removed). The simple variant took into account the specific label of the rule.

As discussed in the initial part of this section, the second variant (cautious) focuses on the ``content'' of the rules. It is partially inspired by some statutory interpretation principles, mostly the principle of non-redundancy of the prescriptions in a statute.  Suppose that we have the rules
\begin{align*}
    \dRule\alpha:a=>\Cost b && \dRule\beta:a=>\Cost\neg b.
\end{align*}
The logic of Section~\ref{sec:Method} is very well-equipped to handle such a case. If $a$ is derivable, both rules are applicable, and they are for opposite conclusions. The logic is sceptical, and it prevents to derive positive conclusions. It derives that both $b$ and $\neg b$ are refuted. Given that the two rules have exactly the same antecedent, it is \emph{not} possible that one of them is applicable and the other is not. Therefore, (i) the combination of the two rules introduces a normative gap: when $a$ holds, we are not able to determine if $b $ or $\neg b$ is the case; (ii) suppose that the theory specifies that one rule is stronger than the other. In this case, we can derive the conclusion of the stronger rule. Given that the two rules have the same antecedent, and the weaker rule is always defeated, there is no situation where the weaker rule can produce its conclusion. Accordingly, the weaker rule is redundant. 

The principle of non-redundancy in statutory interpretation states that every term in a statute has a purpose \cite{bowman}. Therefore, if a provision in a statute never produces an effect, then there is no purpose for such a provision, and the provision is redundant. The proof conditions we present in the rest of this section implement two key elements: (i) at every step, they look for a rule whose content is incompatible with the content of the rule we want to prove, and (ii) they extend the superiority relation mechanism to check not only if a meta-rule is stronger than another (meta-)rule, but also the superiority for the (standard) rules themselves.

We have now all the tools to reformulate the proof tags for proving/refuting rules in Definitions~\ref{def:ConstMetaPT-V1}--\ref{def:PermMetaPT-V1} to accommodate the new ideas of conflicts introduced at the beginning of Section~\ref{sec:Logic}.

\begin{definition}[Constitutive Meta-Proof Conditions -- Cautious Variant]\label{def:ConstMetaPT-V2}

\begin{tabbing}
  $+\partial_\Cost^m \alpha$: \=If $P(n+1)=+\partial_\Cost^m \alpha$ then\+\\
  (1) $\alpha\in R$ or\\
  (2) \=(1) $\forall \omega\in R$, $\omega$ does not cautiously conflict with $\alpha$, and\+\\
    (2) \= $\exists \beta \in R_\To^\Cost[\alpha]$ s.t.\+\\
        (1) $\beta$ is applicable, and\\
        (2) \=$\forall \gamma \in R^\Cost$ s.t. $\gamma$ cautiously conflicts with $\beta$, either\+\\
            (1) $\gamma$ is discarded, or \\
            (2) \= $\exists \zeta \in R^\Cost$ s.t. $\zeta$ cautiously conflicts with $\gamma$, and \+\\
		        (1) \= $\zeta$ is applicable, and \\
		        (2) \= (1) \=$\zeta > \gamma$, or\+\\
		               (2) $\gamma \not> \zeta$, and $C(\zeta) > C(\gamma)$.
\end{tabbing} 

\begin{tabbing}
  $-\partial_\Cost^m \alpha$: \=If $P(n+1)=-\partial_\Cost^m \alpha$ then\+\\
  (1) $\alpha\notin R$ and\\
  (2) \=(1) $\exists \omega\in R$, $\omega$ cautiously conflicts with $\alpha$, or\+\\
    (2) \= $\forall \beta \in R_\To^\Cost[\alpha]$ either\+\\
        (1) $\beta$ is discarded, or\\
        (2) \=$\exists \gamma \in R^\Cost$ s.t. $\gamma$ cautiously conflicts with $\beta$, s.t.\+\\
            (1) $\gamma$ is applicable, or \\
            (2) \= $\forall \zeta \in R^\Cost$ s.t. $\zeta$ cautiously conflicts with $\gamma$, either\+\\
		        (1) \= $\zeta$ is discarded, or\\
		        (2) (1) $\zeta \not> \gamma$, and\+\\
		            (2) $\gamma> \zeta$ or $C(\zeta) \not> C(\gamma)$.
\end{tabbing} 
\end{definition}
Notice that the only new parts wrt Definition~\ref{def:ConstMetaPT-V1} are: (i) Condition (2.1) uses cautiously conflict instead of simply conflict, and (ii) Condition (2.2.2.2.2) verifies whether the meta-rule supporting the conclusion is stronger than the meta-rule attacking the conclusion ($\gamma$); if this is not the case, we can see if the is an instance of the superiority relation for the standard rules (i.e., $C(\zeta) > C(\gamma)$).

\begin{example}\label{ex:MetaCost-v2}
Consider again the theory of Example~\ref{ex:MetaConstV1-1}, with $\lambda>\gamma$, and:
\begin{align*}
&\drule{\beta}{\dots}{(\drule{\alpha}{a}{b})} &
&\drule{\eta}{\dots}{c}&
&\Drule{\lambda}{\dots}{(\drule{\_}{a}{b})}\\
&\drule{\gamma}{c}{\neg(\drule{\epsilon}{a}{b})} &
&\drule{\theta}{\dots}{\neg(\drule{\eta}{\dots}{c})} &
&\drule{\mu}{d}{\neg(\drule{\alpha}{a}{b})}.
\end{align*}
All rules that were in conflict according to the Simple Conflict variant are in conflict in the Cautious Conflict version. Now, we do not have to ensure that the label of the standard rule that is the conclusion of $\lambda$ is either the label of the  rule that is the conclusion of $\beta$, or the conclusion of $\gamma$. Any rule conflicting with $\epsilon$ will do. 
\end{example}

The following two definitions, that determine the proof conditions for obligations and permissions, are obtained directly from Definitions~\ref{def:OblMetaPT-V1} and \ref{def:PermMetaPT-V1} using the changes we just explained for Definition~\ref{def:ConstMetaPT-V2}.

\begin{definition}[Obligation Meta-Proof Conditions -- Cautious Variant]\label{def:OblMetaPT-V2}

\begin{tabbing}
  $+\partial_\Obl^m \alpha$: \=If $P(n+1)=+\partial_\Obl^m \alpha$ then\+\\
    $\exists \beta \in R_\To^\Obl[\alpha, i]$ s.t. \\
    (1) $\beta$ is applicable at index $i$, and\\
    (2) \=$\forall \gamma \in R^\Obl[\nu, j] \cup R^\Perm[\nu]$ s.t. $\gamma$ cautiously conflicts with $\beta$, either\\
        \>(1) $\gamma$ is discarded (at index $j$), or \\
      \>(2) \= $\exists \zeta \in R^\Obl[\chi, k]$ s.t. $\zeta$ cautiously conflicts with $\gamma$\\
		\>\> (1) \= $\zeta$ is applicable (at index $k$), and either\\
		\>\> (2) (1) $\zeta > \gamma$, or\\
		    \>\>\> (2) $\gamma \not> \zeta$, and $C(\zeta) > C(\gamma)$.
\end{tabbing} 

\begin{tabbing}
  $-\partial_\Obl^m \alpha$: \=If $P(n+1)=-\partial_\Obl^m \alpha$ then\+\\
    $\forall \beta \in R_\To^\Obl[\alpha, i]$ either \\
    (1) $\beta$ is applicable at index $i$, or\\
    (2) \=$\exists \gamma \in R^\Obl[\nu, j] \cup R^\Perm[\nu]$ s.t. $\gamma$ cautiously conflicts with $\beta$, s.t.\\
        \>(1) $\gamma$ is applicable (at index $j$), and \\
      \>(2) \= $\forall \zeta \in R^\Obl[\chi, k]$ s.t. $\zeta$ cautiously conflicts with $\gamma$ either\\
		\>\> (1) \= $\zeta$ is discarded (at index $k$), or\\
		\>\> (2) (1) $\zeta \not> \gamma$, and\\
		    \>\>\> (2) $\gamma  > \zeta$ or $C(\zeta) \not> C(\gamma)$.
\end{tabbing} 
\end{definition}

\begin{definition}[Permission Meta-Proof Conditions -- Cautious Variant]\label{def:PermMetaPT-V2}

\begin{tabbing}
  $+\partial_\Perm^m \alpha$: \=If $P(n+1)=+\partial_\Perm^m \alpha$ then\+\\
  $\exists \beta \in R_\To^\Perm[\alpha]$ s.t.\\ 
  (1) $\beta$ is applicable, and\\
  (2) \=$\forall \gamma \in R^\Obl[\nu, j]\cup R^\Perm[\nu]$ s.t. $\gamma$ cautiously conflicts with $\beta$,  either\\
        \>(1) $\gamma$ is discarded at index $j$, or \\
      \>(2) \= $\exists \zeta \in R^\Obl[\chi, k] \cup R^\Perm[\chi]$ s.t. $\zeta$ cautiously conflicts with $\gamma$\\
		\>\> (1) \= $\zeta$ is applicable (at index $k$), and either\\
		    \>\>(2) \=(1) $\zeta > \gamma$, or\\
		    \>\>\>(2) $\gamma \not > \zeta$, and $C(\zeta) > C(\gamma)$.
\end{tabbing} 

\begin{tabbing}
  $-\partial_\Perm^m \alpha$: \=If $P(n+1)=-\partial_\Perm^m \alpha$ then\+\\
  $\forall \beta \in R_\To^\Perm[\alpha]$ either\\ 
  (1) $\beta$ is discarded, or\\
  (2) \=$\exists \gamma \in R^\Obl[\nu, j]\cup R^\Perm[\nu]$ s.t. $\gamma$ cautiously conflicts with $\beta$, s.t.\\
        \>(1) $\gamma$ is applicable at index $j$, and \\
      \>(2) \= $\forall \zeta \in R^\Obl[\chi, k] \cup R^\Perm[\chi]$ s.t. $\zeta$ cautiously conflicts with $\gamma$ either\\
		\>\> (1) \= $\zeta$ is discarded (at index $k$), or\\
		    \>\>(2) \=(1) $\zeta \not> \gamma$, and\\
		    \>\>\>(2) $\gamma > \zeta$ or $C(\zeta) \not> C(\gamma)$.
\end{tabbing} 

\end{definition}

\begin{example}\label{ex:MetaV2-superiority}
Consider theory $D = (F = \set{c}, R, >=\set{(\eta,\theta)})$, with $R= \{$
\begin{align*}
    &\dRule\alpha:\dots=>\Cost {(\dRule\beta:a=>\Perm b)} &
    &\dRule\gamma:\dots=>\Cost {(\dRule\epsilon:a=>\Perm \neg b)}\\
    &\dRule\eta: c=>\Perm d &
    &\dRule\theta: c=>\Perm {\neg d}\}.
\end{align*}
Rules $\alpha$ and $\gamma$  conflict with one another since $\beta$ cautiously conflicts with $\epsilon$
(and the other way around). Rules $\alpha$ and $\gamma$ are both applicable (thus neither a discarded, nor a stronger rule exists). Therefore, $\alpha$ satisfies Condition (2.2.2) of $-\partial^m_\Cost$  (Definition~\ref{def:ConstMetaPT-V2}) to derive $-\partial^m_\Cost\gamma$, and $\gamma$ does the same to prove $-\partial^m_\Cost\alpha$. 

Let us focus on $\eta$ and $\theta$. These two rules are in $R$; thus, even if they are in a cautious conflict (Case 2 of Definition~\ref{def:RationalConflict}), we can prove them with $\Cost$ modality. Their antecedent $c$ is a fact; hence, they are applicable. Can we prove $\Perm p$ and $\Perm\neg p$? The answer is positive. Condition (2.2) of $+\partial_\Perm$ (Definition~\ref{def:StandardPermProof}) specifies that only obligation rules can attack a permission rule. Ergo, the two rules are individually unopposed, and we conclude $+\partial_\Perm p$ and $+\partial_\Perm\neg p$. Rules $\beta$ and $\epsilon$ have the same rule structure of $\eta$ and $\theta$, and one might ask if preventing the inclusion of both rules is meaningful. Again, from the legal drafting point of view, where the aim is to include only provisions that add content that would not be drawn otherwise, a better alternative would be to adopt weak permission, where something is permitted if it is not possible to derive the obligation to the contrary. In this view, $p$ is permitted when we prove $-\partial_\Obl\non p$; for a detailed analysis of how to model weak permission in DDL, see \cite{GovernatoriORS13}.

Note that, even if we had an instance of the superiority relation between $\eta$ and $\theta$, it would play no role whatsoever in establishing the provability of the two permissions. However, the superiority relation is relevant when the two permission rules are the conclusion of meta-rules. The proof conditions for the cautious variant of DDL allow us to consider two cases. 

Case $\alpha>\gamma$. This is the same circumstance adopted by the simple conflict variant. The legal intuition is that meta-rules are norms from a higher order in a legal hierarchy which provides a method to solve the conflict. Here, we conclude $+\partial^m_\Cost\beta$ and $-\partial^m_\Cost\epsilon$, and if $a$ is derivable, we are entitled to conclude $+\partial_\Perm b$. 

Case  $\epsilon>\beta$.
Here, the normative system is equipped with a mechanism to solve the conflict between the (standard) rules, even if such a mechanism (as is the case for permission) turns out to be irrelevant for the direct solution. In this instance, we reverse the result of the case above, deriving $+\partial^m_\Cost\epsilon$ and $+\partial_\Perm\neg b$.
\end{example}

Note that the superiority relation $>$ is defined over the union of  standard rules and
meta-rules; thus, it is possible to have an instance where a standard rule and a meta-rule are 
involved. This is useful for cases with rules like: 
\begin{align*}
    &\dRule\alpha:a=>\Obl b &
    &\dRule\beta:{(\dRule\gamma:c=>\Cost d)}=>\Obl\neg b.
\end{align*}
Here, it is meaningful to add a superiority relation, for instance $\alpha >\beta$, given the
conclusions of the two rules are opposite. Moreover, the proof conditions for the cautious variants
allow us to use instances on meta-rules and instances on rules for the derivation of rules.  
Also, it is possible to have opposing information about the strength of rules. 
\begin{example}\label{ex:referee}
Consider the a defeasible with the following applicable rules\footnote{We are very grateful to one of the anonymous referees for pointing out the example.}
\begin{align*}
    &\dRule\alpha_1:A(\alpha_1)=>\Cost{(\dRule\gamma:a=>\Cost b)}&
    &\dRule\alpha_2:A(\alpha_2)=>\Cost{(\dRule\gamma:a=>\Cost b)}\\
    &\dRule\beta_1:A(\beta_1)=>\Cost{(\dRule\zeta:a=>\Cost\neg b)}&
    &\dRule\beta_2:A(\beta_2)=>\Cost{(\dRule\zeta:a=>\Cost\neg b)}.
\end{align*}
According to Condition 2 of Definition~\ref{def:RationalConflict} $\gamma$ and $\zeta$ cautiously conflict: they have the same antecedent and opposite conclusions. It is hence meaningful to establish that one prevails over the other; let us $\zeta > \gamma$. Given that $\gamma$ and $\zeta$ conflict, the conclusions of the $\alpha$ rules cautiously conflict with the conclusion of the $\beta$ rules (Condition 5 of Definition~\ref{def:RationalConflict}). 

Therefore, again, it is meaningful to have instances of the superiority relation. Suppose we have $\alpha_1 > \beta_1$, $\beta_1 > \alpha_2$ and $\alpha_2 > \beta_2$, and that all meta-rules are body-applicable. We ask whether $+\partial^m_\Cost\gamma$? We thus have to check the conditions set in Definition~\ref{def:ConstMetaPT-V2}:
since $\alpha_1$ is applicable by Condition (2.2.1), we then have to corroborate Condition (2.2.2). The two $\beta$ rules are applicable as well, thus we have to verify if they are defeated by stronger and applicable (meta-)rules; they are given $\alpha_1 > \beta_1$ and $\alpha_2 > \beta_2$. Accordingly, we can use Condition (2.2.2.2.1) in both cases to satisfy (2.2.2), and so conclude $+\partial^m_\Cost\gamma$. 

We now turn our attention to $+\partial^m_\Cost\zeta$. Similarly to the other case we have an applicable rule for it (any of the two $\beta$s will do), and again there exists an applicable (meta-)rule for a rule conflicting with $\zeta$ (the two $\alpha$s). We have to rebut them: (i) For $\alpha_2$ -- as for the case of $\gamma$ above -- we can apply Condition (2.2.2.2.1) and the instance of $>$ on the meta-rules $\beta_2 > \alpha_2$; (ii) For $\alpha_1$ we do not have $\alpha_1>\beta_2$, but we have $C(\beta_2)=\zeta>\gamma=C(\alpha_1)$, that which satisfies Condition (2.2.2.2.2). Hence, we conclude $+\partial^m_\Cost\zeta$. 

The intuitions of Condition (2.2.2.2) of Definition~\ref{def:ConstMetaPT-V2}, as well as of Condition (2.2.2) of Definitions~\ref{def:OblMetaPT-V2} and \ref{def:PermMetaPT-V2}, is to provide a mechanism to solve conflicts over meta-rules. If a preference over meta-rules exists, use it; otherwise, check whether we have a preference over the conflicting (standard) rules. Accordingly, instances of the superiority relation on standard rules can fill gaps in the superiority relation over meta-rules. However, as the example illustrates, we can end with a cycle in the extended superiority relation when we fill these gaps. The superiority relation over the standard rule 
prefers $\beta$ to $\alpha$, while the superiority relation over the 
meta-rules indicates a preference for $\alpha$ over $\beta$.  Thus, 
in some sense, the theory contains some inconsistencies, and conflicting rules can be derived (see Theorem~\ref{th:SoundCompl} for the proper result).
\end{example}
We shall use $D\vdash_C\pm\# c$ to indicate that there is a proof of $\pm\# c$ from $D$ using the proof conditions for Cautious Conflict Defeasible Deontic Logic.

\subsection{Formal Properties of the Logical Apparatus}\label{subsec:Properties}

In this subsection, we prove two properties of the logic showing 
that the variants of the logic do not produce inconsistent results
unless the theory we started with contains some inconsistency. 
In addition, we refine the definition of \textit{extension} and the definition of \textit{theory equivalence}. These two notions play 
an important role to prove the results in Section~\ref{sec:Algo}.

\begin{theorem}[Coherence]\label{prop:coerenza}
Let $D$ be a defeasible deontic theory. There is no pair of tagged modal formulas $+\# x$ and $-\# x$ such that $D\vdash_{L}+\# x$ and $D\vdash_{L}-\#x$, for $L\in\set{S,C}$.
\end{theorem}
\begin{proof}
The result follows from Theorem 1 of \cite{DBLP:journals/igpl/GovernatoriPRS09} proving that if the proof conditions for a positive proof tag and the corresponding negative are the strong negation of each other, then no theory can prove both $+\# x$ and $-\# x$ for any conclusion $x$. It is immediate to verify that all pairs of proof conditions defined in this paper obey the principle of strong negation.
\end{proof}

For the next results we have to provide an additional definition 
(to capture the scenario depicted in Example~\ref{ex:referee}). 
Given a defeasible deontic meta-theory $D$, the extended superiority 
relation $\succ$ is defined as follows: 
\begin{equation}
	{\succ} = {>}\cup\set{(\alpha,\beta)| (\alpha,\beta) \notin {>} \text{ and } (C(\alpha),C(\beta))\in {>}} 
\end{equation} 

\begin{theorem}[Consistence]\label{prop:consistenza}
Let $D$ be defeasible deontic theory such that the superiority relation $>$ is acyclic\footnote{A relation is acyclic if the transitive closure of it does not contain a cycle.}. For any literal $l$, and rules $\alpha$ and $\varphi$:
\begin{enumerate}
    \item If $D\vdash_{L}+\partial_\Cost l$ and $D\vdash_{L}+\partial_\Cost\neg l$, then $l, \neg l\in F$, for $L\in\set{S,C}$;
    
    \item It is no possible to prove both $D\vdash_{L}+\partial_\Obl l$ and  $D\vdash_{L}+\partial_\Obl \neg l$, for $L\in\set{S,C}$.
    
    \item If $D\vdash_{S}+\partial^m_\Box \alpha$ and $D\vdash_{S}+\partial^m_\Box\varphi$, such that $\alpha$ simply conflicts with $\varphi$, then $\alpha, \varphi\in R$; 
 \end{enumerate}
  In addition if $\succ$ is acyclic:
  \begin{enumerate}[resume]  
    \item If $D\vdash_{C}+\partial^m_\Box \alpha$ and $D\vdash_{C}+\partial^m_\Box\varphi$, such that $\alpha$ cautiously conflicts with $\varphi$, then $\alpha, \varphi\in R$. 
\end{enumerate}
\end{theorem}
\begin{proof}
It is immediate to verify that, according to Definitions~\ref{def:MetaApplicability} and \ref{def:MetaDiscardability}, no rule is applicable and discarded (for the same literal)
at the same time.  All proof conditions have the same structure that specifies that there is an applicable rule and every non-discarded rule for a conflicting conclusion is defeated by a stronger applicable rule. Suppose that $l$ and $\neg l$ ($\alpha$ and $\beta$) are both provable for $\Box$. This means that rules $\beta\in R[l]$ and $\gamma\in R[\neg l]$, for $l$ and $\neg l$, exist and are applicable. Given that both $l$ and $\neg l$ are provable for $\Box$, then for every applicable rule $\beta$ for $l$, there is an applicable rule $\gamma'$ for $\neg l$ such that $\gamma' > \beta$, and the other way around. However, since the rule set is finite, the situation is possible if, only if, there is a cycle in the superiority relation (see the proof of Theorem 3.5 of \cite{billington93}). A contradiction. 

For the cases for $\Cost$, condition (1) of the appropriate proof conditions allows us to conclude $+\partial_\Cost l$, $+\partial^m_\Cost\alpha$ when $l\in F$ and $\alpha\in R$; however, when $l\in F$ ($\alpha\in R$) it is not possible to use condition (2) to derive $+\partial_\Cost\neg l$ ($+\partial^m_\Cost\beta$), but Condition (1) applies, thus $\neg l\in F$ ($\beta\in R$). Note
that in addition to the situations captured by $>$ or $\succ$ being cyclic (and for which we can replicate as in the previous case the argument of Theorem 3.5 of \cite{billington93}) Cases 3. and 4. are not possible when $\Box$ is either $\Obl$ or $\Perm$, since we do not allow deontic literal in $\FACTS$ and there are no deontic rule expressions in $R$.
\end{proof}
As the theorem shows, and as illustrated in Example~\ref{ex:referee}, it is still possible to derive conflicting rules (for the 
cautions variants). However, the derivation of one literal and its 
complement is impossible (unless they are given as facts or there is a cycle in the superiority relation, both cases where the theory we started with contains some inconsistency). 

We can now modify the interpretation of extension of Definition~\ref{def:StandardExtension} to adapt the notions discussed above. First, we revise the notion of Herbrand Base to include rule labels. Hence 
$\HB(D)=\set{l,\non l \in\PLit|\,l \text{ appears in }D}\cup\set{\alpha,\neg\alpha\in\LABELS|\, \alpha \text{ appears in }D}$, second we define the so called modal Herbrand Base to close the element of $\HB(D)$ under the various modalities; thus $\MHB(D)=\set{\Box e|\; e\in\HB(D) \wedge \Box\in\set{\Cost,\Obl,\Perm}}$. 

\begin{definition}[Extension]\label{def:Extension}
	Given a defeasible deontic theory $D$, we define the \emph{extension} of $D$ according to variant $L$, $L\in\set{S,R}$ of the logic as $E_L(D) = (+\partial_\Box, -\partial_\Box, +\partial^m_\Box, -\partial^{m}_\Box)$, where $\pm\partial_\Box = \set{l\in\PLit |\, l\in\HB(D)\wedge D\vdash_L \pm\partial_\Box l}$ and $\pm\partial^m_\Box = \set{\alpha\in\LABELS |\, \alpha\in\HB(D)\wedge D\vdash_L \pm\partial^m_\Box l}$, with $\Box \in \set{\Cost, \Obl, \Perm}$.
\end{definition}
We say that two theories are equivalent when they conclude the same conclusions, that is, they have the same extension. Formally:

\begin{definition}[Theory Equivalence]\label{def:Equivalence}
	Two defeasible deontic theories $D$ and $D'$ are \emph{equivalent}, notationally $D \equiv D'$, iff $E_L(D) \equiv E_L(D')$.
\end{definition}
The notion of theory equivalence will play a key role in proving that our algorithms are sound and complete.

%% file: Algorithms.tex
The algorithms presented in this work compute the extension of a defeasible deontic theory with meta-rules; they take inspiration from previous versions \cite{GovernatoriORS13,Olivieri202169,DBLP:conf/icail/GovernatoriO21}. 

The main idea of the algorithms is to compute, at each iteration step, a \emph{simpler} yet equivalent theory than the one we had at the previous step. By `simpler theory', we mean that, by proving and disproving/rejecting literals and standard rules, we can progressively do two actions: 
\begin{enumerate}

    \item \textbf{Simplify the antecedents} of the rules of the theory by removing (previously) proved elements;
    
    \item \textbf{Eliminate (discarded) rules} from the theory itself.
\end{enumerate}
To better understand how such actions operate and, most importantly, why they are possible, we firstly want the reader to focus on some key concepts of Defeasible Logic and its implementations.

The first concept is that, being our logical apparatus non-monotonic, when we want to prove something (no matter whether a statement or a rule), we cannot consider \emph{only a single rule} supporting such a conclusion; we need to consider, at the same time, (i) All the rules supporting it, as well as (ii) All the rules opposing it and thus supporting the opposite conclusion, as such rules must be ``defeated''. (Naturally, if the ``opposing'' team prevails, we shall prove the opposite conclusion, whereas if neither team wins, we refute both.)

Do we really have to defeat all the supporting/opposing rules to prove our conclusion? The answer to that question is: No, we do not \emph{have} to consider the discarded rules, but only the applicable ones.

It is natural then to ask what makes a rule applicable/discarded. The rule set (and the superiority relation establishing the relative strength of rules) should be as general as possible to describe as many situations as possible. 

On the other hand, the set of facts is used to describe the current, specific situation. Imagine, for instance, all the rules describing the normative corpus of driving. There are rules prescribing to fasten the seat belts while driving, others prescribing to wear a helmet while riding a motorbike. What settles if we need to fasten our seat belt or if we have to wear a helmet is the set of facts. Once the set of facts is fed as input, of all the rules present in the initial rule sets, some will be applicable (and they can hence contribute to the team fight), whereas some are discarded (and thus it is like they were not even present in the initial rule sets given that specific set of facts).

When we move from standard to meta-rules, two considerations are in order. 

First, the set of rules supporting a given literal may not be (entirely) determined at the start of the computation. Standard rules can be conclusions of meta-rules; to determine the derivability/refutability of such a literal (and its opposite), we need to wait until all the supporting and opposing standard rules that are conclusions of meta-rules have been either proved or refuted.

Second, a rule being discarded is different from a rule being not provable. Therefore, we cannot discard a meta-rule because its antecedent contains a discarded standard rule: the standard rule can be discarded, but still be provable.

Our algorithms make good use of these concepts: as the input is a theory with a \emph{specific} set of facts (thus describing a \emph{specific situation}), they reduce, step after step, the number of rules of the theory itself by eliminating, at iteration $i+1$, all those rules that have, in their antecedent, an element that has been \emph{refuted} at iteration $i$ (as such rules are discarded by Definition~\ref{def:MetaDiscardability}). 

Symmetrically, if an element has been proved at iteration $i$, at iteration $i+1$, we can safely remove it from the antecedents of all the rules it appears in, as their applicability does not depend any longer on such an element.
 
Roughly said, a rule is applicable when everything in its antecedent has been proved. Consequently, a rule with an empty antecedent is always vacuously applicable, as there is nothing (more) to prove.

A rule is discarded if (at least) one of elements of its antecedent has been previously rejected. When a rule is discarded, it can no longer play a part in neither supporting its conclusion nor rejecting the opposite. Accordingly (as it shall be proved afterwards), theories with or without a discarded rule have the same extension.

Let $D = (\FACTS, R, {>}= \set{(\beta, \theta), (\theta, \zeta), (\beta, \psi)})$ be a theory such that 
\begin{align*}
    R = \{&\alpha\colon \dots \To_\Cost z &&\beta\colon b, z \To_\Cost \neg l && \zeta\colon \dots \To_\Cost \neg l\\ 
    &\theta\colon \neg z \To_\Cost \neg l && \psi\colon \dots \To_\Cost l\},
\end{align*}
and assume that, at iteration $i$, the algorithm proves $+\partial_\Cost z$: at the same at iteration $i$, the algorithm proves  $-\partial_\Cost \neg z$ as well (Procedure~\ref{alg:ProveLiteral}, Lines~\ref{AlgCheckLit:ProveL-ProveL-Cost} and \ref{AlgCheckLit:ProveL-RefuteNonL-Cost}). At iteration $i+1$, we can hence perform two operations:
\begin{enumerate}
    \item Remove $z$ from $\beta$'s antecedent: $\beta$'s applicability no longer depends upon $z$, but solely on $b$, the only remaining element in its antecedent; the ``new $\beta$'' is
    \[
    \beta\colon b \To_\Cost \neg l.
    \]
    \item Eliminate $\theta$ from $R$: $\theta$ is discarded given that $\neg z \in A(\theta)$; therefore,
    \[
    R^{i+1} = R^i \setminus \set{\theta}.
    \]
\end{enumerate}
Naturally, once $\theta$ has been eliminated from the rule set, we can adjust the superiority relation accordingly by removing all those tuples that contain $\theta$ -- in this case, $(\theta, \alpha)$ and ($\beta, \theta$). The only ``potentially'' supporting rule for $l$ is now $\psi$, which is ``potentially'' defeated by $\beta$ (``potentially'', as we need to establish whether $\beta$ and $\psi$ are applicable).

\subsection*{Description of Algorithm~\ref{Alg:ALG} \textsc{Deontic Meta Extension}} The main Algorithm~\ref{Alg:ALG} starts by initialising 
\begin{itemize}
    \item The $\partial$ sets for the extension (Line 1).
    
    \item The Modal Herbrand Base $\MHB$, used to properly cycle over literals (Lines~\ref{ALG:MainFor-Literals-BEGIN}--\ref{ALG:MainFor-Literals-END}), and over standard rules (Lines~\ref{ALG:MainFor-Rules-BEGIN}--\ref{ALG:MainFor-Rules-END}) of the theory.

    \item Support sets $R^\Box[]$ (Line~\ref{ALG:For-LinMHB-R}), one for each literal in the modal Herbrand Base; given a modal literal $X l$, $R^X[l]$ plays a fundamental role in proving/refuting it, as some standard rules ($\alpha$ in the algorithm) supporting $X l$ may not be in the initial rule set, but be the conclusion of meta-rules ($\beta$ in the algorithm). The decidability of $X l$ (and naturally that of $\neg Xl$) must thus be postponed until we have proved/rejected such `$\alpha$ rules'.

    \item Support sets $R[]^\Box_{\infd}$ (Line~\ref{ALG:For-LinMHB-Rinfd}), one for each literal in $\MHB$, and used during the ``team fight''; given a modal literal $X l$, $R[l]^X_{\infd}$ stores the rules supporting $l$ that which are defeated by an \emph{applicable} rule for $\non l$ (for a more detailed explanation, see below during description of Procedure~\ref{alg:CheckLiteral}).

\input{AlgorithmMain}

\enlargethispage{3\baselineskip}
    \item Support sets $R[\alpha]^\Box_{opp}$ and $R[\alpha]^\Box_{supp}$, by invoking Procedure~\ref{alg:Conflicts}. Such sets are determined according to which $variant$ of conflict (simple or cautious) is given as input; such sets correspond to either: (i) Conditions 1--2 of Definition~\ref{def:SimpleConflict} if $variant = simply$, or (ii) Conditions 1--5 of Definition~\ref{def:RationalConflict} if $variant = cautiously$. We decided to report the simple variant sets in full (Lines~\ref{AlgConfl:Ropp-Cost-V1}--\ref{AlgConfl:Rsupp-Cost-V1}, \ref{AlgConfl:Ropp-Obl-V1}--\ref{AlgConfl:Rsupp-Obl-V1}, and \ref{AlgConfl:Ropp-Perm-V1}--\ref{AlgConfl:Rsupp-Perm-V1}) to provide to the reader a deeper understanding of the signatures of the rules involved: $\non \varphi$ has same antecedent, modality, and conclusion of $\alpha$, whereas $\chi$ is either $\alpha$, or $\varphi$. This was not necessary for the cautious variant.

\input{AlgConflicts}

\newpage

    \item Support matrices $\alpha[2][k]$\footnote{We chose indices ranging $1$ to $n$, instead of typical implementations, for the sake of readability.)} for every obligation rule (\textbf{for} cycle at Line~\ref{ALG:OblRulesArrayArray-BEGIN}--\ref{ALG:OblRulesArrayArray-END}), where $k$ is the number of elements in the $\otimes$-chain. Such matrices are to verify the applicability of an obligation rule at a given index $j$ of its $\otimes$-chain according to Condition 4 of Definition~\ref{def:MetaApplicability}.
\end{itemize}

Cycle \textbf{for} of Line~\ref{ALG:ForFacts} proves (as constitutive\footnote{Note that facts are constitutive statements, i.e., plain literal.}) all literals in the initial set of facts, and rejects all opposites. Symmetrically, every rule in the initial rule set (thus not standard rules that can be derived through meta-rules) are proved as constitutive rules (Line~\ref{ALG:ForFactsRules-Prove}), and we refute standard rules that (i) are conclusions of meta-rules, and (ii) conflict with a given rule (Line~\ref{ALG:ForFactsRules-Refute}).

The algorithm now enters the main cycle (\textbf{repeat--until}, Lines~\ref{ALG:Repeat}--\ref{ALG:Until}); cycle \textbf{for} at Lines~\ref{ALG:MainFor-Literals-BEGIN}--\ref{ALG:MainFor-Literals-END} runs over literals, the one at Lines~\ref{ALG:MainFor-Rules-BEGIN}--\ref{ALG:MainFor-Rules-END} runs over rules: they behave in an identical way. Naturally, if there is no rule supporting an element in the Modal Herbrand Base, we refute it (\textbf{if} at Line~\ref{ALG:lIf-RefuteL} for literals, \textbf{if} at Line~\ref{ALG:lIf-RefuteAlfa} for rules). On the other hand, if there exists an applicable rule $\beta$ that has been proved, we invoke \ref{alg:CheckLiteral}, or \ref{alg:CheckRule}, according to whether the element is a literal or a rule (see below the detailed description on how they operate) to verify if, at the current iteration, we can already prove the element at hand; note that the \textbf{if} check at Line~\ref{ALG:CheckLiterals-Obl} is specific for proving an element as obligatory, and differs from the one at Line~\ref{ALG:CheckLiterals-CostPerm} (same for Line~\ref{ALG:CheckRules-Obl} with respect to Line~\ref{ALG:CheckRules-CostPerm}) because we must consider $\otimes$-chains, hence satisfy Condition 4 of Definition~\ref{def:MetaApplicability}.

The algorithm terminates when no modifications to the extension are made. We use the convention that $\Box$ and $\blacksquare$ represent all three modalities whilst $\Diamond$ is restricted to union of obligations and permissions.

\subsection*{Description of Procedure~\ref{alg:CheckLiteral}.} Once an \emph{applicable} and proved rule $\beta$ for $l$ has been found, we can update the set of the opposite and \emph{defeated} rules $R[\non l]_{\infd}$ with all the $\gamma$ rules that are defeated by $\beta$. We remind the reader that 
\begin{itemize}
    \item Only constitutive rules oppose constitutive to rules ($X = \Cost$, Line~\ref{AlgCheckLit:CaseCost-Rinfd});
    
    \item Both obligation and permission rules oppose obligation rules ($X = \Obl$, Lines~\ref{AlgCheckLit:CaseObl-Rinfd-Obl} and \ref{AlgCheckLit:CaseObl-Rinfd-Perm});
    
    \item Only obligation rules oppose permissive rules ($X = \Perm$, Line~\ref{AlgCheckLit:CasePerm-Rinfd}).
\end{itemize}

If there are no opposite rules stronger than $\beta$, we can refute the contrary conclusion (Lines~\ref{AlgCheckLit:lIf-RefuteNonL-Cost}, \ref{AlgCheckLit:lIf-RefuteNonL-Obl}, and \ref{AlgCheckLit:lIf-RefuteNonL-Perm}).

Moreover, if: (i) all the opposite rules are defeated ($R[\non l] \setminus R[\non l]_{\infd} = \emptyset$), (ii) and at least one of the supporting, applicable rules (for $X l$) is a defeasible rule (thus, not a defeater), then we can prove $Xl$ (and we do not even care whether the remaining rules in $R[\non l]$ are applicable or discarded -- Lines~\ref{AlgCheckLit:If-ProveL-Cost-BEGIN}, \ref{AlgCheckLit:If-ProveL-Obl-BEGIN}, and \ref{AlgCheckLit:If-ProveL-Perm-BEGIN}). Note that: (i) when we prove a literal as obligatory, (i.a) we prove it as permitted as well and (i.b) we refute the opposite as both (Lines~\ref{AlgCheckLit:CaseObl-ProveRefuteObl} and \ref{AlgCheckLit:CaseObl-ProveRefutePerm}), and (ii) when we prove a literal as permitted, we refute the opposite as obligatory but not as permitted (if it is permitted to smoke, ``not smoking'' cannot be mandatory but it can be permitted -- Lines~\ref{AlgCheckLit:CasePerm-ProvePerm} and \ref{AlgCheckLit:CasePerm-RefuteObl}).

\input{AlgCheckLiteral}

\subsection*{Description of Procedures \ref{alg:ProveLiteral} and \ref{alg:RefuteLiteral}}

As described at the start of this section, every time we prove a literal, we can remove it from each antecedent it appears in; as negative deontic literals (e.g., $\neg\Obl \non l$ which is satisfied when $+\partial_\Obl l$ or $+\partial_\Perm l$) may be in the antecedents as well, Procedure~\ref{alg:ProveLiteral} takes care of such situations at Lines~\ref{AlgProveLit:CaseObl-UpdateRp}, \ref{AlgProveLit:CaseObl-UpdateRalfa}, \ref{AlgProveLit:CasePerm-UpdateRp}, and \ref{AlgProveLit:CasePerm-UpdateRalfa}.

\input{AlgProveLiteral}

Orthogonally, every time we refute a literal, Procedure~\ref{alg:RefuteLiteral} removes all those rules that have such a literal in their antecedent. 

Finally, we update the support matrices $\alpha[][]$, according to Condition 4.a of Definitions~\ref{def:MetaApplicability} and \ref{def:MetaDiscardability} at Lines~\ref{AlgProveLit:CaseCost-N2}, \ref{AlgProveLit:CaseCost-J2}, and \ref{AlgProveLit:CaseObl-N1} of Procedure~\ref{alg:ProveLiteral}, and Line~\ref{AlgRefuteLit:UpdateBetaN1} of Procedure~\ref{alg:RefuteLiteral}.

\input{AlgRefuteLiteral}

\subsection*{Description of Procedures \ref{alg:CheckRule}, \ref{alg:ProveRule} and \ref{alg:RefuteRule}}

\smallskip

We shall discuss only key distinctions for the procedures for literals, as the sequence of operations and the basic ideas of Procedures~\ref{alg:CheckRule}, \ref{alg:ProveRule} and \ref{alg:RefuteRule} are fundamentally equivalent to each other.

\newpage
\FloatBarrier

\input{AlgCheckRules}
\FloatBarrier

Procedure \ref{alg:CheckRule} is invoked when we are trying to prove a certain rule $\alpha$, with modality $X$, and we found an applicable and proved rule $\beta$ supporting $\alpha$. To verify that the opposite rules are defeated, we need to distinguish between the two conflict variants. If $variant = simply$, then $R[\alpha]^X_{\infd}$ will contains all those $\gamma$ rules that simply conflict with and are defeated by $\beta$. If $variant = cautiously$, we need to consider more rules: not just the $\gamma$s defeated by $\beta$, but also those $\gamma$s \emph{not} stronger than $\beta$ such that their conclusion is defeater by $\alpha$ itself. This perfectly mirrors Condition (2.2.2.2) of Definitions~\ref{def:ConstMetaPT-V2}, \ref{def:OblMetaPT-V2}, and \ref{def:PermMetaPT-V2}.

Lastly, the way of how matrices $\alpha[][]$ are updated at Lines~\ref{AlgProveRule:CaseCost-BetaN2} and \ref{AlgProveRule:CaseOblBetaN1} of Procedure~\ref{alg:ProveRule} (resp. Lines~\ref{alg:RefuteRule-CaseCost-BetaN2} and \ref{algRefuteRule-CaseObl-BetaN1} of Procedure~\ref{alg:RefuteRule}) are because here we have to satisfy Condition 4.b of Definition~\ref{def:MetaApplicability} (and \emph{not} Condition 4.a of Definition~\ref{def:MetaDiscardability}).

\enlargethispage{10\baselineskip}

\input{AlgsRule}
\FloatBarrier

\subsection*{Algorithms Execution}

We end this part by proposing a couple of input theories and analysing the runs of the algorithms in the computation of the extension. Each theory tackles specific characteristics of our algorithms and therefore of our logic. We will pay attention only to the relevant and non-trivial details.

\vspace{5mm}
\noindent Consider $D = (\set{f_1, f_2}, R, \set{(\gamma, \theta)})$ to be a theory such that 
\begin{align*}
    R = \{&\alpha\colon (\gamma\colon \neg f_1 \To_\Cost a ) \To_\Cost b && \beta\colon f_2 \To_\Cost (\gamma\colon \neg f_1 \To_\Cost a)\\ 
    & \zeta\colon (\nu\colon f_1 \To_\Cost c)  \To_\Cost (\kappa\colon f_2 \To_\Perm \neg a) &&\theta\colon f_1, f_2 \To_\Cost \neg a\\
    &\mu\colon f_2 \To_\Obl a \otimes b \otimes c\},
\end{align*}
Support sets $R^\Box[l]$ are instantiated; what is relevant here are $R^\Cost[a] = \set{\gamma}$, $R^\Cost[\neg a] = \set{\theta}$, $R^\Obl[a] = \set{\mu}$, and $R^\Perm[\neg a] = \set{\kappa}$. The only obligation matrix created is $\mu[][]$ ($2 \times 3$), as $\mu$ is the only obligation rule.

Procedure \ref{alg:ProveLiteral} is invoked on $f_1$ and $f_2$, which results in: (i) both literals are added to $+\partial_\Cost$,  (ii) emptying $A(\beta)$, $A(\nu)$, $A(\kappa)$, $A(\theta)$, and (iii) $A(\mu)$.

Procedure~\ref{alg:RefuteLiteral} is symmetrically invoked on $\neg f_1$ and $\neg f_2$, which results in: (i) removing $\gamma$ from $R^\Cost[a]$ as this makes $\gamma$ to be discarded (note that now there are no more supporting rules to conclude $a$ as constitutive), and (ii) removing $(\gamma, \theta)$ from the superiority relation. Important to highlight here that, even if $\gamma$ is discarded, this does not influence the applicability/discardability of $\alpha$, which depends only on proving either $+\partial^m_\Cost\gamma$ or $-\partial^m_\Cost\gamma$. Rules $\alpha$, $\beta$, $\zeta$, $\theta$, and $\mu$ are all proved as constitutive (thus $\alpha, \beta, \zeta, \theta, \mu \in +\partial^{m}_\Cost$).

The algorithm finally enters the main \textbf{repeat--until} cycle; we shall assume that the loop \textbf{for} at Lines~\ref{ALG:MainFor-Literals-BEGIN}--\ref{ALG:MainFor-Literals-END} first controls whether we can prove $b$ as constitutive. Indeed there exists a constitutive rule for $b$ that has been proved: $\alpha$. Procedure~\ref{alg:CheckLiteral} is thus invoked with $b$, $\Cost$, and $\alpha$, as parameters. The first half of the check $\{\, R[\neg b] \setminus R[\neg b]_{\infd} = \emptyset \wedge \exists \zeta \in R^\Cost_\de[l].\, A(\zeta) = \emptyset \,\}$ is in fact satisfied as there are no rules supporting $\neg b$, but not the second half as applicability/discardability of $\alpha$ has yet to be determined; anyhow, $\neg b$ has no support and Procedure~\ref{alg:RefuteLiteral} is hence invoked, with the only result to eliminate it from $\MHB$ and add it to $-\partial_\Cost$. 

Assume that the loop \textbf{for} at Lines~\ref{ALG:MainFor-Literals-BEGIN}--\ref{ALG:MainFor-Literals-END} now checks $\neg a$ for constitutive; $\theta$ is applicable and no rules support $a$ (we do not need to wait for $\gamma$ to be proved/rejected as we already know that, if proved, $\gamma$ is discarded as previously we computed that $\neg f_1 \in -\partial_\Cost$). Accordingly, Procedure~\ref{alg:ProveLiteral}: (i) computes $\neg a \in +\partial_\Cost$, and (ii) updates $\mu[2][1]$ to $+$. Analogously to the previous case, we cannot yet decide for $a$ as obligation since there is potentially the permissive rule $\kappa$ for $\neg a$, and we thus need to wait until we compute either $\pm\partial_\Cost\kappa$.

The algorithm passes then to run the loop \textbf{for} at Lines~\ref{ALG:MainFor-Rules-BEGIN}--\ref{ALG:MainFor-Rules-END}, and let us consider $\gamma$; $\beta$ is applicable and no rule conflicts with it. We hence compute $\gamma\in +\partial_\Cost^{meta}$, which in turn updates $\{\,A(\alpha) \setminus \set{\alpha} = \emptyset\,\}$ ($\alpha$ is now applicable). On the contrary, no rules support $\nu$, thus in cascade: (i) $\nu\in -\partial_\Cost^{meta}$, (ii) $\zeta$ is discarded, hence (iii) no rules support $\kappa$ and so $\kappa\in -\partial_\Cost^{meta}$.

At the next iteration of the \textbf{repeat--until},  (i) we compute $b \in +\partial_\Cost$ and update $\mu[2][2]$ to $-$, (ii) $a \in +\partial_\Obl$ and we update $\mu[1][1]$ to $+$. This makes $\mu$ applicable at index $2$ for $b$, and we compute $b \in +\partial_\Obl$. Lastly, as $\mu[1][2] = +$ and $\mu[2][2] = -$ makes $\mu$ discarded at index $3$ for $c$, the algorithm computes $c \in -\partial_\Obl$.

\medskip

\noindent The next example illustrates: (i) how the algorithms work on the chains of meta-rules, and (ii) the computation of the supporting sets $opp$ and $supp$ for the cautious conflict variant, including cases where the conflict is not restricted to the negations of the rules.

\enlargethispage{2\baselineskip}
Consider $D = (\FACTS, R, {>}= \set{(\alpha, \beta), (\alpha, \lambda)})$ to be a theory such that 
\begin{align*}
    R = \{&\dRule\alpha:\dots=>\Obl {(\dRule\eta:a=>\Perm b) \otimes c \otimes (\dRule\kappa:c=>\Obl d\otimes e)}\\ 
    & \dRule\beta:\dots=>\Perm{(\dRule\theta:a=>\Obl\neg b)}\qquad \ \dRule\gamma:\dots=>\Obl{\neg(\dRule\zeta:a=>\Obl\neg b)}\\
    &\dRule\lambda:\dots=>\Obl {(\dRule\mu:c=>\Obl d)}\qquad \ \dRule\psi:\dots=>\Cost \neg c\},
\end{align*}
where for the sake of simplicity we assume that all the rules' antecedents have been emptied, and would thus have satisfied Conditions 1--3 of Definition~\ref{def:MetaApplicability}. According to Definitions~\ref{def:SimpleConflict} and \ref{def:RationalConflict}, we notice that:
\begin{enumerate}
    \item\label{enum:1} $\eta$ cautiously conflicts with $\theta$, hence $\alpha$ conflicts with $\beta$ (at index 1);
    \item\label{enum:2} $\theta$ simply conflicts with $\zeta$, hence $\beta$ conflicts with $\gamma$;
    \item For \ref{enum:1} and \ref{enum:2}, $\gamma$ supports $\alpha$ (and the other way around);
    \item $\kappa$ cautiously conflicts with $\mu$, hence $\lambda$ cautiously conflicts with $\alpha$ (at index 3).
\end{enumerate}
Ergo, $R[\alpha]_{opp}^\Obl = \set{\beta, \lambda}$, $R[\alpha]_{supp}^\Obl = \set{\gamma}$, and so on. When Procedure~\ref{alg:CheckRule} is invoked on $\eta$, \textbf{if} check of Procedure~\ref{alg:ProveRule} at Line~\ref{AlgCheckRules:If-ProveAlfa-Obl-BEGIN} is satisfied we compute $\eta, \neg\zeta \in +\partial_\Obl^{meta}$, $\theta \in -\partial_\Perm^{meta}$ as well as $\theta \in -\partial_\Obl^{meta}$.

As later on, the algorithms compute $\neg c \in +\partial_\Cost$ as well as $c \in +\partial_\Obl$ (and thus $\alpha[j][2]$, $j = 1, 2$, are updated to $+$), then $\alpha$ becomes applicable at index 3: given that $\alpha > \lambda$, we compute $\kappa \in +\partial_\Obl^{meta}$ and $\mu \in -\partial_\Obl^{meta}$.

Note that, if we had considered the simple conflict variant, $\alpha$ would have not conflicted with $\beta$ and since the superiority relation does not solve the conflict between $\beta$ and $\gamma$, we would have computed $\neg\zeta \in -\partial_\Obl^{meta}$ instead of $\neg\zeta \in +\partial_\Obl^{meta}$.

\subsection{Computational Properties}\label{subsec:CompProperties}

We discuss the computational properties of the algorithms presented. In order to discuss termination and computational complexity, we start by defining the \emph{size} of a theory $D$ as $\Sigma(D)$ to be the number of the occurrences of literals plus the number of occurrences of rules plus 2 for every tuple in the superiority relation. Consequently, theory $D = (\FACTS = \{a, b, c\}$, $R = \set{(\alpha\colon a \To_\Obl d), (\beta\colon b \To \non d), \big(\gamma\colon c \To (\zeta\colon a \To d)\big)}$, $> = \{(\zeta, \beta)\})$ has size $3 + 11 + 2 = 16$.

Note that, by implementing hash tables with pointers to rules where a given literal occurs, each rule can be accessed in constant time. We also implement hash tables for the tuples of the superiority relation where a given rule appears as either one of the two elements, and even those can be accessed in constant time.

\begin{lemma}\label{lemm:ComplexityProveRefute}
    Procedures~\ref{alg:ProveLiteral}, \ref{alg:RefuteLiteral}, \ref{alg:ProveRule}, and \ref{alg:RefuteRule} terminate and their complexity is $O(\Sigma^2)$.
\end{lemma}
\begin{proof}
    Termination of such procedures is straightforward, as (i) the size of the input theory is finite, (ii) we modify finite sets only, and (iii) all the \textbf{for} cycles loop on a finite number of elements, as the Modal Herbrand Base is finite.
    
    Given that all set assignments/modifications are linear in the size of the theory and that all the \textbf{for} cycles are iterated $|MHB| \in O(\Sigma)$ times, this proves our claim setting their complexity to $O(\Sigma^2)$. 
\end{proof}

\begin{lemma}\label{lemm:ComplexityCheckLiteral}
    Procedure~\ref{alg:CheckLiteral} terminates and its complexity is $O(\Sigma^3)$.
\end{lemma}
\begin{proof}
    Termination is straightforward (see motivations above) and by the fact that Lemma~\ref{lemm:ComplexityProveRefute} proves that its inner Procedures~\ref{alg:ProveLiteral} and \ref{alg:RefuteLiteral} terminate.
    
    Again, all set assignments/modifications are linear in the size of the theory, and so are all the \textbf{if} checks that which invoke Procedures~\ref{alg:ProveLiteral} and \ref{alg:RefuteLiteral}. Accordingly, $O(\Sigma) + O(\Sigma) * O(\Sigma^2) = O(\Sigma^3)$. 
\end{proof}

\begin{lemma}\label{lemm:ComplexityConflicts}
    Procedure~\ref{alg:Conflicts} terminates and its complexity is: $O(\Sigma^3)$ if $variant = simply$, or $O(\Sigma^4)$ if $variant = cautiously$.
\end{lemma}
\begin{proof}

Termination is guaranteed by the same considerations above and from what follows, depending on the variant.

If $variant = simply$, set assignments operate on finite sets and: (i) $R^X_{opp}$ are linear, (ii) $R^X_{supp}$ are in $O(\Sigma^2)$. This sets the overall complexity to $O(\Sigma^3)$.

If $variant = cautiously$, we report hereafter how a set $R^X_{opp}$ is.
\[
\begin{split}
    R[\alpha]_{opp}^X  \gets  R[\alpha]_{opp}^X &\cup \set{\gamma\in R[\varphi]|\, C(\alpha) = c_1\otimes \dots\otimes c_m \\
     & \wedge C(\varphi) = d_1\otimes\dots\otimes d_n \wedge
        \big ( A(\alpha) = A(\varphi) \big ) \wedge \\
     &\big ( \exists i \leq m, n.\, \forall k < i.\, ( c_k = d_k \wedge c_i = \non d_i ) \\
     &\vee ( m < n \wedge \forall j \leq m.\, c_j = d_j ) \big ) }.
\end{split}
\]
Such a set is finite, and is computed in $O(\Sigma^2)$  (which implies that $R^X_{supp}$ is in $O(\Sigma^3)$). As the main \textbf{for} cycle at Lines~\ref{AlgConfl:ForAlfaSupports-BEGIN}--\ref{AlgConfl:ForAlfaSupports-END} is in $O(\Sigma)$, this proves our claim. 
\end{proof}
As before, Procedure~\ref{alg:Conflicts} uses hash tables to store such information, and so after its iteration to verify whether a certain rule $variant$ conflicts with another requires constant time, whereas to identify all the rules that $variant$ conflict with a certain rule requires linear time.

\begin{lemma}\label{lemm:ComplexityCheckRule}
    Procedure~\ref{alg:CheckRule} terminates and its complexity is $O(\Sigma^4)$.
\end{lemma}
\begin{proof}
    Termination is straightforward (see motivations above) and given by the fact that Lemma~\ref{lemm:ComplexityProveRefute} proves that its inner Procedures~\ref{alg:ProveRule} and \ref{alg:RefuteRule} terminate.
    
    Again, all set assignments/modifications are linear in the size of the theory, even the ones requiring the verification of \emph{variant conflicts with} based on the considerations in (and after) Lemma~\ref{lemm:ComplexityConflicts}.
    
    Given that all the \textbf{if} checks and the \textbf{for} cycles are linear as well, this proves that the overall complexity is $O(\Sigma) + (\Sigma) * (O(\Sigma^2) + O(\Sigma) * O(\Sigma^2)) = O(\Sigma^4)$. 
\end{proof}

\begin{theorem}\label{th:Complexity}
	Algorithm~\ref{Alg:ALG} \textsc{Deontic Meta Extension} terminates and its complexity is $O(\Sigma^5)$.
\end{theorem}
    
\begin{proof}
	
Termination of Algorithm~\ref{Alg:ALG} \textsc{Deontic Meta Extension} is given by Lemmas~\ref{lemm:ComplexityProveRefute}--\ref{lemm:ComplexityCheckRule}, and bound to the termination of the \textbf{repeat-until} cycle at Lines~\ref{ALG:Repeat}--\ref{ALG:Until}, as all other cycles loop over finite sets of elements of the order of $O(\Sigma)$. Given that (i) the Modal Herbrand Base $\MHB$ is finite and (ii) since, every time a literal or a rule is proved/refuted, they are removed from $\MHB$, thus the algorithm eventually empties such sets, and, at the next iteration, no modification to the extension can be made. This proves the termination of Algorithm~\ref{Alg:ALG}.

Regarding its complexity, let us notice that: (i) all set modifications are in linear time, and (ii) the aforementioned \textbf{repeat-until} cycle is iterated at most $O(\Sigma)$ times, and so are the two \textbf{for} loops at Lines \ref{ALG:MainFor-Literals-BEGIN}--\ref{ALG:MainFor-Literals-END} and \ref{ALG:MainFor-Rules-BEGIN}--\ref{ALG:MainFor-Rules-END}. This would suggest that the \textbf{repeat-until} cycle would contribute to the overall complexity for a factor of $O(\Sigma^2)$. A more discerning analysis shows that the complexity is actually $O(\Sigma)$, as the complexity of each \textbf{for} loop cannot be considered separately from the complexity of the external loop (they are strictly dependent on one another). Indeed, the overall number of operations made by the sum of all loop iterations cannot outrun the number of occurrences of the literals or rules ($O(\Sigma)+ O(\Sigma)$), because the operations in the inner cycles directly decrease, iteration after iteration, the number of the remaining repetitions of the out-most loop, and the other way around. 

Based on the results of Lemmas~\ref{lemm:ComplexityProveRefute}--\ref{lemm:ComplexityCheckRule}, and the fact that the complexity of the \textbf{repeat-until} cycle, which is $O(\Sigma^5) = O(\Sigma) * (O(\Sigma^2) + O(\Sigma^3) + O(\Sigma^2) + O(\Sigma^4))$, dominates all the operations in the first part of the algorithm, we have that the overall complexity is $O(\Sigma^5)$.
\end{proof}
The final result concerns the soundness and correctness of the algorithm, in the sense that the extension computed by the algorithm corresponds to the set of provable/refutable literals/rules.
\begin{theorem}\label{th:SoundCompl}
	Algorithm~\ref{Alg:ALG} \textsc{Deontic Meta Extension} is sound and complete, that is, for $\Box\in\set{\Cost,\Obl,\Perm}$: 
\begin{enumerate}
	 \item $D \vdash_L +\partial_\Box p$ iff $p \in +\partial_\Box p$ of $E_L(D)$, $p \in \LIT$
	 \item $D \vdash_L +\partial^{m}_\Box \alpha$ iff $p \in +\partial^m_\Box \alpha$ of $E_L(D)$, $\alpha \in \LABELS$
	 \item  $D \vdash_L -\partial_\Box p$ iff $p \in -\partial_\Box p$ of $E_L(D)$, $p \in \LIT$
	 \item $D \vdash_L -\partial^{m}_\Box \alpha$ iff $p \in -\partial^{m}_\Box \alpha$ of $E_L(D)$, $\alpha \in \LABELS$.
\end{enumerate}

\end{theorem}

\begin{proof} (Sketch)
The aim of Algorithm~\ref{Alg:ALG} \textsc{Deontic Meta Extension} is to compute the extension of the input theory through successive transformations on the set of facts, rules, and the superiority relation. Such transformations allow us: (i) to obtain a simpler theory, (ii) while retaining the same extension. By simpler theory we mean a theory with less symbols in it. Note that if $D\vdash_L +\partial_\Box l$ then $D\vdash_L -\partial_L \non l$, and that if $D\vdash_L +\partial^{m}_\Box \alpha$ then $D\vdash_L -\partial^{m}_\Box \gamma$, with  $\alpha$ conflicting with $\gamma$.

Suppose that the algorithm computes $+\partial_\Box l$ or $+\partial^m_\Box\alpha$ (meaning that $l \in +\partial_\Box$, or $\alpha\in +\partial^{m}_\Box$). Accordingly, we remove $l/\Box l$ or $\alpha/\Box\alpha$ from every antecedent where it appears in as, by Definition~\ref{def:MetaApplicability}, the applicability of such rules will not depend any longer on $l/\Box l$ or $\alpha\Box\alpha$, but only on the remaining elements in their antecedents. Moreover, we can eliminate from the rule sets all those rules with $\non l/\Box\non l$ or $\gamma/\Box\gamma$ in their antecedent (with $\alpha$ conflicting with $\gamma$), as such rules are discarded by Definition~\ref{def:MetaDiscardability} (and then adjust the superiority relation accordingly). Finally,  proving $+\partial^m_\Cost\alpha$ makes $\alpha$ to become active in supporting its conclusion, and rebutting the opposite.

The proof follows the schemas of the ones in \cite{GovernatoriORS13,GovernatoriOSRC16}, and proves that the original theory $D$ and the simpler theory $D'$ are equivalent. 

Formally, suppose that $D\vdash_L +\partial_\Box l$ (symmetrically $D\vdash_L +\partial^{m}_\Box \alpha$) at $P(n)$. $R'$ of $D'$ is obtained by the following transformation. 
Given a standard rule $\alpha\colon A(\alpha)\hookrightarrow_\blacksquare C(\alpha)$ and a literal $l$, $\alpha\ominus l$ is the rule
\[
 \alpha\colon A(\alpha)\setminus \set{l} \hookrightarrow_\blacksquare C(\alpha).
\]
Given a meta-rule rule $\alpha\colon A(\alpha)\hookrightarrow_\blacksquare C(\alpha)$ and a literal $l$, $\alpha\ominus l$ is the rule
\[
 \alpha\colon A(\alpha)[\beta/\beta\ominus l]\setminus\set{l} \hookrightarrow_\Box C(\alpha),
\]
for all rules $\beta\in A(\alpha)$, where $A(\alpha)[\beta/\beta\ominus l]$ denotes the substitution of $\beta$ in $\alpha$ with $\beta\ominus l$.

Furthermore, if $\alpha\ominus\beta$ is the rule
\[
    \alpha\colon A(\alpha)\setminus\set{\beta}\hookrightarrow_\blacksquare C(\alpha),
\]
then\footnote{If $\Box=\Cost$, in the transformation $\Box l$ is just $l$, following the conditions of Definition~\ref{def:MetaApplicability} to make a rule applicable, and $\non\Box l$ is $\non l$ according to Definition~\ref{def:MetaDiscardability} to discard a rule.} 
\[
R'=\set{\alpha\ominus \Box l|\,\alpha\in R}\setminus \set{\alpha\in R|\, \non\Box l\in A(\alpha)}
\]
for  $D\vdash_L +\partial_\Box l$, while for $D\vdash_L +\partial^m_\Box\alpha$ we use
\[
R'=\set{\beta\ominus \Box\alpha|\,\beta\in R}\setminus \set{\beta\in R|\, \non\Box \alpha\in A(\beta)}
\]
Finally, if $>'$ of $D'$ is obtained from $>$ of $D$ as follows
\[
	{>'} = {>} \setminus \set{(\beta,\zeta),(\zeta,\beta)\, |\, \non l \in A(\zeta) \text{ or } \non \gamma \in A(\zeta)},
\]
then by induction on the length of a proof, we can show that
\begin{itemize}
	\item $D\vdash_L \pm\partial_\Box p$ iff $D'\vdash_L \pm\partial_\Box p$, and
	\item $D\vdash_L \pm\partial^{m}_\Box \chi$ iff $D'\vdash_L \pm\partial^{m}_\Box \chi$.
\end{itemize}

The key step to proof the equivalences above is that: when we prove $l$ (resp., $\alpha$), we can transform a derivation $P$ in $D$ into a derivation $P'$ in $D'$. We do so by removing from $P$, first, all steps where $l$ occurs, and then all subsequent steps justified by steps where $l$ occurred. It is immediate to see that if a rule $\beta$ is applicable in $D$ at a particular step in $P$, then the version of $\beta$ in $D'$ (if any) is applicable in $D'$ at some step in $P'$.

The next proprieties to conclude the proof of the theorem are the following: if  $R$ contains a rule $\alpha$ such that $A(\alpha)=\emptyset$, then
\begin{enumerate}
    \item If $\set{\beta\in R|\, \beta \text{ conflicts with } \alpha\wedge \beta>\alpha}=\emptyset$, then $D\vdash_L-\partial_\Box \non C(\alpha)$. 
    \item If $\set{\beta\in R|\, \beta \text{ conflicts with  }\alpha\wedge \beta>\alpha}\setminus\{\beta\in R|\, \beta$  conflicts with  $\alpha\wedge \exists\gamma, \gamma$ conflicts with $\beta\wedge A(\gamma)=\emptyset\wedge
    \gamma>\beta\}=\emptyset$,\footnote{For $L=S$ we require the additional condition that $\gamma$ is either $\alpha$ or $\non\beta$, and for meta-rules and $L=R$, we have to consider not only the meta-rules stronger than $\beta$, but the standard rules stronger than $C(\beta)$.} then $D\vdash_L-\partial_\Box \non C(\alpha)$. 
\end{enumerate}
First of all, $\alpha$ is applicable. Then for 1., it is a witness of an applicable and undefeated rule for $\non C(\alpha)$. It is mundane to verify that
it satisfies the  $\exists\gamma\dots$ clause of the proof conditions for the various $-\partial$.  Similarly, for 2., if all rules conflicting with $\alpha$ have been defeated (the construction, denoted by $R[]_{\infd}$ in the algorithms, ensures that every rule opposing $\alpha$ is defeated by an applicable rule), satisfying the $\exists\zeta$ clauses of the proof conditions for the various $+\partial$. Condition 1. is used by Procedures~\ref{alg:RefuteLiteral} and \ref{alg:RefuteRule}, and Condition 2. by the Procedures~\ref{alg:ProveLiteral} and \ref{alg:ProveRule} to populate the extension of a theory.

\end{proof}

%% file: AlgorithmMain.tex
\begin{algorithm}[p]
    \fontsize{9}{9}
    \SetAlgoLined
    \LinesNumbered
    
\KwIn{Defeasible deontic theory $D$, \texttt{global} $variant = \{$simply, cautiously$\}$}
    
\KwOut{The extension $E_L(D)$}

$\pm\partial_\Box \gets \emptyset$; $\pm\partial_\Box^{meta} \gets \emptyset$\;\label{ALGv1:Rapp}
  
\textsc{modalHerbrandBase($\MHB$)}\label{ALG:InitMHB}

\For{$\Box l \in \MHB$}{\label{ALG:For-LinMHB-BEGIN}
    $R^\Box[l] \gets \big\{\alpha \text{ appears in } R |\,  \alpha \in R^\Box[l]  \vee  \big (\exists \beta \in R^\blacksquare[\alpha] \wedge l \in C(\alpha) \text{ s.t. } \alpha\colon A(\alpha) \hookrightarrow_\Box C(\alpha) \big ) \big\}$\;\label{ALG:For-LinMHB-R}
    
    $R[l]^\Box_{\infd} \gets \emptyset$\;\label{ALG:For-LinMHB-Rinfd}

}\label{ALG:For-LinMHB-END}
    
\textsc{Conflicts}\;\label{ALG:Conflict}

\For{$\alpha \in R^\Obl$}{\label{ALG:OblRulesArrayArray-BEGIN}
    $k = $ \texttt{length($C(\alpha)$)}, \textsc{Build matrix} $\alpha[2][k]$\;
    \lFor{$i = 1$ to $2$, $j = 1$ to $k$}{\textsc{Set} $\alpha[i][j]$ to $null$}
}\label{ALG:OblRulesArrayArray-END}

\lFor{$l\in \FACTS$}{
    \textsc{Prove($l, \Cost$)}; \textsc{Refute($\non l, \Cost$)}
}\label{ALG:ForFacts}

\For{$\alpha\in R$}{\label{ALG:ForFactsRules-BEGIN}
    
    \textsc{Prove($\alpha, \Cost$)}\;\label{ALG:ForFactsRules-Prove}
    
    \textsc{Refute($\gamma, \Cost$)}, s.t. $\gamma\notin R$ and $\alpha$ $variant$ conflicts with $\gamma$\;\label{ALG:ForFactsRules-Refute}

}\label{ALG:ForFactsRules-END}

\Repeat{$\partial^+_\Box = \emptyset \wedge \partial^-_\Box = \emptyset \wedge \partial^+_{meta, \Box} = \emptyset \wedge \partial_{meta, \Box}^- = \emptyset$}{\label{ALG:Repeat}
    
$\partial^\pm_\Box \gets \emptyset$\;\label{ALG:ResetPartialPM}
        
\For{$\Box l\in \MHB$}{\label{ALG:MainFor-Literals-BEGIN}
        
    \lIf{$R^\Box[l]=\emptyset$}{$\textsc{Refute}(l, \Box)$}\label{ALG:lIf-RefuteL}
    
    \Switch{modality $\Box$}{\label{ALG:SwitchLiteral-BEGIN}  
        
        \Case{$\Box \in \set{\Cost, \Perm}$}{
            \lIf{$\exists \beta \in R^\Box[l].\, A(\beta)=\emptyset \wedge \beta \in +\partial^{meta}_\Box$}{\textsc{CheckLiteral($l, \Box, \beta$)}}\label{ALG:CheckLiterals-CostPerm}         
        }    
    
        \Case{$\Box = \Obl$}{
            \lIf{$ \exists \beta \in R^\Obl[l, i].\, A(\beta)=\emptyset \wedge \forall j < i.\, \beta[1][j] = + \text{ and } \beta[2][j] =  + \wedge \beta \in +\partial^{meta}_\Box$}{\textsc{CheckLiteral($l, \Obl, \beta$)}}\label{ALG:CheckLiterals-Obl}
        }
    }\label{ALG:SwitchLiteral-END}
 
}\label{ALG:MainFor-Literals-END}
    
$\pm\partial_\Box \gets \pm\partial_\Box \cup \partial^\pm_\Box$\;\label{ALG:UpdatePartialPM}

$\partial_{meta, \Box}^\pm \gets \emptyset$\;\label{ALG:ResetPartialPmMeta}
     
\For{$\Box\alpha \in MHB$}{\label{ALG:MainFor-Rules-BEGIN}
     
    \lIf{$R^\Box[\alpha] = \emptyset$}{$\textsc{Refute}(\alpha, \Box)$}\label{ALG:lIf-RefuteAlfa}
    
    \Switch{modality $\Box$}{\label{ALG:SwitchRule-BEGIN}
        
        \Case{$\Box \in \set{\Cost, \Perm}$}{
            \lIf{$\exists \beta \in R^\Box[\alpha].\, A(\beta) = \emptyset$}{\textsc{CheckRule($\alpha, \Box, \beta$)}}\label{ALG:CheckRules-CostPerm}
        }
        
        \Case{$\Box = \Obl$}{
            \lIf{$\exists \beta \in R^\Obl[\alpha, i].\, A(\beta)=\emptyset \wedge \forall j < i.\, \beta[1][j] = + \wedge \beta[2][j] = + $}{\textsc{CheckRule($\alpha, \Obl, \beta$)}}\label{ALG:CheckRules-Obl}
        }
    }\label{ALG:SwitchRule-END}
}\label{ALG:MainFor-Rules-END}
    
$\pm\partial^{meta}_\Box \gets \pm\partial^{meta}_\Box \cup \partial_{meta, \Box}^\pm$\;\label{ALG:UpdatePartialPmMeta}

}\label{ALG:Until}

\Return{$E_L(D) = \big (+\partial_\Box, -\partial_\Box, +\partial^{meta}_\Box, -\partial^{meta}_\Box\big )$}
    
\caption{Deontic Meta Extension}\label{Alg:ALG}
\end{algorithm}

\FloatBarrier

%% file: AlgConflicts.tex
\hspace{-.57cm}
\begin{procedure}[H]
\footnotesize
\SetAlgoLined
\LinesNumbered

\For{$\big ( \alpha\colon A(\alpha) \hookrightarrow_\Box C(\alpha)\big )$ s.t. $\alpha \text{ appears in } R$}{\label{AlgConfl:ForAlfaSupports-BEGIN}

$R^\blacksquare[\alpha] \gets \set{\beta \in R^\blacksquare |\, \alpha \in C(\beta)}$\;

$R[\alpha]^\blacksquare_{\infd} \gets \emptyset$\;
    
\uIf(\tcp*[f]{Simple conflicts}){$variant = simply$}{

\Switch{modality $X$}{\label{AlgConfl:Switch-V1-BEGIN}
    
    \Case{$X = \Cost$}{
        
        $R[\alpha]_{opp}^\Cost  \gets \set{\gamma \in R^\Cost[\non\varphi] |\,  \non\big (\varphi: A(\alpha) \hookrightarrow_\Box C(\alpha) )}$\;\label{AlgConfl:Ropp-Cost-V1}
        
        $R[\alpha]_{supp}^\Cost  \gets \big\{ \zeta \in R^\Cost[\chi] |\,  \exists \gamma \in R[\alpha]_{opp}^\Cost \cap R[\non\varphi].\, \chi \in \set{\alpha, \varphi} \big \}$\;\label{AlgConfl:Rsupp-Cost-V1}
    }
    
    \Case{$X = \Obl$}{
        
        $R[\alpha]_{opp}^\Obl  \gets \set{\gamma \in R^\Diamond[\non\varphi] |\,  \non\big (\varphi: A(\alpha) \hookrightarrow_\Box C(\alpha) )}$\;\label{AlgConfl:Ropp-Obl-V1}
    
        $R[\alpha]_{supp}^\Obl  \gets \big\{ \zeta \in R^\Obl[\chi] |\,  \exists \gamma \in R[\alpha]_{opp}^\Obl \cap R[\non\varphi].\, \chi \in \set{\alpha, \varphi} \big \}$\;\label{AlgConfl:Rsupp-Obl-V1}
    
    }
    
    \Case{$X = \Perm$}{
        
        $R[\alpha]_{opp}^\Perm  \gets \set{\gamma \in R^\Obl[\non\varphi] |\,  \non\big (\varphi: A(\alpha) \hookrightarrow_\Box C(\alpha) )}$\;\label{AlgConfl:Ropp-Perm-V1}
        
        $R[\alpha]_{supp}^\Perm  \gets \big\{ \zeta \in R^\Diamond[\chi] |\,  \exists \gamma \in R[\alpha]_{opp}^\Perm \cap R[\non\varphi].\, \chi \in \set{\alpha, \varphi} \big \}$\;\label{AlgConfl:Rsupp-Perm-V1}
    
    }

}\label{AlgConfl:Switch-V1-END}
}
\Else(\tcp*[f]{Cautious conflicts}){

\Switch{modality $X$}{
    \Case{$X = \Cost$}{
        
        $R[\alpha]_{opp}^\Cost \gets \set{\gamma \in R^\Cost[\varphi] |\, \varphi \text{ cautiously conflicts with } \alpha}$\;\label{AlgConfl:Ropp-Cost-V2}
        
        $R[\alpha]_{supp}^\Cost \gets \set{\zeta \in R^\Cost[\chi] |\, \exists \gamma \in R[\alpha]_{opp}^\Cost  \cap R^\Cost[\varphi].\, \chi \text{ cautiously conflicts with } \varphi}$\;\label{AlgConfl:Rsupp-Cost-V2}
    }
    \Case{$X = \Obl$}{
    
        $R[\alpha]_{opp}^\Obl \gets \set{\gamma \in R^\Diamond[\varphi] |\, \varphi \text{ cautiously conflicts with } \alpha}$\;\label{AlgConfl:Ropp-Obl-V2}
        
        $R[\alpha]_{supp}^\Obl \gets \set{\zeta \in R^\Obl[\chi] |\, \exists \gamma \in R[\alpha]_{opp}^\Obl  \cap R^\Diamond[\varphi].\, \chi \text{  cautiously conflicts with } \varphi}$\;\label{AlgConfl:Rsupp-Obl-V2}
    }
    \Case{$X = \Perm$}{
        
        $R[\alpha]_{opp}^\Perm \gets \set{\gamma \in R^\Diamond[\varphi] |\, \varphi \text{ cautiously conflicts with } \alpha}$\;\label{AlgConfl:Ropp-Perm-V2}
        
        $R[\alpha]_{supp}^\Perm \gets \set{\zeta \in R^\Diamond[\chi] |\, \exists \gamma \in R[\alpha]_{opp}^\Perm  \cap R^\Obl[\varphi].\, \chi \text{ cautiously conflicts with } \varphi}$\;\label{AlgConfl:Rsupp-Perm-V2}
    }
}
}

}\label{AlgConfl:ForAlfaSupports-END}

\caption{Conflicts()}\label{alg:Conflicts}
\end{procedure}

%% file: AlgCheckLiteral.tex
\begin{procedure}[H]
\SetAlgoLined
\LinesNumbered

\KwIn{A plain literal $l \in \PLit$, a modality $X \in \set{\Cost, \Obl, \Perm}$, a rule $\beta$}

\Switch{modality $X$}{\label{AlgCheckLit:Switch-BEGIN}

\Case{$X = \Cost$}{\label{AlgCheckLit:CaseCost-BEGIN}

    $R[\non l]^\Cost_{\infd} \gets R[\non l]^\Cost_{\infd} \cup \set{\gamma \in R^\Cost[\non l]\, |\, \beta > \gamma}$\;\label{AlgCheckLit:CaseCost-Rinfd}
    
    \If{$\set{\gamma \in R^\Cost[\non l]\, |\, \gamma > \beta} =
    \emptyset$}{
        $\textsc{RefuteLiteral}(\non l, \Cost)$\;\label{AlgCheckLit:lIf-RefuteNonL-Cost}
    }
    
    \If{$\big (R^\Cost[\non l] \setminus R[\non l]^\Cost_{\infd} = \emptyset\big ) \wedge \big (\exists \zeta \in R^\Cost_\de[l].\, A(\zeta) = \emptyset \big )$}{\label{AlgCheckLit:If-ProveL-Cost-BEGIN}
    
    $\textsc{ProveLiteral}(l, \Cost)$\;\label{AlgCheckLit:ProveL-ProveL-Cost}
    
    $\textsc{RefuteLiteral}(\non l, \Cost)$\;\label{AlgCheckLit:ProveL-RefuteNonL-Cost}
    
    }\label{AlgCheckLit:If-ProveL-Cost-END}
    
}\label{AlgCheckLit:CaseCost-END}

\Case{$X = \Obl$}{\label{AlgCheckLit:CaseObl-BEGIN}

    $R[\non l]^\Obl_{\infd} \gets R[\non l]^\Obl_{\infd} \cup \set{\gamma \in R^\Diamond[\non l]\, |\, \beta > \gamma}$\;\label{AlgCheckLit:CaseObl-Rinfd-Obl}

    $R[\non l]_{\infd}^\Perm \gets R[\non l]^\Perm_{\infd} \cup \set{\gamma \in R^\Diamond[\non l]\, |\, \beta > \gamma}$\;\label{AlgCheckLit:CaseObl-Rinfd-Perm}    
    
    \If{$\set{\gamma \in ( R^\Obl[\non l] \cup R^\Perm[\non l] )\, |\, \gamma > \beta} =
    \emptyset$}{\label{AlgCheckLit:lIf-RefuteNonL-Obl-BEGIN}
        $\textsc{RefuteLiteral}(\non l, \Obl)$\;\label{AlgCheckLit:lIf-RefuteNonL-Obl}
    }\label{AlgCheckLit:lIf-RefuteNonL-Obl-END}
    
    \If{$\big ( ( R^\Obl[\non l] \cup R^\Perm[\non l] ) \setminus R[\non l]^\Obl_{\infd} = \emptyset\big ) \wedge \big (\exists \zeta \in R^\Obl_\de[l].\, A(\zeta) = \emptyset \big )$}{\label{AlgCheckLit:If-ProveL-Obl-BEGIN}
        $\textsc{ProveLiteral}(l, \Obl)$; $\textsc{RefuteLiteral}(\non l, \Obl)$\;\label{AlgCheckLit:CaseObl-ProveRefuteObl}
        $\textsc{ProveLiteral}(l, \Perm)$; $\textsc{RefuteLiteral}(\non l, \Perm)$\;\label{AlgCheckLit:CaseObl-ProveRefutePerm}
    }\label{AlgCheckLit:If-ProveL-Obl-END}

}\label{AlgCheckLit:CaseObl-END}

\Case{$X = \Perm$}{\label{AlgCheckLit:CasePerm-BEGIN}

    $R[\non l]^\Perm_{\infd} \gets R[\non l]^\Perm_{\infd} \cup \set{\gamma \in R^\Obl[\non l]\, |\, \beta > \gamma}$\;\label{AlgCheckLit:CasePerm-Rinfd}
    
    \If{$\set{\gamma \in R^\Obl[\non l]\, |\, \gamma > \beta} =
    \emptyset$}{
        $\textsc{RefuteLiteral}(\non l, \Obl)$\;\label{AlgCheckLit:lIf-RefuteNonL-Perm}
    }
    
    \If{$\big ( R^\Obl[\non l] \setminus R[\non l]^\Perm_{\infd} = \emptyset \big ) \wedge \big (\exists \zeta \in R^\Perm_\de[l].\, A(\zeta) = \emptyset \big )$}{\label{AlgCheckLit:If-ProveL-Perm-BEGIN}
        $\textsc{ProveLiteral}(l, \Perm)$\;\label{AlgCheckLit:CasePerm-ProvePerm}
        $\textsc{RefuteLiteral}(\non l, \Obl)$\;\label{AlgCheckLit:CasePerm-RefuteObl}
        
    }\label{AlgCheckLit:If-ProveLE-Cost}
    
}\label{AlgCheckLit:CasePerm-END}
}\label{AlgCheckLit:Switch-END}


\caption{CheckLiteral($l, X, \beta$)}\label{alg:CheckLiteral}
\end{procedure}

%% file: AlgProveLiteral.tex
\begin{procedure}[H]
\SetAlgoLined
\LinesNumbered

\KwIn{A literal $l \in \PLit$, a modality $X \in \set{\Cost, \Obl, \Perm}$}

$\partial^+_X \gets \partial^+_X \cup \set{l}$\;\label{AlgProveLit:Update+partial}

$\MHB \gets \MHB \setminus \set{X l}$\;\label{AlgProveLit:Update+FLc}

\Switch{modality $X$}{

\Case{$X = \Cost$}{
   
    \For{$\Box p \in \MHB$}{
        $R^\Box[p] \gets \set{A(\beta) \setminus \set{l} \hookrightarrow_\blacksquare C(\beta) |\, \beta \in R^\Box[p]}$\;\label{AlgProveLit:CaseCost-UpdateRp}
    }
    
    \For{$\Box \alpha \in \MHB$}{
        $R^\Box[\alpha] \gets \set{A(\beta) \setminus \set{l} \hookrightarrow_\blacksquare C(\beta) |\, \beta \in R^\Box[\alpha]}$\;\label{AlgProveLit:CaseCost-UpdateRalfa}
    }

    \lFor{$\beta \in R^\Obl[l, i]$}{$\beta[2][i] \gets -$}\label{AlgProveLit:CaseCost-N2}
        
    \lFor{$\gamma \in R^\Obl[\non l, j]$}{$\gamma[2][j] \gets +$}\label{AlgProveLit:CaseCost-J2}
}

\Case{$X = \Obl$}{

    \For{$\Box p \in \MHB$}{
        $R^\Box[p] \gets \set{A(\beta) \setminus \set{\Obl l, \neg\Obl \non l, \neg\Perm\non l} \hookrightarrow_\blacksquare C(\beta) |\, \beta \in R^\Box[p]} \setminus \set{\beta \in R^\Box[p] |\ \set{\neg\Obl l} \in A(\beta)}$\;\label{AlgProveLit:CaseObl-UpdateRp}
    }
    
    \For{$\Box \alpha \in \MHB$}{
        $R^\Box[\alpha] \gets \set{A(\beta) \setminus \set{\Obl l, \neg\Obl \non l, \neg\Perm\non l} \hookrightarrow_\blacksquare C(\beta) |\, \beta \in R^\Box[\alpha]}  \setminus \set{\beta \in R^\Box[p] |\ \set{\neg\Obl l} \in A(\beta)}$\;\label{AlgProveLit:CaseObl-UpdateRalfa}
        
        $R[\alpha]^\Box_{opp} \gets R[\alpha]^\Box_{opp} \setminus \set{\beta \in R^\Box[p] |\ \set{\neg\Obl l} \in A(\beta)}$\;\label{AlgProveLit:CaseObl-UpdateRalfa-Ropp}
        
        $R[\alpha]^\Box_{supp} \gets R[\alpha]^\Box_{supp} \setminus \set{\beta \in R^\Box[p] |\ \set{\neg\Obl l} \in A(\beta)}$\;\label{AlgProveLit:CaseObl-UpdateRalfa-Rsupp2}
    }
    
    \lFor{$\beta \in R^\Obl[l, i]$}{$\beta[1][i] \gets +$}\label{AlgProveLit:CaseObl-N1}
}

\Case{$X = \Perm$}{

    \For{$\Box p \in \MHB$}{
        $R^\Box[p] \gets \set{A(\beta) \setminus \set{\Perm l, \neg\Obl\non l} \hookrightarrow_\blacksquare C(\beta) |\, \beta \in R^\Box[p]}  \setminus \set{\beta \in R^\Box[p] |\ \set{\neg\Perm l} \in A(\beta)}$\;\label{AlgProveLit:CasePerm-UpdateRp}
    }
    
    \For{$\Box \alpha \in \MHB$}{
        $R^\Box[\alpha] \gets \set{A(\beta) \setminus \set{\Perm l, \neg\Obl\non l} \hookrightarrow_\blacksquare C(\beta) |\, \beta \in R^\Box[\alpha]} \setminus \set{\beta \in R^\Box[p] |\ \set{\neg\Perm l} \in A(\beta)}$\;\label{AlgProveLit:CasePerm-UpdateRalfa}
        
        $R[\alpha]^\Box_{opp} \gets R[\alpha]^\Box_{opp} \setminus \set{\beta \in R^\Box[p] |\ \set{\neg\Perm l} \in A(\beta)}$\;\label{AlgProveLit:CasePerm-UpdateRalfa-Ropp}
        
        $R[\alpha]^\Box_{supp} \gets R[\alpha]^\Box_{supp} \setminus \set{\beta \in R^\Box[p] |\ \set{\neg\Perm l} \in A(\beta)}$\;\label{AlgProveLit:CasePerm-UpdateRalfa-Rsupp}
    }
}

}

\caption{ProveLiteral($l$, $X$)}\label{alg:ProveLiteral}
\end{procedure}

%% file: AlgRefuteLiteral.tex
\begin{procedure}[H]
\SetAlgoLined
\LinesNumbered

\KwIn{A literal $l \in \PLit$, a modality $X \in \set{\Cost, \Obl, \Perm}$}

$\partial^-_X \gets \partial^-_X \cup \set{l}$\;\label{AlgRefuteLit:Update-partial}
	
$\MHB \gets \MHB \setminus \set{Xl}$\;\label{AlgRefuteLit:UpdateFLc}

\uIf{$X =\Cost$}{
    \lFor{$\Box p\in \MHB$}{
        $R^\Box[p]\gets R^\Box[p]\setminus\set{\beta\in R^\Box[p]|\, l\in A(\beta)}$}
    
	\For{$\Box \alpha\in \MHB$}{
        
        $R^\Box[\alpha]\gets R^\Box[\alpha]\setminus\set{\beta\in R^\Box[\alpha]|\, l\in A(\beta)}$\;
    
        $R[\alpha]^\Box_{opp} \gets R[\alpha]^\Box_{opp}\setminus\set{\beta\in R[\alpha]^\Box_{opp} |\, l\in A(\beta)}$\;
        
        $R[\alpha]^\Box_{supp} \gets R[\alpha]^\Box_{opp}\setminus\set{\beta\in R[\alpha]^\Box_{supp} |\, l\in A(\beta)}$
    }

    $> \gets > \setminus \set{(\beta,\gamma), (\gamma, \beta) \in > |\, l\in A(\beta)}$\;\label{AlgRefuteLit:UpdateSuperiorityCost}
}
\Else{

    \For{$\Box p \in \MHB$}{
        $R^\Box[p] \gets \set{A(\beta) \setminus \set{\neg Xl} \hookrightarrow_\blacksquare C(\beta) |\, \beta \in R^\Box[p]} \setminus \set{\beta \in R^\Box[p] |\, Xl \in A(\beta)}$\;}
    
    \For{$\Box \alpha\in \MHB$}{
        
        $R^\Box[\alpha]\gets \set{A(\beta) \setminus \set{\neg Xl} \hookrightarrow_\blacksquare C(\beta) |\, \beta \in R^\Box[\alpha]}  \setminus\set{\beta\in R^\Box[\alpha]|\, Xl\in A(\beta)}$\;
    
        $R[\alpha]^\Box_{opp} \gets  R[\alpha]^\Box_{opp} \setminus \set{\beta\in R[\alpha]^\Box_{opp} |\, Xl\in A(\beta)}$\;
        
        $R[\alpha]^\Box_{supp} \gets  R[\alpha]^\Box_{supp} \setminus \set{\beta\in R[\alpha]^\Box_{supp} |\, Xl\in A(\beta)}$
    }

    $> \gets > \setminus \set{(\beta,\gamma), (\gamma, \beta) \in > |\, X l\in A(\beta)}$\;\label{AlgRefuteLit:UpdateSuperiorityMod}
}

\If{$X = \Obl$}{\lFor{$\beta \in R^\Obl[l, i]$}{$\beta[1][i] \gets -$}\label{AlgRefuteLit:UpdateBetaN1}
}

\caption{RefuteLiteral($l$,$X$)}\label{alg:RefuteLiteral}
\end{procedure}

%% file: AlgCheckRules.tex
\begin{procedure}[p]
\footnotesize
\SetAlgoLined
\LinesNumbered

\KwIn{A rule $\alpha$, a modality $X$, a rule $\beta$}

\Switch{modality $X$}{\label{AlgCheckRules:Switch-BEGIN}

\Case{$X = \Cost$}{\label{AlgCheckRules:CaseCost-BEGIN}
    
    \uIf{$variant = simply$}{
        $R[\alpha]^\Cost_{\infd} \gets R[\alpha]^\Cost_{\infd} \cup \set{\gamma \in R[\alpha]^\Cost_{opp} |\, \beta\ simply \text{ conflicts with }  \gamma \wedge \beta > \gamma}$\;\label{AlgCheckRules:UpdateRinfd-Cost-V1}

    }\Else{
        $R[\alpha]^\Cost_{\infd} \gets R[\alpha]^\Cost_{\infd} \cup \set{\gamma \in R[\alpha]^\Cost_{opp} \cap R^\Cost[\varphi]\, |\, \beta\ cautiously \text{ conflicts with }  \gamma  \wedge (\beta > \gamma) \vee ( \gamma \not> \beta \wedge \alpha > \varphi)  }$\;\label{AlgCheckRules:UpdateRinfd-Cost-V2}
    }
    
    \If{$\big( R[\alpha]^\Cost_{opp} \setminus R[\alpha]^\Cost_{\infd} = \emptyset \big )  \wedge \big (\exists \chi \in R^\Cost_\de[\alpha] \cap R[\alpha]^\Cost_{supp}.\, A(\chi) = \emptyset \big )$}{\label{AlgCheckRules:If-ProveAlfa-Cost-BEGIN}
            
        \textsc{ProveRule}($\alpha, \Cost$)\;
        
        \For{$\varphi \in C(\gamma).\, \gamma \in R[\alpha]^\Cost_{opp} \wedge \varphi\ variant$ conflicts with $\alpha$}{ 
            \textsc{RefuteRule}($\varphi, \Cost$)\;
        }\label{AlgCheckRules:lForRefuteAlfa-Cost}
                
    }\label{AlgCheckRules:If-ProveAlfa-Cost-END}
}\label{AlgCheckRules:CaseCost-END}

\Case{$X = \Obl$}{\label{AlgCheckRules:CaseObl-BEGIN}

    \uIf{$variant = simply$}{
        $R[\alpha]^\Obl_{\infd} \gets R[\alpha]^\Obl_{\infd} \cup \set{\gamma \in R[\alpha]^\Obl_{opp} |\, \beta\ simply \text{ conflicts with }  \gamma \wedge \beta > \gamma}$\;\label{AlgCheckRules:UpdateRinfd-OblObl-V1}

    }\Else{
        $R[\alpha]^\Obl_{\infd} \gets R[\alpha]^\Cost_{\infd} \cup \set{\gamma \in R[\alpha]^\Obl_{opp} \cap R^\Obl[\varphi]\, |\, \beta\ cautiously \text{ conflicts with }  \gamma  \wedge (\beta > \gamma) \vee ( \gamma \not> \beta \wedge \alpha > \varphi)  }$\;\label{AlgCheckRules:UpdateRinfd-OblObl-V2}
    }
    
    \uIf{$variant = simply$}{
        $R[\alpha]^\Perm_{\infd} \gets R[\alpha]^\Perm_{\infd} \cup \set{\gamma \in R[\alpha]^\Obl_{opp} |\, \beta\ simply \text{ conflicts with }  \gamma \wedge \beta > \gamma}$\;\label{AlgCheckRules:UpdateRinfd-OblPerm-V1}

    }\Else{
        $R[\alpha]^\Perm_{\infd} \gets R[\alpha]^\Perm_{\infd} \cup \set{\gamma \in R[\alpha]^\Obl_{opp} \cap R^\Obl[\varphi]\, |\, \beta\ cautiously \text{ conflicts with }  \gamma  \wedge (\beta > \gamma) \vee ( \gamma \not> \beta \wedge \alpha > \varphi)  }$\;\label{AlgCheckRules:UpdateRinfd-OblPerm-V2}
    }
    
    \If{$\big( R[\alpha]^\Obl_{opp} \setminus R[\alpha]^\Obl_{\infd} = \emptyset \big )  \wedge \big (\exists \chi \in R^\Obl_\de[\alpha].\, A(\chi) = \emptyset \big )$}{\label{AlgCheckRules:If-ProveAlfa-Obl-BEGIN}
            
        \textsc{ProveRule($\alpha, \Obl$)}; \textsc{ProveRule($\alpha, \Perm$)}\;         
        
        \For{$\gamma \in R[\alpha]^\Obl_{opp}.\, \varphi \in C(\gamma) \wedge \varphi\ variant \emph{ conflicts with } \alpha$}{
                \textsc{RefuteRule($\varphi,\Obl$)}; \textsc{RefuteRule($\varphi,\Perm$)}\;
        }\label{AlgCheckRules:lForRefuteAlfa-Obl-V2}
            
    }\label{AlgCheckRules:If-ProveAlfa-Obl-END}

}\label{AlgCheckRules:CaseObl-END}

\Case{$X = \Perm$}{\label{AlgCheckRules:CasePerm-BEGIN}
    
    \uIf{$variant = simply$}{
        $R[\alpha]^\Perm_{\infd} \gets R[\alpha]^\Perm_{\infd} \cup \set{\gamma \in R[\alpha]^\Perm_{opp} |\, \beta\ simply \text{ conflicts with }  \gamma \wedge \beta > \gamma}$\;\label{AlgCheckRules:UpdateRinfd-Perm-V1}

    }\Else{
        $R[\alpha]^\Perm_{\infd} \gets R[\alpha]^\Perm_{\infd} \cup \set{\gamma \in R[\alpha]^\Perm_{opp} \cap R^\Diamond[\varphi]\, |\, \beta\ cautiously \text{ conflicts with }  \gamma  \wedge (\beta > \gamma) \vee ( \gamma \not > \beta \wedge \alpha > \varphi)  }$\;\label{AlgCheckRules:UpdateRinfd-Perm-V2}
    }
    
    \If{$ R[\alpha]^\Perm_{opp} \setminus R[\alpha]^\Perm_{\infd} = \emptyset \wedge \exists \chi \in R^\Perm_\de[\alpha].\, A(\chi) = \emptyset$}{\label{AlgCheckRules:If-ProveAlfa-Perm-BEGIN}
            
        \textsc{ProveRule($\alpha, \Perm$)}\;
                
        \For{$\gamma \in R[\alpha]^\Perm_{opp}.\, \varphi \in C(\gamma) \wedge \varphi\ variant \emph{ conflicts with }  \alpha$}{
                $\textsc{RefuteRule}(\varphi, \Obl)$
        }\label{AlgCheckRules:lForRefuteAlfa-Cost-V2}
                
    }\label{AlgCheckRules:If-ProveAlfa-Perm-END}
    
}\label{AlgCheckRules:CasePerm-END}

}\label{AlgCheckRules:Switch-END}

\caption{CheckRule($\alpha, X, \beta$)}\label{alg:CheckRule}
\end{procedure}

%% file: AlgsRule.tex
\begin{procedure}[H]
\SetAlgoLined
\LinesNumbered

\KwIn{A standard rule $\alpha$, a modality $X \in \set{\Cost, \Obl, \Perm}$}

$\partial^+_{X, meta} \gets \partial^+_{X, meta} \cup \set{\alpha}$\;\label{AlgProveRule:Update+partial}

    $\MHB \gets \MHB \setminus \set{X\alpha}$\;\label{AlgProveRule:UpdateFLc}

\Switch{modality $X$}{\label{AlgProveRule:Switch-BEGIN}

\Case{$X = \Cost$}{\label{AlgProveRule:CaseCost-BEGIN}

    \lFor{$\Box p \in \MHB$}{
    $R^\Box[p] \gets \set{A(\beta) \setminus \set{\alpha} \hookrightarrow_\blacksquare C(\beta) |\, \beta \in R^\Box[p]}$
    }
    
    \lFor{$\Box \zeta \in \MHB$}{
    $R^\Box[\zeta] \gets \set{A(\beta) \setminus \set{\alpha} \hookrightarrow_\blacksquare C(\beta) |\, \beta \in R^\Box[\zeta]}$
    }
    
    \lFor{$\beta \in R^\Obl[\alpha, i]$}{$\beta[2][i] \gets -$}\label{AlgProveRule:CaseCost-BetaN2}
}\label{AlgProveRule:CaseCost-END}

\Case{$X = \Obl$}{\label{AlgProveRule:CaseObl-BEGIN}
    
    \For{$\Box p \in \MHB$}{
        $R^\Box[p] \gets \set{A(\beta) \setminus \set{\Obl\alpha, \neg\Obl\non \alpha, \neg\Perm\non \alpha} \hookrightarrow_\blacksquare C(\beta) |\, \beta \in R^\Box[p]} \setminus \set{\beta \in R^\Box[p] |\ \set{\neg\Obl \alpha} \in A(\beta)}$\;\label{AlgProveRule:CaseObl-UpdateR}
    }
    
    \For{$\Box \zeta \in \MHB$}{
    $R^\Box[\zeta] \gets \set{A(\beta) \setminus \set{\Obl\alpha, \neg\Obl\non \alpha, \neg\Perm\non \alpha} \hookrightarrow_\blacksquare C(\beta) |\, \beta \in R^\Box[\zeta]} \setminus \set{\beta \in R^\Box[\zeta] |\ \set{\neg\Obl \alpha} \in A(\beta)}$\;
    
    $R[\zeta]^\Box_{opp} \gets R[\zeta]^\Box_{opp} \setminus \set{\beta \in R^\Box[\zeta] |\ \set{\neg\Obl \alpha} \in A(\beta)}$\;\label{AlgProveLit:CaseObl-UpdateRalfa-Ropp2}
        
    $R[\zeta]^\Box_{supp} \gets R[\zeta]^\Box_{supp} \setminus \set{\beta \in R^\Box[\zeta] |\ \set{\neg\Obl \alpha} \in A(\beta)}$\;\label{AlgProveLit:CaseObl-UpdateRalfa-Rsupp}
    }

    \lFor{$\beta \in R^\Obl[\alpha, i]$}{$\beta[1][i] \gets +$}\label{AlgProveRule:CaseOblBetaN1}
}\label{AlgProveRule:CaseObl-END}

\Case{$X = \Perm$}{

	\For{$\Box p \in \MHB$}{
        $R^\Box[p] \gets \set{A(\beta) \setminus \set{\Perm\alpha, \neg\Obl\non \alpha} \hookrightarrow_\blacksquare C(\beta) |\, \beta \in R^\Box[p]} \setminus \set{\beta \in R^\Box[p] |\ \set{\neg\Perm \alpha} \in A(\beta)}$\;\label{AlgProveRule:CasePerm-UpdateRp}
    }
    
    \For{$\Box \zeta \in \MHB$}{
        $R^\Box[\zeta] \gets \set{A(\beta) \setminus \set{\Perm\alpha, \neg\Obl\non \alpha} \hookrightarrow_\blacksquare C(\beta) |\, \beta \in R^\Box[\zeta]}  \setminus \set{\beta \in R^\Box[\zeta] |\ \set{\neg\Perm \alpha} \in A(\beta)}$\;\label{AlgProveRule:CasePerm-UpdateRalfa}
        
        $R[\zeta]^\Box_{opp} \gets R[\zeta]^\Box_{opp} \setminus \set{\beta \in R^\Box[\zeta] |\ \set{\neg\Perm \alpha} \in A(\beta)}$\;\label{AlgProveLit:CaseSupp-UpdateRalfa-Ropp}
        
    $R[\zeta]^\Box_{supp} \gets R[\zeta]^\Box_{supp} \setminus \set{\beta \in R^\Box[\zeta] |\ \set{\neg\Perm \alpha} \in A(\beta)}$\;\label{AlgProveLit:CaseSupp-UpdateRalfa-Rsupp}
    }

}\label{AlgProveRule:Switch-CasePerm-END}
}\label{AlgProveRule:Switch-END}
    \caption{ProveRule($\alpha$, X)}\label{alg:ProveRule}
\end{procedure}

\newpage
\FloatBarrier

\begin{procedure}[H]
\SetAlgoLined
\LinesNumbered

\KwIn{A standard rule $\alpha: A(\alpha) \hookrightarrow_\Box C(\alpha)$, a modality $X \in \set{\Cost, \Obl, \Perm}$}

$\partial^-_{X, meta} \gets \partial^-_{X, meta} \cup \set{\alpha}$\;\label{algRefuteRule-UpdatePartial}

$\MHB \gets \MHB \setminus \set{X\alpha}$\;\label{algRefuteRule-UpdateMHB}

\uIf{$X = \Cost$}{
    \lFor{$\Box p\in \MHB$}{
        $R^\Box[p]\gets R^\Box[p]\setminus\set{\beta\in R^\Box[p]|\, \alpha\in A(\beta)}$}\label{algRefuteRule-CaseCost-Rp}
        
	\For{$\Box \zeta\in \MHB$}{
        
        $R^\Box[\zeta]\gets R^\Box[\zeta]\setminus\set{\beta\in R^\Box[\zeta]|\, \alpha\in A(\beta)}$\;\label{algRefuteRule-CaseCost-Ralfa}
        
        $R[\zeta]^\Box_{opp} \gets R[\zeta]^\Box_{opp} \setminus \set{\beta\in R^\Box[\zeta]|\, \alpha\in A(\beta)}$\;\label{algRefuteRule-CaseCost-Ralfa-Ropp}
        
        $R[\zeta]^\Box_{supp} \gets R[\zeta]^\Box_{supp} \setminus \set{\beta\in R^\Box[\zeta]|\, \alpha\in A(\beta)}$\;\label{algRefuteRule-CaseCost-Ralfa-Rsupp}
    }

    \lFor{$\beta \in R^\Obl[\alpha, i]$}{$\beta[2][i] \gets +$}\label{alg:RefuteRule-CaseCost-BetaN2}
    
    $>\, \gets\, >\, \setminus \{(\beta, \gamma),(\gamma, \beta) \in > |\, \alpha \subseteq A(\beta) \}$\;\label{algRefuteRule-CaseCost-Superiority}
}
\Else{
    \For{$\Box p \in \MHB$}{
        $R^\Box[p] \gets \set{A(\beta) \setminus \set{\neg X\alpha} \hookrightarrow_\blacksquare C(\beta) |\, \beta \in R^\Box[p]} \setminus \set{\beta \in R^\Box[p] |\, X\alpha \in A(\beta)}$\;\label{algRefuteRule-Else-Rp}
    }
        
    \For{$\Box\zeta \in \MHB$}{
        $R^\Box[\zeta] \gets \set{A(\beta) \setminus \set{\neg X\alpha} \hookrightarrow_\blacksquare C(\beta) |\, \beta \in R^\Box[\zeta]} \setminus \set{\beta \in R^\Box[\zeta] |\, X\alpha \in A(\beta)}$\;\label{algRefuteRule-Else-Ralfa}
        
        $R[\zeta]^\Box_{opp} \gets R[\zeta]^\Box_{opp} \setminus \set{\beta \in R^\Box[\zeta] |\, X\alpha \in A(\beta)}$\;\label{algRefuteRule-Else-Ralfa-Ropp}
        
        $R[\zeta]^\Box_{supp} \gets R[\zeta]^\Box_{supp} \setminus \set{\beta \in R^\Box[\zeta] |\, X\alpha \in A(\beta)}$\;\label{algRefuteRule-Else-Ralfa-Rsupp}
    }    
        
    $>\, \gets\, >\, \setminus \{(\beta, \gamma),(\gamma, \beta) \in > |\, X\alpha \subseteq A(\beta) \}$\;\label{algRefuteRule-Else-Superiority}    
}

\If{$X = \Obl$}{
\lFor{$\beta \in R^\Obl[\alpha, i]$}{$\beta[1][i] \gets -$}\label{algRefuteRule-CaseObl-BetaN1}
}

\caption{RefuteRule($\alpha$, X)}\label{alg:RefuteRule}
\end{procedure}

%% file: relatedworkmetalogic.tex
\section{Related Work}\label{sec:RelatedWork}

The philosophical issue of admitting meta-rules in the logical language was critically discussed for a long time in conditional logics \cite{nute:1980}. At the end of the `60s, such logics have been studied to give formal account to linguistic structures such as indicative conditionals and subjunctive conditionals to express counterfactuals. Later on, conditional logics have also been used to deal with non-monotonic reasoning \cite{Delgrande:87,Farinas:94}. While some,  like \cite{adams:1975}, rejected nested rules when a probabilistic treatment of conditionals is considered, others admitted this possibility even though it was argued that the intuitive meaning of such formulas is not clear \cite{Lewis:1973,nute:1980,Delgrande:87}; meaningful uses of nested conditionals were advanced in \cite{Boutilier:92}.

In mathematical logic and theoretical computer science meta-logic and meta-reasoning have been a discussed theme over the years. A historical perspective towards the argument shows that the topic has been dealt with, from different perspectives, since 1989, when Paulson \cite{Paulson1989363} presented the issues related to the construction of a generic theorem prover and showed that reasoning with the notion of \emph{relevance} as a means to choose the most useful subset of axioms can boost a significant speed-up of the process itself. 

Thenceforth, the discussion stood open for a few years. Numerous scholars borrowed the notion of meta-reasoning from the original presentation of the issue proposed by Tarski \cite{Tarski1986143} where the notion of \emph{metamathematics} is devised as applied to logical frameworks. This was employed in the foundational work by Russell and Wefald \cite{Russell1991361}. The level of discussion provided in Russell and Wefald's study was yet too far from concrete cases, but it has the great merit of providing a reference for the basic idea needed for devising the \emph{motivation} for introducing meta-reasoning systems, defined as \emph{bounded rationality}. More concretely, it became possible to employ these notions in many-valued logics \cite{Murray1994237}, in image processing, a field that is not relevant to this investigation, and in knowledge-based systems by Rowel \emph{et al.} \cite{Rowe19921}.

Almost immediately a significant attention has been posed on the definition of meta-logic methods in logic programming. This investigation line started by the pioneering study by Costantini and Lanzarone \cite{Costantini1994239}, but was inspired by the early investigation by Lloyd and Topor \cite{Lloyd1984225} and further investigated by Brogi \emph{et al.} \cite{Brogi19997,Brogi1998123}, Lifschitz \emph{et al.} \cite{Lifschitz1999369}, and many others. 

Subsequently, a very important novelty in meta-logical systems was introduced by Grundy in 1996 \cite{Grundy1996}: the notion of hierarchical reasoning. The idea behind hierarchical reasoning is that the setting of decision-making does not depend on one single level, but on more than one. Numerous applications have been devised, and in many of these, hierarchical reasoning has shown to be decisive. Papadias \emph{et al.} \cite{Papadias1997251} provided specific algorithms for spatial reasoning, and Seidel \emph{et al.} \cite{Seidel1997239} applied it to probabilistic Constraint Satisfaction Problems. 

Unlike hierarchical reasoning, where different layers of logical frameworks are employed to represent a hierarchy of decisions, meta-reasoning can be based on higher-order logics, where reasoning becomes an element of the logic itself. This approach is the inspiration for methods for the construction of systems with nested rules that we introduced as a base for this research in Section~\ref{sec:Intro}. Original investigation on higher-order logic and meta-reasoning should be attributed to McDowell \emph{et al.} \cite{Mcdowell200280} and independently to Momigliano \emph{et al.} \cite{Momigliano2003375}.

Meta-reasoning has also been the motivation for investigations on a more theoretical level. On the one hand, the notion of continual computation has been deeply investigated as a base for machine reasoning by Horvitz \cite{Horvitz2001159}. His revolutionary notion has shown to be the base of many unified models of reasoning for humans and machines, as discussed widely by Gershman \emph{et al.} in the impactful study on the notion of computational rationality \cite{Gershman2015273}. 
Another aspect of meta-logic that has been dealt with is the \emph{temporal} one, that has been studied for the rewriting methods by Clavel and Meseguer \cite{Clavel2002245}, and by Baldan \emph{et al.} \cite{Baldan20021}. 

To come to meta-reasoning and meta-logic in non-monotonic systems, this work is an extensive revision and substantial extension to the research by Olivieri \emph{et al.} \cite{Olivieri202169}. Their investigation provided a framework for reasoning with meta-rules as a generalisation of the approach employed by Governatori and Rotolo \cite{Governatori2009157} and by Cristani \emph{et al.} \cite{Cristani201739}. These studies have mainly focused on the need for specific meta-rules for the revision of norms and to accommodate, in the form of meta-reasoning, the issues related to the effects of rules over time. Concerning business process compliance, some studies have been developed by Governatori \emph{et al.} \cite{GovernatoriOSRC16,GovernatoriORS13,DBLP:conf/ruleml/GovernatoriOSC11,Olivieri2015603} where some of the issues that we solved here emerged at first.

The specific issue in developing methods for reasoning with rules as the object of an argument has been the focus of the study by Modgil \emph{et al.} \cite{Modgil2011959}. This research focuses mainly on meta-arguments that have been investigated as the base for a general theory of argumentation structures. Finocchiaro has explored these issues critically \cite{Finocchiaro2007253} and interpreted the notion of meta-argument from a general point of view. Similarly, Dhovhannisyan and Djijian \cite{Hovhannisyan2017345} recently developed a general theory of meta-argumentation. 
On the other hand, Boella \emph{et al.} have defined when a meta-argument is acceptable \cite{Boella2009259}.

%% file: Conclusions.tex
The focus of this paper is the efficient computation of rules from meta-rules. In general, the topic of how to use (meta-)rules to generate other rules has received little attention, as discussed in Section \ref{sec:RelatedWork}. In this section, we refer to some works of the late literature that 
constitutes the terrain on which we build this study.

As previously claimed, meta-rules have been considered as a base for norm modification \cite{Governatori2009157,Cristani201739}, and more generally, in \cite{ruleml05insu}, that is dedicated to a logic for deriving rules from meta-rules; however, none of these works investigated the computational complexity, nor addressed the issue of defining algorithms. 

The large majority of the studies that have used meta-rules have focused upon the usage of these as a means to determine the \emph{scope} of rule application or the \emph{result} of the application of the rules. In particular, we can identify two research lines: Logic Programming and   Meta-logic. Logic programming studies investigated the issue of enhancing the expressiveness by allowing nested expressions \cite{Lloyd1984225,Lifschitz1999369}. Nevertheless, these approaches are based on the so-called Lloyd-Toper transformation, that transforms nested expressions into (equivalent) logical expressions. Similarly, in \cite{datalog} disjunctive DATALOG is enriched with nested rules; however, such nested rules may be eliminated by using (stratified) negation, but these are kept because they allow for a more natural correspondence with the natural language description of the underlying problem. 

We have seen in Section ~\ref{sec:Intro} that this approach suffers from several problems, and it is not appropriate for many uses of meta-rules, particularly when the aim is to represent meta-rules as a means to derive rules. Some works, like \cite{GabbayGMO96}, extended Logic Programming with negation with nested hypothetical implications, plus constraints or negation as failure. Specifically, they consider rules with conditional goals in the body but not implications in the head.

The notion of meta-rules and close concepts, including meta-logic \cite{Basin2004528} and meta-reasoning \cite{Dyoub2018}, have been employed widely in Logic Programming \cite{Azab2008211}, but also outside it, specifically in \emph{Context Theory} \cite{Ghidini2001221}. In general, we can look at these studies as methodologically
coherent with the notion of \emph{hierarchical} reasoning, where a method is devised to choose which reasoning process is more appropriate for the specific
scenario in which the process is employed.  
%
A specific line of research (strictly connected with the studies in the semantics of Logic Programming) is the Answer Set Programming (ASP) and preferences \cite{Faber03}. Further on, many studies on ASP where meta-rules took place: unfortunately, such investigations did not focus upon \emph{nested rules}. 

Generation of rules from other rules is one of the research goals in Input/Output logic (IOL) \cite{makinson2000input}. The idea of Input-Output Logic is to define a set of operations on input/output rules (where an input/output rule is a pair $(x,y)$, where $x$ and  $y$ are formulas in a logical language) to derive new input/output pairs. 
The language underlying IOL has been generalised to allow for statements such as $\neg(in A \Rightarrow out B)$ (interpreted as “$A$ does not give a reason for the obligation $B$”), $in A \Rightarrow \neg out B$ ($A$ is a reason against the obligation $B$), and similar combinations, resembling somehow our rule-expressions \cite{strasser}.
Differently from what we do: (1) IOL does not consider nested rules, and (2) the derivation mechanism depends on the properties of the operations on which the IOL variant is defined, not on the rules on which the logic operates.  

Closer to the issues of Deontic Logic are the researches in \emph{argumentation}. The basic concept derived from the combination of meta-logical structures and argumentation is the meta-level argumentation \cite{Modgil2011959}. Applied meta-level has been investigated to develop a framework where, for instance, admissibility of arguments and other issues in this field are dealt with \cite{DupinDeSaint-Cyr201657}. 

Although some previous work has been done, the problem of nested rules in non-monotonic frameworks from a computational complexity viewpoint deserves a deeper study, and this paper fills this gap. In \cite{Olivieri202169}, we have dealt with Defeasible Logic without modal operators and temporal expressions (most of the work on meta-rules considers combinations of such features). In the present work, we have discussed the modalised formulae and their usage in meta-rules.  A related area regards the treatment of preferences over rules.  The issue of generating the preference relation dynamically has been investigated by \cite{PrakkenS95,Antoniou2004463}, and how to revise it is
discussed by Governatori \emph{et al.} \cite{Governatori2019205}. We plan to extend the work presented here to handle the preference relation by adding and removing instances contextually using meta-rules and to study the computational complexity of the resulting logic. 
 The variants of the logic we presented are restricted to one level of nesting. Accordingly, another area of future investigation is how to extend the logic to handle more levels of nesting and, again, determine the complexity of the extension.

Further direct issues concern the introduction of temporal operators: as discussed in \cite{Rotolo:2021}, applicability issues are strongly concerned with time, and, for what concerns the specific of private international law, also space (wrt countries and territories). We plan to extend and combine the algorithms presented here with the algorithms for temporal and spatial logical framework with defeasible deontic meta-logic, and we expect that the computational results to carry over.
